\documentclass{article}

\usepackage{amsthm}
\usepackage{arXiv}

\usepackage[utf8]{inputenc} 
\usepackage[T1]{fontenc}    
\usepackage{hyperref}       
\usepackage{booktabs}       
\usepackage{amsfonts}       
\usepackage{nicefrac}       
\usepackage{microtype}      
\usepackage{lipsum}		
\usepackage{graphicx}

\usepackage{natbib}
\bibpunct[, ]{(}{)}{,}{a}{}{,}%

\usepackage{bm}
\usepackage{listings}
\usepackage{fancyvrb}
\usepackage{subcaption}

\usepackage{amsmath}
\usepackage{amssymb}
\usepackage{ifthen}
\usepackage{url}
\usepackage{graphicx}
\usepackage{color}
\usepackage{array}
\usepackage{url}            

\newcommand{\norm}[1]{\left\lVert#1\right\rVert}

\newcommand{\fracpartial}[2]{\frac{\partial #1}{\partial  #2}}

\newcommand{\Var}[1]{\mathbb{V}\left[#1\right]}

\newtheorem{prop}{Proposition}
\newtheorem{theorem}{Theorem}

\def\x{\mathbf{x}}
\def\X{\mathbf{X}}

\def\Z{\mathbf{Z}}
\def\y{\mathbf{y}}
\def\Y{\mathbf{Y}}

\def\W{\mathbf{W}}
\def\U{\mathbf{U}}

\def\S{\mathbf{S}}
\def\I{\mathbf{I}}
\def\H{\mathbf{H}}
\def\xnew{\mathbf{x}^{(\textrm{new})}}
\def\Xnew{\mathbf{X}^{(\textrm{new})}}

\def\xnewi{{\x}_{i}^{(\textrm{new})}}
\def\xnewt{\mathbf{x}^{(\textrm{new})T}}
\def\treated{\textrm{treated}}
\def\untreated{\textrm{untreated}}
\def\R{\mathbb{R}}
\def\blambda{\bm{\lambda}}
\def\bxi{\bm{\xi}}
\def\bbeta{\bm{\beta}}

\def\one{\mathbf{1}}
\def\grad{\bigtriangledown}
\def\tpn{t_{(1-\alpha/2),(n-p-1)}} 
\def\SSE{\textrm{SSE}} 

\def\Zp{\bm{Z}_{\textrm{p}}}
\def\Zo{\bm{Z}_{\textrm{o}}}
\def\Zh{\bm{Z}_{\textrm{h}}}

\def\fp{f_{\textrm{p}}}
\def\fo{f_{\textrm{o}}}
\def\fh{f_{\textrm{h}}}

\def\true{\textrm{true}}
\def\emp{\textrm{emp}}

\title{A Theory of Statistical Inference for Ensuring\\ the Robustness of Scientific Results}

\author{Beau Coker\footnotemark \\
	Department of Biostatistics \\
	Harvard University\\
	\texttt{beaucoker@g.harvard.edu} \\
	\And
	{Cynthia Rudin\footnotemark[\value{footnote}]} \\
	Department of Computer Science \\
	Duke University\\
	\texttt{cynthia@cs.duke.edu} \\
	\And
	{Gary King} \\
	Institute for Quantitative Social Science \\
	Harvard University\\
	\texttt{king@harvard.edu} \\
}

\date{}

\renewcommand{\headeright}{Preprint}
\renewcommand{\undertitle}{Preprint}
\renewcommand{\shorttitle}{Hacking Intervals}

\hypersetup{
pdftitle={hackingintervals},
pdfsubject={},
pdfauthor={Beau~Coker, Cynthia~Rudin, Gary~King},
pdfkeywords={Robustness, Replicability, Observational Data, Model Dependence, Causal Inference, Matching},
}

\begin{document}
\maketitle

\renewcommand{\thefootnote}{\fnsymbol{footnote}}

\footnotetext[1]{Equal contribution.}
\renewcommand{\thefootnote}{\arabic{footnote}}
\setcounter{footnote}{0} 

\begin{abstract}
	Inference is the process of using facts we know to learn about facts we do not know. A theory of inference gives assumptions necessary to get from the former to the latter, along with a definition for and summary of the resulting uncertainty. Any one theory of inference is neither right nor wrong, but merely an axiom that may or may not be useful. Each of the many diverse theories of inference can be valuable for certain applications. However, no existing theory of inference addresses the tendency to choose, from the range of plausible data analysis specifications consistent with prior evidence, those that inadvertently favor one's own hypotheses. Since the biases from these choices are a growing concern across scientific fields, and in a sense the reason the scientific community was invented in the first place, we introduce a new theory of inference designed to address this critical problem. We introduce \textit{hacking intervals}, which are the range of a summary statistic one may obtain given a class of possible endogenous manipulations of the data. Hacking intervals require no appeal to hypothetical data sets drawn from imaginary superpopulations. A scientific result with a small hacking interval is more robust to researcher manipulation than one with a larger interval, and is often easier to interpret than a classical confidence interval. Some versions of hacking intervals turn out to be equivalent to classical confidence intervals, which means they may also provide a more intuitive and potentially more useful interpretation of classical confidence intervals.
\end{abstract}


\section{Introduction}

The numerous choices in even ``best practice'' data analysis procedures lead to high levels of unmeasured and unreported uncertainty in research publications. These choices include, among others, variable selection and transformations, data subsetting, identification and elimination of outliers, functional forms, distributional assumptions, priors, estimators, nonparametric preprocessing (such as matching), and procedures to control for unmeasured confounders (such as difference in differences or instrumental variables). See \citealt{Wicherts2016} for an attempt to enumerate a complete list. Classical statistical inference conditions on whichever choices the analyst makes and focuses on uncertainty induced by observing only one possible sample of data. This is uncertainty \emph{across hypothetical datasets}, where one is observed and the rest might have come from an imagined superpopulation. However, within the single observed dataset, the often considerable variation \emph{across potential ``plausible'' analysis choices} can lead to a wide range of empirical estimates, a range that is often considerably larger than the uncertainty induced by hypothetical sampling.

We thus propose that researchers (and readers) ask a simple question that gets to the heart of whether or not a quantitative conclusion can be trusted: ``Would another honest researcher, choosing different but still reasonable analysis techniques, come to a different conclusion?'' The best way to answer this question is the very process of science, where numerous researchers work in cooperation and competition in pursuit of a common goal. If one researcher publishes a result that can be questioned by another, a healthy scientific community will ensure that will happen and together with others they will be more likely to find the right answer. But what happens in the interim, when we write or read a paper today? How do we increase the likelihood that the conclusions in this paper could not be overturned by minor changes in the analysis methods that another reasonable researcher might choose? We offer a quantitative framework for answering these questions.

We use the term \emph{hacking} to describe an earnest researcher working hard to choose appropriately among many data analysis choices. Although this term is sometimes used to describe dishonest manipulation of results, we use it solely (in the positive sense of a ``hackathon'') to refer to honest scientists genuinely trying to get the right answer by making analysis choices among many reasonable alternatives. For a given model class and loss function, a \emph{hacking interval} is the smallest and largest value of a summary statistic (e.g., a coefficient in a regression, first difference, risk ratio, or other quantity of interest) that can be achieved over a set of constraints for which the researcher, readers, and the scientific community would like robustness. It quantifies the extent to which a different, also reasonable, analyst could come to different conclusions. Researchers who report hacking intervals are being more transparent about the evidence available to support their hypotheses. Hacking intervals are designed to reveal information any research publication should provide to make it less likely to mislead researchers and readers of their work. A major benefit of hacking intervals is that they are easy to understand, interpret, and teach, we think much easier than introducing hypothetical draws from imaginary superpopulations. They can be taught alongside, before, or even without reference to classical confidence intervals or any other theory of inference. They do not require knowledge of probability.

The hacking intervals we propose come in two varieties. \emph{Prescriptively constrained} hacking intervals allow for an explicit definition of the analysis choices reasonable researchers make and they identify the range of a summary statistic over these choices. They are useful when one can limit which analysis choices are valid. The second type, \emph{tethered hacking intervals}, avoid the explicit enumeration of analysis choices and require only that the predictive model chosen by the researcher has a small enough loss on the observed data. Each type of hacking interval is a consequence of the defined set of researcher constraints. In a maximum likelihood scenario, tethered hacking intervals are mathematically equivalent to profile likelihood confidence intervals (as shown in 
\textbf{Appendix C.1}). Our work therefore provides a new interpretation of profile likelihood confidence intervals that requires no understanding of probability.

Quantifying the potential impact of hacking is especially --- but not only --- important if researchers are (inadvertently) biased toward a favored hypothesis. This is crucial since standard data analysis procedures leave researchers in a situation that meets all the conditions social-psychologists have identified that lead to biased choices: In the presence of high levels of discretion, many analysis choices, little objective way to know which is best, and access to the estimates each choice results in, even honest, hard working, earnest researchers are likely to inadvertently bias results toward their favorite hypotheses \citep{Gilbert98, Banaji13, Kahneman11}. If a researcher (or reader) is concerned that analysis choices were only chosen because they yielded results consistent with the bias of the researcher, a hacking interval informs them of the degree to which this can matter. A small hacking interval says that \emph{any} researcher making choices within our defined constraints, whether biased towards a conclusion or not, could only have a small impact on the result. Hacking intervals, defined via specific norms such as the ones we suggest here, are a natural solution for conveying the impact analysis choices can have for any one publication, without the costly, time consuming, and sometimes dubious or tendentious process of ad hoc sensitivity testing designed anew for each article. Hacking intervals characterize the space of analysis choices systematically with precise computational and mathematical tools. This process can also provide insight into the state of researcher bias in an entire literature: if the hacking interval is large, and the range of conclusions from many published studies is small, then this suggests researchers may be collectively biased towards a specific conclusion.

There exist some formalized procedures that aim to mitigate the impact of bias, for example pre-registration, lists of ``best practices,'' 
enforced ignorance (e.g., double blinding experiments and journal reviews), or requiring replication datasets \citep{King95}, but the problem of reasonable researchers being able to reach a different conclusions would still exist even if researchers were each unbiased. The sheer number of possible analysis choices leaves unchecked uncertainty in scientific results unless the space of choices is rigorously defined and explored.

The prescriptive constraints and amount of loss tolerated for tethered hacking are up to the user to choose, so one could argue these choices are themselves subject to hacking. Although we argue \textit{any} choice of hacking interval constraints is better than none at all, a set of best practices can not only remove the burden of making this choice but facilitate comparison of hacking intervals across studies. Accompanying this paper we provide the \texttt{R} package \verb!hackint!, which computes constraint-based and tethered hacking intervals for linear models. Like \texttt{R}'s built-in function for standard confidence intervals \verb!confint!, \verb!hackint! requires as input only a model fitted with \verb!lm!. This linear model represents a ``base'' model in the sense that hacking is defined relative to this model. That is, the threshold for tethered hacking is a percentage of the base model's loss and the prescriptive-constraints are specified as modifications of the base model (e.g., removing a feature from the base model). The package itself is available on GitHub\footnote{Repository for \texttt{hacking} package: \url{https://github.com/beauCoker/hacking}.} and a quick demo is available in \textbf{Appendix A}. The code used to produce results in the paper is also available on GitHub.\footnote{Repository for paper results: \url{https://github.com/beauCoker/hacking_paper_results}.}

Throughout this work, we offer examples and illustrations of hacking intervals, in the context of $k$-nearest neighbors ($k$-NN), matching, variable selection, support vector machines, and, in more detail, linear regression. In Section \ref{sec:app} we present an analysis of recidivism prediction, where 
we find evidence that the COMPAS score, which is a commonly used risk-scoring system used in bail and parole decisions, is often miscalculated. This can lead in practice to high-risk criminals being released, as well as low-risk individuals being unfairly sentenced or denied bail or parole. Our evidence for this conclusion is a set of individuals for which \textit{all} reasonable models (by our definition of reasonable, and according to our dataset) from a particular model class disagree with their COMPAS score.
This is followed by 
related work and further discussion in Section \ref{sec:related_work}. All proofs are available in \textbf{Appendix D}.

\section{Theories of Inference}\label{sec:theory_of_inference}

Each of the diverse theories of inference is united by a common goal --- to understand if an observed effect is robust over counterfactual worlds imagined to have occurred. These theories can be distinguished by which set of counterfactual worlds are assumed to be of interest. For example, $p$-values consider if an effect is robust to counterfactual \emph{data} from a superpopulation. Fisher's exact $p$-values fix the data and measure if an effect is robust to counterfactual \emph{treatment assignments} from every possible randomization. Causal sensitivity analysis considers if an effect is robust to counterfactual \emph{unmeasured confounders} from a defined set \citep{Ding2016,liu}. Bayesian credible intervals define results as robust to counterfactual \emph{worlds}, generated by redrawing the data from the same data generating process, given the observed data and assumed prior and likelihood model.

In part because the sum of uncertainties from different forms of inference is usually too large to be able to conclude almost anything at all, current practice is to present, in every applied publication, intervals or another summary from \emph{only one} chosen form of uncertainty, stemming from a single theory of inference, and to temporarily assume away other forms of uncertainty. Another reason for temporarily ignoring all but one form of uncertainty is that one theory of inference may seem to be of more use than another depending on context. For example, despite studies showing a strong correlation between smoking and lung cancer, the question of whether or not smoking caused lung cancer was unsettled in the 1950s because of the possibility of an unmeasured confounding genetic variable. The Cornfield Conditions assumed that the causal effect was zero and deduced properties of the unmeasured confounding genetic variable, properties that were deemed biologically infeasible \citep{cornfield1959}. This approach to inference was vital to taking the scientific community from facts that were known (smoking correlates with lung cancer and there is an approximate biological limit on how much a genetic variable and smoking could be related) to a fact that was unknown (smoking causes lung cancer). Many other sources of uncertainty also afflicted this inference, but confounding bias was the largest perceived threat to validity, and so it was well worth it for researchers to at least temporarily set aside other sources of uncertainty.

We introduce our hacking theory of inference to address the growing crisis in science across fields, based on the mistrust of published scientific results due to high degrees of researcher discretion. As such, our theory of inference considers if a substantive result is robust to counterfactual \emph{researchers} making counterfactual \emph{analysis choices} from a defined set larger than any one researcher would normally consider. We try to define this set of analytical decisions based on what all reasonable researchers from the entire scientific community might choose. Results from our theory of inference, like all others, is based on a set of counterfactual worlds, but it is designed precisely to respond to the current concern in the community.

We hypothesize which analysis choices reasonable researchers might make, either by explicitly constraining their choices or by allowing a tolerance in the loss function. From this, we then deduce the range of effects --- the hacking interval --- of results that would have been found within these constraints. A hacking interval can therefore be used to judge whether or not the observed effect is robust to researcher choices. While a hacking interval is designed to estimate the range of conclusions that \emph{reasonable} researchers could report, \emph{any} researcher acting within the constraints will report results within the hacking interval. Because hacking intervals are designed to characterize conservatively all reasonable researcher choices, any researcher should report almost the same hacking interval.

An alternative to our approach is a greatly expanded Bayesian model (perhaps via robust Bayes combined with Bayesian model averaging) that formally specifies all possible modeling decisions, enables a choice of priors or classes of priors and the many associated hyperprior values over this large set, and computes classes of posteriors as a result. We do not recommend this approach because it adds numerous researcher choices for which prior information is rarely available, and thus may exacerbate the very problem of hacking we seek to address. Our preferred theory of inference explicitly gives up the goal of full posterior distributions or classes of posterior distributions. In their place, it seeks the more limited goal of an interval as a summary of uncertainty. What we get in return for limiting our goal to intervals is clearer ways of specifying assumptions, more effective ways of limiting researcher discretion, and easy-to-interpret results.

Hacking intervals, classical frequentist confidence intervals, Bayesian credible intervals, and others each convey important but different components of the strength of evidence in the observed data. However, hacking intervals may offer an especially natural starting point in analysis and in teaching. When researchers calculate numerical results of scientific interest, they need to quantify how strongly the observed data supports their result. As with $p$-values, classical confidence intervals quantify the robustness of the result to sampling variability. If the result could be reversed under different datasets that are likely to have occurred under a specific sampling scheme, the result is not robust.
Similarly, if a result could be reversed under different but also reasonable analysis choices, then the result is not robust. A large interval of either type should be regarded as lack of robustness of a type. However, hacking intervals may be a more natural starting point. Compared to classic confidence intervals, hacking intervals:
\begin{enumerate}
	\item \emph{represent uncertainty that always exists,}
	\item \emph{are easier to understand and explain,}
	\item \emph{are natural even when the superpopulation imagined in classical inference is not,}
	\item \emph{are often wider than classical confidence intervals}.
\end{enumerate}
On the second point, hacking intervals are the solutions to an optimization problem that requires no understanding of probability. In contrast, despite repeated clarifications of their interpretation \citep{Wasserstein2016}, frequentist confidence intervals are routinely misinterpreted and mis-explained, to the point where they have even been banned in some circles \citep{Trafimow2015} (see Section \ref{sec:related_work}). 

In regards to the third point in the above list, consider problems from the political science fields of comparative politics and international relations, where country level or time-series cross-sectional data are available. The cause of (for instance) civil wars is deeply important for understanding the past, and we may like to determine patterns that characterize political situations that have led to civil wars. One might hypothesize that countries with many people in poverty, having many young men, with neighboring countries in civil war, and with no strong government could be prone to have civil wars. The data are observational; randomization is impossible for events that happened in the past; and no more relevant data may ever be collected (at least until more civil wars of the same type occur). In situations like these, researchers often use some type of regression to estimate causal relationships. If the researcher learns that a variable has a large coefficient in the regression for predicting aspects relating to a civil war, then she may use confidence intervals to determine whether this result is robust --- robust across possible model specifications. She may use traditional inference notation (confidence intervals, null hypotheses), but since the idea of a superpopulation may not even make sense, the null hypothesis does not exist, and she may find it more natural to compute a hacking interval. Researchers in this field are not interested in constructing an imaginary superpopulation of world systems with different countries; we really only care about the actual countries and their real civil wars. The question of interest, which hacking intervals address, is whether the researcher can claim a robust empirical relationship, or whether she demonstrated only that it was merely possible to find one of a million model specifications that was consistent with her causal hypothesis. In this case, the researcher may wish to focus on the uncertainty in a hacking interval, rather than a classic confidence interval. However, to do this requires a specific mathematical framework for this interval, a subject to which we now turn.

Given these four relative advantages of hacking intervals, and that the analyst simply wants to find patterns in the data that are robust, we recommend that researchers calculate a hacking interval first and then decide if calculating a classical interval adds value.

\section{Prescriptively-Constrained Hacking Intervals}\label{sec:prescriptive_hacking}

Denote $\X\in\mathcal{X}\subset \R^{n\times p}$ as covariates, $\Y\in\mathcal{Y} \subset \R^{n\times}$ as outcomes, $\Z \in \mathcal{Z} = \mathcal{X} \times \mathcal{Y}$ as datasets, and $f:\mathcal{X}\to\mathcal{Y}$ as prediction functions from a class $\mathcal{F}_\psi$, where $\psi\in\Psi$ denotes a vector of hyperparameters. For example, $\mathcal{F}_\psi$ could be the space of all binary decision trees of maximum depth $\psi$. Let $L:\mathcal{Z}\times\mathcal{F}_\psi\times\Psi \to\mathbb{R}$ be a loss or regularized loss function and $t:\mathcal{Z}\times \mathcal{X}\times\mathcal{F}_\psi \to \mathbb{R}$ be a summary statistic of interest. The loss function may or may not depend on the hyperparameters $\psi$, so if not we omit writing $\psi$. Similarly, while the summary statistic must depend on $f$, it may or may not depend on $\Z$, which is the observed training data, or $\Xnew$, which are covariates for observations the model is not trained on. Depending on the context we may omit writing $\Z$ and/or $\Xnew$ in the definition of $t$. For hyperparameters $\psi$, training data $\Z\in\mathcal{Z}$, and, optionally, test data $\Xnew$, we assume the user finds $f^*$ that minimizes the loss $L(\Z,f^*,\psi)$ and then computes the summary statistic $t^* := t(\Z,\X^{(\text{new})}, f^*)$ based on this result. For instance, in linear regression, the user finds the linear function $f^*(\x)=\x^T\bbeta^{*}$ that minimizes the quadratic loss on the dataset $\Z=[\X,\y]$. Possible summary statistics include an estimate of a single regression coefficient $t(f^*)=\beta^*_{j}$, a goodness-of-fit measurement of $f^*$ on $\Z$, or a prediction $t(\xnew,f^*)=f^*\left(\xnew\right) = \x^{(\textrm{new})T}\bbeta^*$ on a single test observation $\xnew \in \mathcal{X}$. Our interest is in the range of summary statistics $t^*$ that could be achieved if the researcher were permitted to adjust the dataset $\Z$ and hyperparameters $\psi$. 

The approach to this problem is to explicitly constrain data adjustments $\phi: \mathcal{Z}\to\mathcal{Z}$ to a set $\Phi$ and hyperparameters to a set $\Psi$. We assume that $\phi$ can be separated into two functions $\phi_\X$ and $\phi_\Y$ such that for any $\Z=[\X,\Y]$ we have $\phi(\Z) = [\phi_\X(\X), \phi_\Y(\Y)]$. We then wish to calculate the minimum and maximum summary statistics over these two sets, $\Psi$ and $\Phi$, which constrain researcher choices:
\begin{align}
&a_{\min}:=\min_{\psi \in \Psi, \phi \in \Phi}  t\left(\phi(\Z), \phi_\X(\Xnew), \underset{f\in\mathcal{F}_\psi}{\text{argmin}}\ L\left(\phi(\Z),f,\psi\right)\right)\label{eq:cons_min},
\\
&a_{\max}:=\max_{\psi \in \Psi, \phi \in \Phi}  t\left(\phi(\Z), \phi_\X(\Xnew), \underset{f\in\mathcal{F}_\psi}{\text{argmin}}\ L\left(\phi(\Z),f,\psi\right)\right)\label{eq:cons_max}.
\end{align}
Notice hyperparameters $\psi$ impact $a_{\min}$ and $a_{\max}$ through $\mathcal{F}_\psi$ (e.g., by controlling the max depth of decision trees) as well as through the loss directly (e.g., by controlling the regularization). In other words, $\psi$ is assumed to contain all relevant hyperparameters, both to determine hard constraints on the function class, as well as soft constraints through regularization. We define the interval $[a_{\min},a_{\max}]$ as the \textit{prescriptively-constrained hacking interval}. For example, if the summary statistic $t$ is a prediction of $f$ on a new point $\xnew$, then Equations (\ref{eq:cons_min}) and (\ref{eq:cons_max}) can be written as:
\begin{align*}
a_{\min} &= \min_{\psi\in\Psi, \phi\in\Phi} \quad f\left(\phi_\X\left(\xnew\right)\right) 
\quad \text{s.t.} \quad 
f\in \underset{\mathcal{F}_\psi}{\text{argmin}}\  L(\phi(Z), f, \psi) \\
a_{\max} &= \max_{\psi\in\Psi, \phi\in\Phi} \quad f\left(\phi_\X\left(\xnew\right)\right) 
\quad \text{s.t.} \quad 
f\in \underset{\mathcal{F}_\psi}{\text{argmin}}\  L(\phi(Z), f, \psi).
\end{align*}

While a prescriptively-constrained hacking interval is designed for a single loss function, one could include in $\psi$ a hyperparameter that switches between more than one loss function, allowing for specification of the loss function to be among the researcher choices. 

Instead of using their own discretion, a researcher may pick all or some of the hyperparameters based on cross validation. To compute the prescriptively-constrained hacking interval in this case, the objective function $t$ in Equations (\ref{eq:cons_min})  and (\ref{eq:cons_max}) is evaluated by first computing the optimal hyperparameters based on cross validation (which will depend on the data-adjustment function $\phi$ and the remaining, non-cross validated hyperparameters, if any) and then plugging them into $t$.

\subsection{Examples}
We present examples of prescriptively-constrained hacking intervals for, first, the simple example of k-nearest neighbors (where the researcher chooses $k$ within a reasonable range) and then the more complex example of adding a new feature (where the researcher adds a new feature constrained by its relationship to existing data). Using results from \citet{Noor1} we also show the example of matching for causal inference (where the researcher chooses a matching algorithm) in \textbf{Appendix B}.

\subsubsection{$k$-NN}
This is a simple example. Suppose we have observed data $\Z=\{\bm{x}_i,y_i\}_{i=1}^n$ and we wish to predict on a new point $\xnew$ by averaging nearby observations.
In this example we will keep the data $\Z$ fixed but allow the researcher to choose the hyperparameter $k$, the number of nearest neighbors over which to average.
To construct a simple prescriptively-constrained hacking interval, we define a subset of reasonable hyperparameter choices $\Psi$, which in this case we can write as a range $[k_{\min},k_{\max}]$, and find the range of predictions on a new point $\xnew$ subject to the constraint that $k\in[k_{\min},k_{\max}]$:
$$
\underset{k\in[k_{\min},k_{\max}]}{\text{max/min}} \frac{1}{k} \sum_j \eta^{(k)}_{i^{(\text{new})}j}y_j
$$
where $\eta^{(k)}_{ij}$ is an indicator that is one if $\bm{x}_j$ is within the $k$ nearest neighbors of $\bm{x}_i$ and zero otherwise. This range of predictions is the prescriptively-constrained hacking interval. Notice that there is no loss function. The hyperparameter $k$ allows for only one function in the function space $\mathcal{F}_k$ (namely, the one that averages over the $k$ nearest neighbors).
To solve this problem, we evaluate the nearest neighbor average for each $k$ within the range $\Psi = [k_{\min},k_{\max}]$.

Prescriptively-constrained hacking intervals require that the researcher justify to readers their choice of $\Psi$, and we recommend that this discussion be briefly included in every paper. This approach therefore does not remove all research discretion, and arguably not all hacking, but it changes the nature of scholarly papers from a justification of a single specification to one where they justify a definition for the range of reasonable specifications.

\begin{figure}[!h]
	\centering
	\includegraphics[scale=0.3]{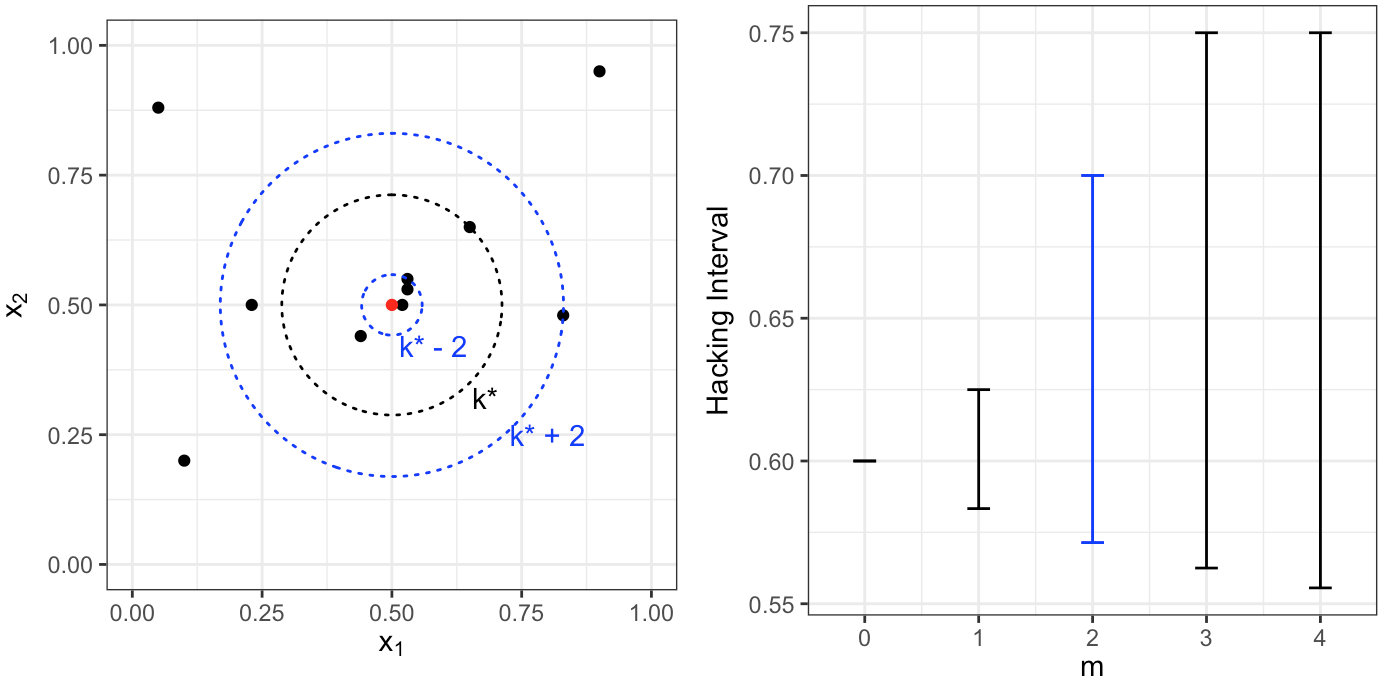}
	\caption{\textit{Left:} Observed data with distance to $k=3,5,\text{ and }7$ nearest neighbors highlighted, where $k^*=5$. \textit{Right:} Hacking intervals as a function of the hyperparameters space width $m$. $m=2$ corresponds to a hacking interval over researcher choice $[k^*-2,k^*+2]=[3,7]$.}
	\label{fig:$k$-NN}
\end{figure}

One possibility for this choice is to center $\Psi = [k_{\min},k_{\max}]$ around a fixed value and calculate the hacking interval over $[k_{\min},k_{\max}]$ constraints of increasing width. 
For example, find $k^*\in[1,n-1]$ that minimizes the training error and then find the hacking interval over $\Psi(m) := [k^* - m, k^* + m] \cap [1,n-1]$ for each $m=1,2,3,\dotsc,m_{\max}$.
Figure~\ref{fig:$k$-NN} shows the results of such a procedure for a dataset in two dimensions and $\xnew=(0.5,0.5)$. We find that $k^*=5$ minimizes the training error and the resulting prediction on $\xnew$ is $0.6$. However, if the researcher is allowed to pick any $k$ in $[k^*-2,k^*+2]=[3,7]$, for example, then the prediction ranges from $.57$ to $.70$. This is the hacking interval for $m=2$. Displaying the hacking interval as a function of $m$ illustrates the sensitivity of the hacking interval to the freedom given to the researcher.

Another choice for the range of $k$ could be to use prior information of acceptable past researcher choices. We might choose the range of $k$ large enough to include the smallest and largest $k$'s used in $k$-nearest neighbor in any article in the last 5 years in that field. In practice, that interval may actually be the smallest and largest values that would not be objected to by reviewers.

Other researcher choices for $k$-NN that we did not consider in this example could include the distance function or the weighting of the $k$ nearest neighbors. The use of $k$-NN as the predictive function class could also be considered a researcher degree of freedom. We could use a binary hyperparameter to switch between $k$-NN and any other regression algorithm. 


\subsubsection{Adding a New Feature}\label{sec:ex_feat} 

The addition of a new feature to a given collection of features, possibly from existing features (such as an interaction term) or from new data, is a data adjustment that can impact the conclusions about prediction models. A prescriptively-constrained hacking interval in this context is the range of a summary statistic that can be achieved over all of the possible choices made by the researcher about new features, subject to explicit constraints on those choices. If the researcher is given the freedom to choose each of $n$ feature values (one for each observation), then solving this problem requires optimizing over a potentially large space, since there are $n$ choices made by the researcher. Fortunately, it may only be necessary to specify a small number of attributes about the new feature to calculate its impact on the summary statistic. The prescriptively-constrained hacking interval would then be an optimization problem over a smaller space of attributes, subject to explicit constraints on those attributes. 

In a causal inference setting, where the researcher observes a treatment feature among other possibly confounding features, sensitivity analysis deals with this exact problem. The goal is to find the impact on a causal effect (the summary statistic) of an unmeasured confounder $u$ (the new feature).\footnote{Alternatively, the goal may be to assume the unmeasured confounder reduced the causal effect to zero and see what this would imply about the unmeasured confounder.} To do this one needs to choose a value for several attributes about the unmeasured confounder. There are a number of approaches to this problem that require different attributes of $u$ to be chosen \citep[see][for a review]{liu}, but generally only a few attributes are required: its distribution, its relationship to the outcome, and its relationship to the treatment. In applications of causal sensitivity analysis, a researcher will often display the adjusted causal effect for each of a few choices of these attributes. If we explicitly define a range of choices for each attribute, then the maximum and minimum causal effect over these ranges is a prescriptively-constrained hacking interval.

The motivation of causal sensitivity analysis and prescriptively-constrained hacking are different. In causal sensitivity analysis, $u$ exists but is unmeasured by the researcher. Constraints on the values of the attributes of $u$ are based on what we believe is scientifically reasonable. In prescriptively-constrained hacking, $u$ is created by the researcher. Constraints on the values of the attributes of $u$ are based on what we believe is a reasonable amount of researcher freedom.

We now define our approach in more detail. Let 
$\Y=(y_1,\dotsc,y_n)^T \in \{-1,1\}^n$ be a $n\times 1$ matrix of observed binary outcomes, $\X=(\x_i^T,\dotsc,\x_n^T)^T$ an $n\times p$ matrix of observed covariates, and 
$\W=(w_1,\dotsc,w_n)^T \in \{0,1\}^n$ an $n\times 1$ matrix of observed binary covariates. In a causal inference setting, $\W$ is the treatment. 
The researcher degrees of freedom constitute the choice of an additional binary covariate $\U=(u_1,\dotsc,u_n)^T \in \{0,1\}^n$. This is equivalent to the choice of a data-adjustment function $\phi:[\Y,\X,\W] \mapsto [\Y,\X,\W,\U]$. Once $\phi$ has been chosen, we assume the researcher finds a model $f$ from a set of linear functions $\mathcal{F}$ of the form
$f([\x,w,u]) = \beta_0 + \x^T \bbeta_\x +\beta_w w + \beta_u u$ 
by minimizing the logistic loss function: 
$$
L([\X,\W,\U],f)=\sum_{i=1}^n \log(1+e^{-y_if([\x_i,w_i,u_i])}).
$$ 
Notice this is equivalent to maximizing the likelihood under the model: $\text{logit}\ \text{Pr}(Y=1 \mid \x, w, u) = f([\x,w,u])$, where $Y$ is the random variable corresponding to the observed $y$. In other words, the researcher performs logistic regression. (For simplicity, the objective is fixed and there are no user choices except to add the extra feature.)
We further assume the researcher is interested in the odds ratio of $y$ and $w$ controlling for covariates $\x$ and $u$:
$$
OR_{y w\mid \x, u} := \frac{\text{Pr}(Y=1 \mid \x,w=1,u)}{\text{Pr}(Y=1 \mid \x,w=0,u)},
$$
so we set the test statistic to be $t(f):=e^{\beta_w}$. The steps followed by the researcher can be summarized as follows:
\begin{itemize}
	\item \textit{1a}: Choose a data adjustment $\phi\in\Phi$ (we discuss $\Phi$ below).
	\item \textit{1b}: Find $\hat{f}([\x,w,u])=\hat{\beta}_0 + \x^T \hat{\bbeta}_\x +\hat{\beta}_w w + \hat{\beta}_u u $ that minimizes the logistic loss on the adjusted data, $(\Y,\X,\W,\U) = \phi(\Y,\X,\W)$.
	\item \textit{1c}: Calculate the summary statistic $\widehat{OR}_{y w\mid \x, u} = t(\hat{f}) =e^{\hat{\beta}_w}$.
\end{itemize}
The prescriptively-constrained hacking interval is the maximum and minimum values of $t(\hat{f})$ that can be achieved over all of the possible researcher choices of $\phi\in\Phi$. There are no hyperparameters in this example.

Interestingly, we can calculate $\widehat{OR}_{y w\mid \x, u}$ without knowing the researcher-created covariate $u$ exactly. We need only know
the relationship of $u$ to the binary covariate $w$, specified by $p_0 := p(U \mid w=0)$ and $p_1:=\text{Pr}(U\mid w=1)$ (where $U$ is the random variable corresponding to $u$), 
and the relationship of $u$ to the binary outcome $y$, specified by $OR_{y u}:=\text{Pr}(Y=1\mid u=1)/\text{Pr}(Y=1\mid u=0)$. 
When $p_0$, $p_1$, and $OR_{y u}$ are known, \citet{lin} show\footnote{\citet{lin} show this result exactly for log-linear regression, but they argue it should hold approximately for logistic regression.}
that we can derive the odds ratio adjusting for $\x$ and $u$, $\widehat{OR}_{y w\mid \x, u}=t(\hat{f})$, from the odds ratio that only adjusts for $\x$, $\widehat{OR}_{y w\mid \x}$, by the following formula:
\begin{equation}
\widehat{OR}_{y w\mid \x, u} =  \frac{1}{AF}\widehat{OR}_{y w\mid \x}
,\  \text{where }
AF=\frac{(OR_{y u}-1) p_1 + 1}{(OR_{y u}-1) p_0 + 1}. \label{eq:ORywxu}
\end{equation}
We write $OR_{y u}$ rather than $\widehat{OR}_{y u}$ because the former quantity is the true odds ratio, not one estimated from the data.

Since $\widehat{OR}_{y w\mid \x}$ can be estimated from the observed data, Equation (\ref{eq:ORywxu})  implies the impact of the researcher choice of $u$ is completely summarized by $p_1$, $p_0$, and $OR_{y u}$, since they determine $AF$. Conversely, if we knew the data adjustment $\phi$ we could estimate $p_1$, $p_0$, and $OR_{y u}$, calling the estimates $\hat{p}_1$, $\hat{p}_0$, and $\widehat{OR}_{y u}$, respectively, from the adjusted data.
Steps $\textit{1a-1c}$ are therefore equivalent to Steps $\textit{2a-2d}$ defined by:
\begin{itemize}
	\item \textit{2a}: Calculate $\widehat{OR}_{y w\mid \x}$.
	\item \textit{2b}: Choose a data adjustment $\phi\in\Phi$.
	\item \textit{2c}: Calculate $\widehat{OR}_{y u}$, $\hat{p}_1$, and $\hat{p}_0$ using the adjusted data $[\Y,\X,\W,\U] = \phi([\Y,\X,\W])$.
	\item \textit{2d}: Calculate the summary statistic $\widehat{OR}_{y w\mid \x, u}=\frac{1}{\widehat{AF}} \widehat{OR}_{y w\mid \x}$, where $\widehat{AF}$ is analogous to Equation (\ref{eq:ORywxu}) but depends on the estimated quantities $\widehat{OR}_{y u}$, $\hat{p}_1$, and $\hat{p}_0$:
	\begin{equation}
	\widehat{AF}=\frac{(\widehat{OR}_{y u}-1) \hat{p}_1 + 1}{(\widehat{OR}_{y u}-1) \hat{p}_0 + 1}.\label{eq:AFhat}
	\end{equation}
\end{itemize}
Notice that the researcher's choice of a data adjustment $\phi$ implies a value for $u$ and the three attributes about $u$ --- $\widehat{OR}_{y u}$, $\hat{p}_1$, and $\hat{p}_0$ --- but it is through these three attributes only that $\phi$ impacts the summary statistic. If we instead allow the researcher to choose only the three attributes, we can find the impact on the summary statistic without ever knowing $u$. We just need to define the space of allowable data adjustments $\Phi$ in terms of its impact on these three attributes:
\begin{equation*}
\Phi := \left\{ \phi: (\Y,\X,\W)\mapsto(\Y,\X,\W,\U) \mid \widehat{OR}_{y u} \in [a,b], |\hat{p}_1-\hat{p}_0|\le c, \hat{p}_0 > d \right\},
\end{equation*}
for constants $a$, $b$, and $c<d$ (the reason for these exact constraints will be come clear later). Then, Steps $\textit{2a-2d}$ can be replaced with Steps $\textit{3a-3c}$ defined by:
\begin{itemize}
	\item \textit{3a}: Calculate $\widehat{OR}_{yw\mid \x}$.
	\item \textit{3b}: Choose $OR_{y u}$, $p_0$, and $p_1$ such that $OR_{y u} \in [a,b]$, $|p_1-p_0|\le c$, and $p_0 \ge d$.
	\item \textit{3c}: Calculate the summary statistic $\widehat{OR}_{y w\mid \x, u}=\frac{1}{AF}\widehat{OR}_{y w\mid \x}$, where $AF$ depends on $OR_{y u}$, $p_0$, and $p_1$.
\end{itemize}

For any equivalent choice of constraints, the maximum and minimum values of $\widehat{OR}_{y w\mid \x, u}$ that could be achieved by any of the three sequences of Steps (\textit{1a-1c}, \textit{2a-2d}, and \textit{3a-3c}) are all equal. We can think of finding the maximum and minimum values of $\widehat{OR}_{y w\mid \x,u}$ for each of the three sequences as the three following optimization problems (each solved for the maximum and minimum):
\begin{align}
\text{Steps 1a-1c:}&\quad
\underset{\phi\in\Phi}{\text{max/min}}\  \left\{OR_{yw\mid \x, u}\right\}
\\
\text{Steps 2a-2d:}&\quad
\underset{\phi\ \text{s.t.} 
	\begin{cases}
	{\scriptstyle \widehat{OR}_{yu} \in [a,b] } \\[-.6em]
	{\scriptstyle |\hat{p}_1-\hat{p}_0|\le c} \\[-.6em]
	{\scriptstyle \hat{p}_0 > d }
	\end{cases}}
{\text{max/min}}\  \left\{\frac{1}{\widehat{AF}}\widehat{OR}_{yw\mid \x}\right\} 
\\
\label{eq:3a3c}
\text{Steps 3a-3c:}&\quad
\underset{ 
	\begin{aligned}
	& {\scriptstyle OR_{yu} \in [a,b]} \\[-.4em]
	& {\scriptstyle |p_1-p_0|\le c } \\[-.4em]
	& {\scriptstyle p_0 \ge d }
	\end{aligned}}
{\text{max/min}}\  \left\{\frac{1}{AF}\widehat{OR}_{yw\mid \x}\right\}.
\end{align}
Optimization Problem (\ref{eq:3a3c}) will prove the most useful as it does not require knowledge of $u$. Since $\widehat{OR}_{yw\mid \x}$ is estimated from the observed data, solving Optimization Problem  (\ref{eq:3a3c}) is equivalent to solving for the maximum and minimum values of $AF$ subject to the same constraints and dividing $\widehat{OR}_{yw\mid \x}$ by each value. Using Equation (\ref{eq:ORywxu}) for $AF$, we find the maximum and minimum values of $AF$ by solving the following optimization problem:
\begin{equation}
\underset{OR_{y w},p_1,p_0}
{\text{max/min}} \quad 
\frac{(OR_{yu}-1) p_1 + 1}{(OR_{yu}-1) p_0 + 1}
\quad\text{s.t.}\quad
\begin{cases}
OR_{yu} \in [a,b] \\
|p_1-p_0|\le c\\
p_0 \ge d
\end{cases}.\label{eq:feat_hack}
\end{equation}
Dividing $\widehat{OR}_{yw\mid \x}$ by the maximum and minimum values given by Optimization Problem (\ref{eq:feat_hack}) gives the minimum and maximum values, respectively, of $OR_{yw\mid \x, u}$, which define the hacking interval in this case.

We can solve Equation (\ref{eq:feat_hack}) for the case where $OR_{yu}$ is fixed greater than one (implying $\text{Pr}(Y=1\mid u=1)>\text{Pr}(Y=1\mid u=0)$). In this case, the maximization problem in Equation (\ref{eq:feat_hack}) (i.e., the hacking interval upper bound) becomes:
\begin{align*}
\underset{ 
	\begin{aligned}
	&{\scriptstyle |p_1-p_0|\le c} \\[-1.25em]
	&{\scriptstyle p_0 \ge d} 
	\end{aligned}}
{\text{max}}\  \frac{(OR_{yu}-1) p_1 + 1}{(OR_{yu}-1) p_0 + 1}
=
\max_{p_0 \ge d}
\  \frac{(OR_{yu}-1) (p_0+c) + 1}{(OR_{yu}-1) p_0 + 1} 
=
\max_{p_0 \ge d}
\  1+\frac{(OR_{yu}-1)c}{(OR_{yu}-1) p_0 + 1},
\end{align*}
while the minimization problem (i.e., the hacking interval lower bound) becomes:
\begin{align*}
\underset{ 
	\begin{aligned}
	&{\scriptstyle |p_1-p_0|\le c} \\[-1.25em]
	&{\scriptstyle p_0 \ge d} 
	\end{aligned}}
{\text{min}}\  \frac{(OR_{yu}-1) p_1 + 1}{(OR_{yu}-1) p_0 + 1}
=
\min_{p_0 \ge d}
\  \frac{(OR_{yu}-1) (p_0-c) + 1}{(OR_{yu}-1) p_0 + 1} 
=
\min_{p_0 \ge d}
\  1-\frac{(OR_{yu}-1)c}{(OR_{yu}-1) p_0 + 1}.
\end{align*}
In each case, the optimum occurs at $p_0 = d$. Therefore, Equation (\ref{eq:feat_hack}) can be solved when $OR_{y u}$ is fixed greater than one. We apply this result in Section \ref{sec:recid_feat}.

This section shows how results from causal sensitivity analysis can be leveraged to solve problems where the researcher is permitted to hack a new feature. Here, we have been in a non-causal inference setting of logistic regression modeling. In Section \ref{sec:recid_feat} we apply these results to a recidivism dataset.

\section{Tethered Hacking Intervals}\label{sec:tethered_hacking}

In prescriptively-constrained hacking intervals, discussed in Section \ref{sec:prescriptive_hacking}, we optimize over a data-adjustment function $\phi$ and hyperparameters $\psi$ constrained to be in sets $\Phi$ and $\Psi$, respectively. An advantage of this approach is that we can clearly define acceptable researcher adjustments. A disadvantage is that the possible adjustments may be difficult to enumerate or optimize over efficiently. One way to circumvent this requirement is to allow \textit{any} choice of $\psi$ and $\phi$ so long as the loss using the unadjusted data $\Z$ and a set of default hyperparameters $\psi_d$ is not too large. The \textit{tethered hacking interval} is the minimum and maximum summary statistic under this constraint. In other words, it is given by the interval $[b_{\min},b_{\max}]$, 
\begin{align} 
&b_{\min}:=\min_{f\in\mathcal{F}_{\psi_d}}  t\left(\Z, \Xnew, f\right) \quad \text{s.t.} \quad L\left(\Z,f,\psi_d\right) \le \theta, \label{eq:tethering_min}
\\
&b_{\max}:=\max_{f\in\mathcal{F}_{\psi_d}}   t\left(\Z, \Xnew, f\right) \quad \text{s.t.} \quad L\left(\Z,f,\psi_d\right) \le \theta, \label{eq:tethering_max}
\end{align}
given a fixed, chosen value of $\theta$. 
The default hyperparameters $\psi_d$ could be specified based solely on problem-specific standards, or on cross validation, or a combination of both. To do the combination in the case that there are multiple viable values of the problem-specific hyperparameters, we would first choose values for the problem-specific hyperparameters and perform cross validation on the rest, repeating this procedure for every viable value of the problem-specific hyperparameters. Once a single set of hyperparameters $\psi_d$ is selected we can proceed with computing the tethered hacking interval by Equations (\ref{eq:tethering_min}) and (\ref{eq:tethering_max}). In contrast to the computation of prescriptively-constrained hacking intervals in the case of cross validation, as described in Section \ref{sec:prescriptive_hacking}, any cross validation of the hyperparameters is done prior to solving the optimization problems. 

For example, suppose $\mathcal{F}$ is the set of constant functions $f(x) = \lambda$, $t(\Z,\Xnew,f)=\lambda$ is the parameter $\lambda$ that defines $f$, and $L$ is the quadratic loss for each of $n$ observations in dataset $\Z$. There are no hyperparameters $\psi_d$ so we suppress their notation. Then Equations (\ref{eq:tethering_min}) and (\ref{eq:tethering_max}) become:
\begin{align*}
b_{\min} &= \min_{\lambda}  \quad \lambda \quad \text{s.t.} \quad \sum_{i=1}^n \left(\lambda - y_i\right)^2 \le \theta \\
b_{\max} &= \max_{\lambda}  \quad \lambda \quad \text{s.t.} \quad \sum_{i=1}^n \left(\lambda - y_i\right)^2 \le \theta.
\end{align*}
For another example, if $\mathcal{F}$ is the set of linear functions $f(x) = \lambda_0+\lambda_1 x$, $t(Z,f)=\lambda_0 + \lambda_1 \xnew$ is a prediction of $f$ on a new point $\xnew$, and $L$ is the same quadratic loss, then Equations (\ref{eq:tethering_min}) and (\ref{eq:tethering_max}) become:
\begin{align*}
b_{\min} &= \min_{\lambda_0, \lambda_1} \quad \lambda_0 + \lambda_1 \xnew \quad \text{s.t.} \quad \sum_{i=1}^n \left(\lambda_0 + \lambda_1 x_i - y_i\right)^2 \le \theta \\
b_{\max} &= \max_{\lambda_0, \lambda_1} \quad \lambda_0 + \lambda_1 \xnew \quad \text{s.t.} \quad \sum_{i=1}^n \left(\lambda_0 + \lambda_1 x_i - y_i\right)^2 \le \theta.
\end{align*}
In general, when the summary statistic is a prediction on a new point $\xnew$, Equations (\ref{eq:tethering_min}) and (\ref{eq:tethering_max}) become:
\begin{align*}
b_{\min} &= \min_{f\in\mathcal{F}_{\psi_d}} \quad f\left(\xnew\right) \quad \text{s.t.} \quad L(\Z,f,\psi_d) \le \theta \\
b_{\max} &= \max_{f\in\mathcal{F}_{\psi_d}} \quad f\left(\xnew\right) \quad \text{s.t.} \quad L(\Z,f,\psi_d) \le \theta.
\end{align*}

The interpretation of a tethered hacking interval is that a researcher could have hacked the data or adjusted the hyperparameters to obtain values of the test statistic in the interval. In other words, for each point 
$b' \in [b_{\min}, b_{\max}]$
there could exist a data-adjustment function $\phi'$ and a set of hyperparameters $\psi'$ such that $b'$ is the output of the summary statistic when applied to the minimum loss predictive model $f$ using $\phi'$ and $\psi'$. That is, 
$$
b'=t\left(\phi'(\Z), \phi_\X'(\Xnew),\underset{f\in\mathcal{F}_{\psi'}}{\text{argmin}}\ L\left(\phi'(\Z),f,\psi'\right)\right).
$$
This interpretation describes how results are hacked in practice. First, a researcher chooses how to adjust a dataset and which hyperparameters are appropriate, and then summarizes the resulting best function in a class. The purpose of a tethered hacking interval is to bound the results of this procedure by specifying a single constraint on the loss function.


The set of models achieving small loss is also called the \textit{Rashomon Set} \citep{fisher}, based on terminology originally due to Leo Breiman's analogy to the 1950 Akira Kurosawa film \textit{Rashomon} \citep{breiman_2001}.  \citet{fisher} introduce a measure of variable importance for a class of prediction functions based on the Rashomon set. While the computation and interpretation of their ``empirical model class reliance'' measure of variable importance could be viewed as similar to those of hacking intervals, their ultimate goal is to study the population version of this quantity, in order to study the Rashomon set for the population.

We note two things about tethered hacking intervals. First, when the loss function corresponds to a likelihood function, tethered hacking intervals are equivalent to profile likelihood confidence intervals for an appropriate choice of the loss threshold $\theta$. See \textbf{Appendix C.1} 
for details. Second, as with prescriptively-constrained hacking intervals, a tethered hacking interval is a statement about the degree to which summaries of a single observed dataset could be hacked by a researcher. It does not require an assumption about a true data generating procedure. If we make such an assumption about the true data generating procedure, we can derive an appropriate generalization bound in order to unite traditional inference with our new inference paradigm. See 
\textbf{Appendix C.2} 
for details.

Next we discuss tethered hacking intervals for predictions made by SVM. The examples of predictions made by kernel regression and features selected using PCA can be found in \textbf{Appendix C.3} 
and 
\textbf{Appendix C.4}, respectively.

\subsection{Example: SVM}\label{sec:ex_svm}

\begin{figure}[!h]
	\centering
	\includegraphics[scale=.6]{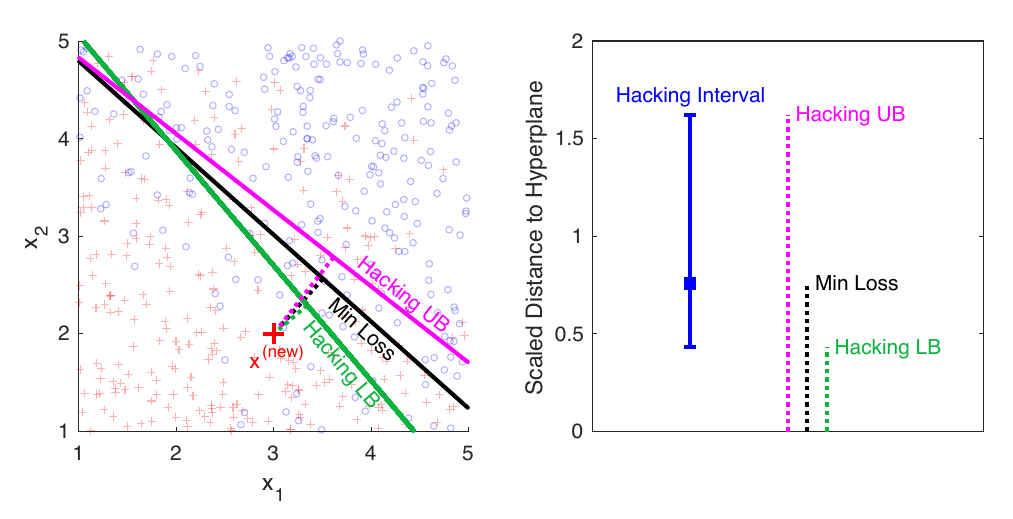}
	\caption{Lower bound (LB) and upper bound (UB) of the hacking interval for an SVM prediction. The summary statistic being hacked is the distance from the separating hyperplane to a new observation, $\xnew$. For a default regularization tradeoff of $\psi_d=1$ and a 5\% tolerance on the loss relative to the minimum loss solution, SVM will always predict a $+1$ label, but the scaled distance to the hyperplane, $\blambda^T \xnew + \lambda_0$, can range from about 0.4 to about 1.6.}
	\label{fig:svm}
\end{figure}

In this section we demonstrate how hacking intervals can be calculated in the context of support vector machines (SVM) with a linear kernel. Recall that SVM is trained by minimizing the following loss function:
\begin{equation*}
L(\Z, f, \psi_d) = \frac{1}{2}\norm{\blambda}_2^2 + \psi_d\sum_{i=1}^n (1-y_i f(\bm{x}_i))_+,
\end{equation*}
where $f(\bm{x}) := \blambda^T \bm{x}  + \lambda_0$ is the scaled distance of $\bm{x} $ to the separating hyperplane and $\psi_d\in\mathbb{R}^+$ is a hyperparameter that controls the degree of regularization. Here, we define the summary statistic as the distance of a new point $\xnew$ to the separating hyperplane. The hacking interval is then given by:
\begin{equation}
\underset{\blambda, \lambda_0}{\text{max/min}}\  \blambda^T \xnew + \lambda_0
\quad \text{s.t.} \quad
\frac{1}{2}\norm{\blambda}_2^2 + \psi_d\sum_{i=1}^n (1-y_i (\blambda^T \bm{x}_i + \lambda_0))_+ \le \theta, \label{eq:svm0}
\end{equation}
where $\theta$ controls the loss tolerance. Figure \ref{fig:svm} illustrates this problem.

For simplicity we can write both the min and max problems from Equation (\ref{eq:svm0}) as a single minimization problem that depends on the choice of a binary variable $s\in\{-1,+1\}$ ($s=1$ for min, $s=-1$ for max). If we also write the loss constraint in terms of slack variables $\bxi$ then Equation (\ref{eq:svm0}) becomes:
\begin{align} \label{eq:svm}
\min_{\blambda, \lambda_0, \bxi} s\blambda^T \xnew + s\lambda_0 \quad \text{s.t.} \quad 
\begin{cases}
y_i (\blambda^T x_i + \lambda_0) \ge 1- \xi_i,\  \forall i \\
\xi_i \ge 0, \  \forall i \\
\frac{1}{2} \norm{\blambda}_2^2 + \psi_d \sum_{i=1}^n \xi_i \le \theta.
\end{cases}
\end{align}
This is a convex optimization problem. The objective is linear. The first two constraints are the same as in non-separable SVM and are linear. The last constraint is the sum of a norm (always convex) and a linear function in $\bxi$, so it is convex; also, it is the objective function for non-separable SVM. Therefore, we can apply the KKT conditions to obtain the dual problem. The following theorem shows the result.

\begin{prop}[Hacking Intervals for SVM]
	\label{thmSVM}
	The solution to optimization problem (\ref{eq:svm}) is given by
	\begin{equation*}
	\blambda^* = \frac{1}{\beta^*} \left(-s\xnew + \sum_{i=1}^n\alpha^*_i y_i \x_i \right) \quad \text{and} \quad \lambda^*_0 = y_{i_{sv}} - \blambda^{*T} \x_{i_{sv}},
	\end{equation*}
	where $i_{sv}$ is such that $0 < \alpha^*_{i_{sv}} < \beta^*\psi_d$ and the optimal dual variables $(\bm{\alpha}^*,\beta^*)$ are the solutions to the following dual problem:	
	\begin{equation} 
	\begin{split}
	\max_{\bm{\alpha}, \beta}
	-\frac{1}{2\beta} \left[
	\xnewt \xnew 
	-2s\sum\alpha_i y_i \x_i^T \xnew
	+\sum_i \sum_k \alpha_i \alpha_k y_i y_k \x_i^T \x_k
	\right]
	+ \sum\alpha_i - \beta \theta \\
	\quad\text{s.t.}\quad
	\begin{cases}
	0\le \alpha_i \le \beta \psi_d,\  \forall i \\
	\sum_{i=1}^n \alpha_i y_i =s \\
	\beta \ge 0 
	\end{cases}.
	\end{split} \label{eq:svm_dual}
	\end{equation}
\end{prop}

In Section \ref{sec:recid_svm} we apply SVM hacking intervals to a recidivism dataset.

\section{Tethered Hacking Intervals for Linear Regression}\label{sec:lin_reg}

We develop hacking intervals in detail for two linear regression scenarios:

\begin{itemize}
	\item \textit{Scenario 1: average treatment effect (ATE)}. We assume a class of linear functions $\mathcal{F}$ with $p$ confounders and an indicator covariate for the treatment (1 if treatment, 0 if control). We write $f\in\mathcal{F}$ as:
	\[
	f(\x,\textrm{treated or control})=\beta_1 x_{.1} + \beta_2 x_{.2} + ... \beta_p x_{.p} + \beta_0 1_{\treated}.
	\]
	The goal is to construct a tethered hacking interval for $\beta_0$, the coefficient of the treatment indicator. In other words, the test statistic is $t(\Z,f)=\beta_0$. The coefficient $\beta_0$ represents the average treatment effect. Section \ref{sec:scen1} develops this in detail.
	
	\item \textit{Scenario 2: individual treatment effect (TE)}. We assume a class of linear functions $\mathcal{F}$ with $p$ confounders for both the treatment and control groups. We write $f\in\mathcal{F}$ as:
	\begin{equation*}
	f(\x,\textrm{treated or control})=1_{\textrm{control}}[\beta_1^c x_{.1} + \beta_2^c x_{.2} + ... \beta_p^c x_{.p}] + \\
	1_{\textrm{treated}}[\beta_1^t x_{.1} + \beta_2^t x_{.2} + ... \beta_p^t x_{.p}],
	\end{equation*}
	where $1_{\textrm{control}}$ is $1$ only for the control group and $1_{\textrm{treated}}$ is 1 only for the treatment group.
	The goal is to construct a tethered hacking interval for a prediction of $f$ on a new point $[\xnew, \textrm{treated or control}]$. In other words, the test statistic is $t(\Z,[\xnew,\textrm{treated or control}],f)=f(\xnew,\textrm{treated or control})$. The value $f(\xnew,\textrm{treated or control})$ represents the prediction for a person with covariates $\xnew$. Section \ref{sec:scen2} develops this in detail.
	
\end{itemize}
In both scenarios we maintain the canonical assumptions of overlap, SUTVA, and conditional ignorability, and we use a quadratic loss function $L(\Z,f) = \sum_{i=1}^n \left(y_i - f(\x_i, 1_{[i \textrm{ treated}]}) \right)^2$, where $\Z=\{[\x_i,1_{[i \textrm{ treated}]}, y_i]\}_{i=1}^n$ is the observed data. There are no hyperparameters so we suppress their notation in the loss function.

\subsection{Scenario 1: Average Treatment Effect} \label{sec:scen1}
The goal is to find the range of treatment effects, $\beta_0$, corresponding to all possible ways the analyst can hack the observed data subject to a constraint on the loss.
Thus our goal is to solve:
\begin{eqnarray}
\max_{\bbeta\in\R^p,\beta_0\in\R} \beta_0 \qquad&\textrm{ s.t. }&\qquad \sum_{i=1}^n\left(y_i-\bbeta\x_i-\beta_01_{[i \textrm{ treated}]}\right)^2 \leq \theta \quad \text{and} \label{thm1max}
\\
\min_{\bbeta\in\R^p,\beta_0\in\R} \beta_0 \qquad&\textrm{ s.t. }&\qquad \sum_{i=1}^n\left(y_i-\bbeta\x_i-\beta_01_{[i \textrm{ treated}]}\right)^2 \leq \theta.\label{thm1min}
\end{eqnarray}
This is a convex quadratically-constrained linear program. Since there are inequality constraints, we require the full KKT conditions (the method of Lagrange multipliers does not handle inequality constraints).
As it turns out, answers to these problems can be found analytically. This is one of the rare problems for which a subset of the KKT conditions can be used to find a closed form solution.  The proof is available in \textbf{Appendix D}.

\begin{theorem}[Hacking Interval for Least-Squares ATE]
	\label{TheoremGaryConfInterval}
	
	Define the following:
	\begin{itemize}
		\item $\bbeta^*_{LS} := (\X^T\X)^{-1}\X^T\mathbf{Y}$, the optimal least square solution from regressing $\Y$ on $\X$. 
		
		\item $\tilde{\bbeta}^*_{LS} := (\tilde{\X}^T\tilde{\X})^{-1}\tilde{\X}^T\mathbf{Y}$, the optimal least square solution from regressing $\Y$ on $\tilde{\X}:=[\X, \one_{[\treated]}]$. The coefficient within this vector for the treatment variable is denoted $\tilde{\beta}^*_{0,LS}$. 
		
		\item $\bm{\gamma}^*_{LS} := (\X^T\X)^{-1}\X^T\one_{[\treated]}$, the optimal least square solution from regressing $\one_{[\treated]}$ on $\X$.
		
		\item $V_{tt} := ( [\tilde{\X}^T\tilde{\X}]^{-1})_{tt}$, the diagonal entry corresponding to the treatment variable of $[\tilde{\X}^T\tilde{\X} ]^{-1}$.
		
		\item $\text{SSE} :=(\Y - \X\bbeta^*_{LS})^T (\Y - \X\bbeta^*_{LS})$, the sum of squared errors of the optimal least square solution.
	\end{itemize}
	Then, the solutions of the optimization problem \eqref{thm1max} are:
	\begin{equation}
	\beta_{0,\max}^{*}=\tilde{\beta}^*_{0,LS}+\sqrt{V_{tt}}\sqrt{\theta - \SSE},\quad
	\bbeta_{\max}^*=\bbeta^*_{LS}-\beta_{0,\max}^{*}\bm{\gamma}^*_{LS}. \label{eq:beta_max}
	\end{equation}
	and the solutions of the optimization problem \eqref{thm1min} are:
	\begin{equation}
	\beta_{0,\min}^{*}=\tilde{\beta}^*_{0,LS}-\sqrt{V_{tt}}\sqrt{\theta - \SSE},\quad 
	\bbeta_{\min}^*=\bbeta^*_{LS}-\beta_{0,\min}^{*}\bm{\gamma}^*_{LS}. \label{eq:beta_min}\
	\end{equation}
	
\end{theorem}
From this theorem, one can see that the range $\beta_{0,\max}^*-\beta_{0,\min}^*$ scales as the square root of the permitted level of optimality $\theta$. The solution is not difficult to find if the relevant KKT conditions are substituted into each other in a particular order. 

Next, we relate the new confidence intervals to the standard ones and then produce new interpretations for confidence intervals, based on in-sample error increases. In the process, we will discuss possible meanings for the user-defined parameter $\theta$.

\subsubsection{Relationship to classical confidence intervals}

We have just produced a confidence interval for $\beta_0$. How does that compare with a typical confidence interval produced using the standard approach where we assume a null distribution? The confidence interval is symmetric in both cases around the least squares solution, so we must be able to equate them. We next equate traditional confidence intervals with our confidence intervals, which relates $\alpha$ for a significance test with $\theta$ for our robust confidence interval.

\begin{theorem}[ATE Hacking Intervals and Standard Confidence Intervals]
	\label{thm_CI}
	
	Start with a standard confidence interval for $\beta_0$ under usual assumptions (normality of errors given a linear model), which is given by:
	\begin{equation}
	\left[
	\tilde{\beta}^*_{0,LS} - \tpn \sqrt{\frac{\SSE}{n-p-1}}\sqrt{V_{tt}},
	\tilde{\beta}^*_{0,LS} + \tpn \sqrt{\frac{\SSE}{n-p-1}}\sqrt{V_{tt}}
	\right]\label{eq:std_CI}
	\end{equation}
	where $\tpn$ is the $1-\alpha/2$ quantile of a $t$ distribution with $n-p-1$ degrees of freedom (we estimate $p$ coefficients plus the treatment variable). Then, in order to keep the new confidence interval from Theorem \ref{TheoremGaryConfInterval} the same as the standard one, we would take the following value for $\theta$:
	\[
	\theta = \SSE\left(1 + \frac{\tpn^2}{n-p-1}\right).
	\]
\end{theorem}
Thus, for teaching purposes, rather than explaining the $t$ distribution or the meaning of $\alpha$ to a student unfamiliar with these topics, we can explain $\theta$ first and later convert to $\alpha$ for those who want this interpretation.

\subsubsection{Non-classical Choices for $\theta$}

In classical hypothesis testing, one would choose the significance level $\alpha$  and say that if the data were drawn repeatedly from the true model, the probability that an estimated value of $\beta_0$ would be within the confidence interval with probability at least $1-\alpha$. We propose \emph{in-sample} alternatives based on the meaning of $\theta$. Here are some natural choices:
\begin{itemize}
	\item \textit{Choose $\theta$ as a percentage of the $\SSE$}: Assume the
	user would not allow a model that would achieve more than $10\%$ 
	higher error than the $\SSE$. Then we set $\theta=1.1\cdot\SSE$.
	Generally if we do not tolerate more than $r\%$ error higher than
	the $\SSE$, we would choose $\theta=(1+r) \SSE$.
	
	To use this, we would ask questions like: ``If we were to tolerate
	any type of change to the data or model that would incur an
	additional error of $10\%$, what are the largest and smallest
	treatment effect one could estimate?'' If the answer is that the
	treatment effect estimate is robust to 10\% error due to user
	hacking, then the estimate is reliable inside the hacking interval.
	
	\item \textit{Choose $\theta$ as the minimum loss suffered to allow
		the treatment effect coefficient to be 0}. Let us say without loss
	of generality that the estimated treatment effect coefficient is
	negative. Then the upper confidence interval is (using Theorem
	\ref{TheoremGaryConfInterval}):
	\begin{eqnarray*}
		\beta_{0,\max}^{*}&=&\beta^*_{0,LS}+\sqrt{V_{tt}}\sqrt{\theta - \SSE},
	\end{eqnarray*}
	We can set this value to 0, which would provide the minimum sacrifice in least square error necessary for that coefficient to become 0. We thus need to solve for $\theta_0$ in the following:
	\begin{equation*}
	0=\beta^*_{0,LS}+\sqrt{V_{tt}}\sqrt{\theta_{0} - \SSE} \iff
	\theta_{0}  = \frac{(\beta^*_{0,LS})^2}{V_{tt}}+\SSE.
	\end{equation*}
	In other words, we would need to sacrifice a least squared error of at least $\frac{(\beta^*_{0,LS})^2}{V_{tt}}$ beyond that of the optimal solution in order that the regression coefficient could be 0.
	
	To use this, we would ask questions like: ``How much loss would need to be sacrificed in order for the treatment to have the opposite estimated effect?''
	
\end{itemize}

\subsubsection{Combining with data variance}\label{sec:hacking_plus_data}

The bounds of the hacking interval, $\beta_{0,\max}^{*}$ and $\beta_{0,\min}^{*}$, are deterministic functions of a fixed dataset $[\tilde{\X},\Y]$. If we assume the outcomes $\Y$ are one possible realization of a ground truth linear process given by:
\begin{equation}
\Y \sim N(\tilde{\X}\bbeta, \sigma^2 I), \label{eq:truth}
\end{equation}
then the bounds of the hacking interval are random variables. The following theorem gives their variance. 

\begin{theorem}[Variance of Least-Squares ATE Hacking Interval Bounds]
	\label{thm_var}
	If outcomes $\Y$ are generated by Equation \ref{eq:truth} and the threshold $\theta$ is set to $(1+r)\SSE$ for any $r>0$, then the variance of both hacking interval bounds $\beta_{0,\min}^{*}$ and $\beta_{0,\max}^{*}$ given by Equations (\ref{eq:beta_min}) and (\ref{eq:beta_max}), respectively, is:
	\begin{align}
	\Var{\beta^*_{0,\min} \mid \tilde{\X}} = \Var{\beta^*_{0,\max} \mid \tilde{\X}} 
	&= \sigma^2 V_{tt}\left( 1 + r(n-p-1 - \mu^2) \right), \label{eq:var}
	\end{align}
	where 
	\begin{equation}
	\mu=\left( \frac{\sqrt{2}\Gamma((n-p)/2)}{\Gamma((n-p-1)/2)}\right). \label{eq:mu}
	\end{equation}
\end{theorem}

\subsubsection{Illustration}\label{subsecillus}

Let us consider an illustrative example. We suppose a ground truth with two covariates called $v_{\cdot 1}$ and $v_{\cdot 2}$, chosen uniformly and independently over the interval [1,5], a 1/2 probability of treatment assignment for each observation, and outcomes generated by the following process:
\begin{equation}
y_i = 2\times 1_{[\treated]} + v_{i 1} + v_{i 2} + \epsilon_i, \label{eq:true_process}
\end{equation}
where $\epsilon_i\sim N(0,1)$.
In this illustration, the researcher observes more than $v_{i 1}$, $v_{i 2}$, and the treatment indicator, $1_{[i\ \treated]}$. We assume they observe monomials $\x_i = (v_{i 1},$ $v_{i 2},$ $v_{i 1}^2,$ $v_{i 2}^2,$ $v_{i 1}v_{i 2},$ $v_{i 1}v_{i 2}^2,$ $v_{i 1}^2v_{i 2},$ $v_{i 1}^2v_{i 2}^2)$ and $1_{[i\ \treated]}$. In the language of 
\textbf{Appendix C.2}
, $\{[v_{i 1},v_{i 2},1_{[i\ \treated]}, y_i]\}_{i=1}^n$ is the pristine data and $\{[\x_i,1_{[i\ \treated]} , y_i]\}_{i=1}^n$ is the observed data. This puts the researcher at risk of overfitting the observed covariates in $\x_i$ that are not part of the ground truth.

\begin{figure}
	\centering
	\begin{subfigure}{.49\textwidth}
		\centering
		\includegraphics[width=3.75cm]{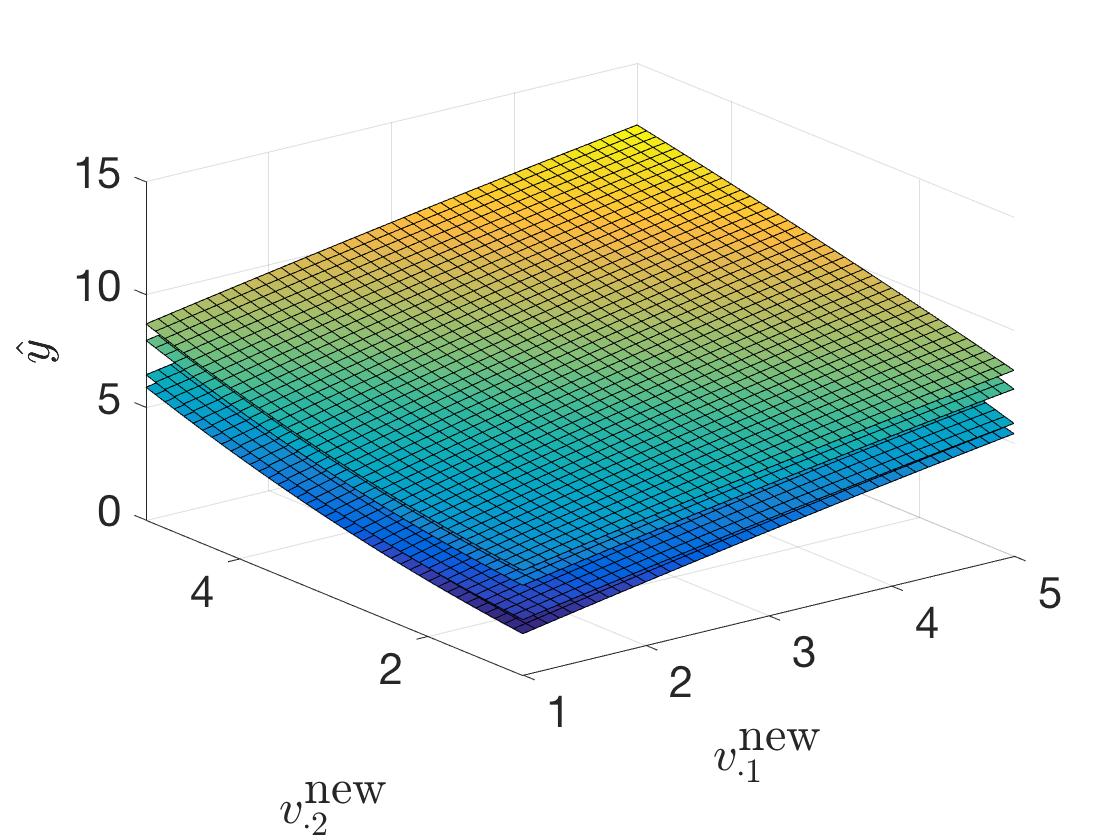}
		\includegraphics[width=3.75cm]{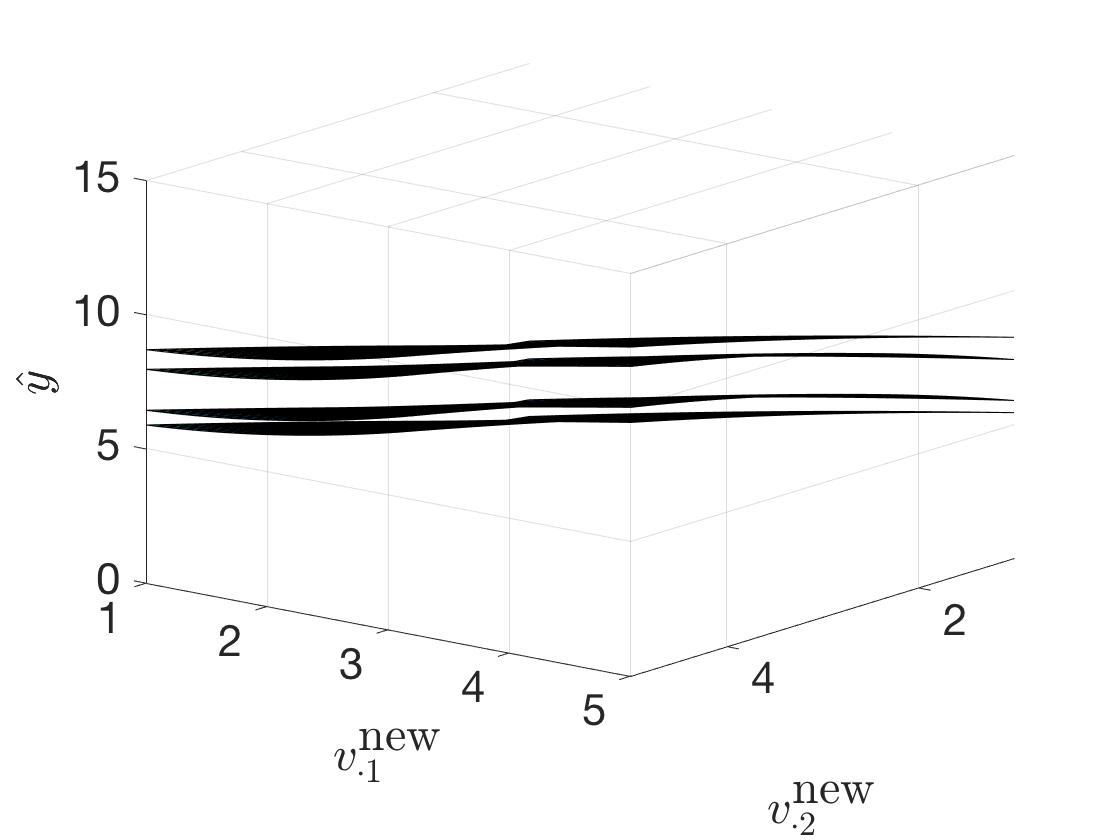}
		\caption{All four treatment prediction curves.}
	\end{subfigure}\hfill

	\begin{subfigure}{.49\textwidth}
		\centering
		\includegraphics[width=3.75cm]{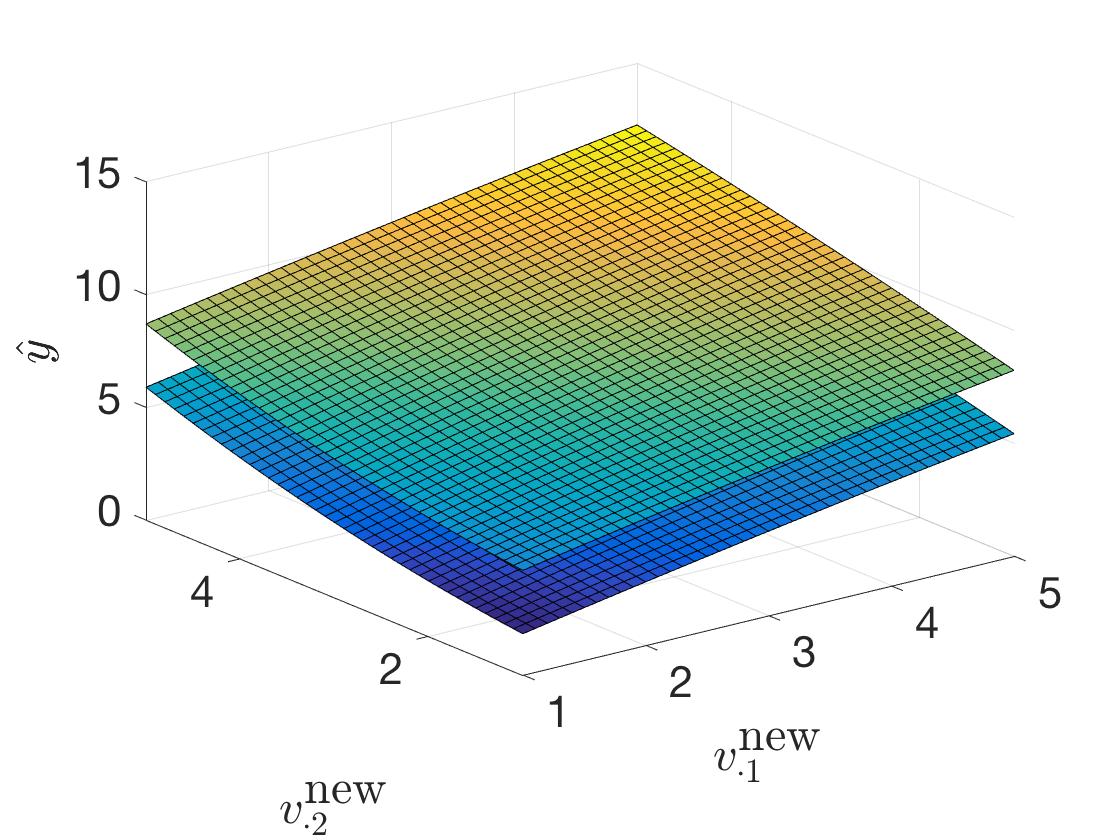}
		\includegraphics[width=3.75cm]{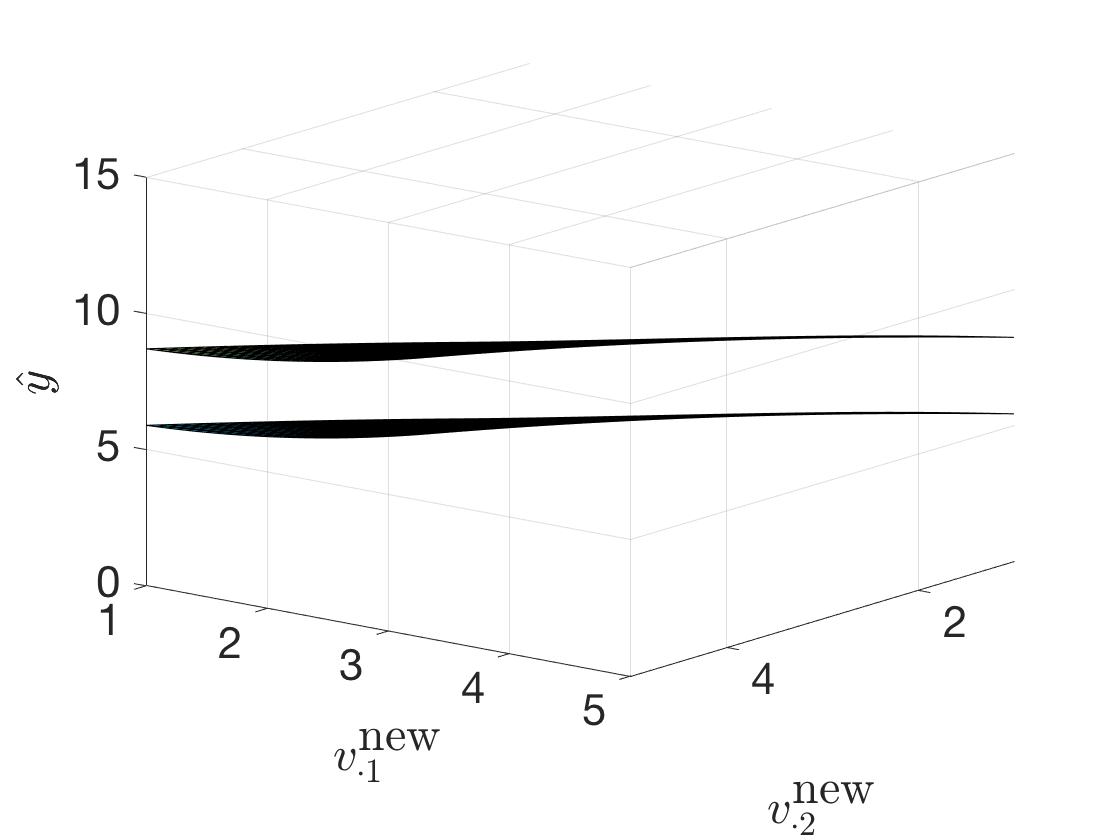}
		\caption{Prediction curves for maximum treatment effect.}
	\end{subfigure}
	\begin{subfigure}{.49\textwidth}
		\centering
		\includegraphics[width=3.75cm]{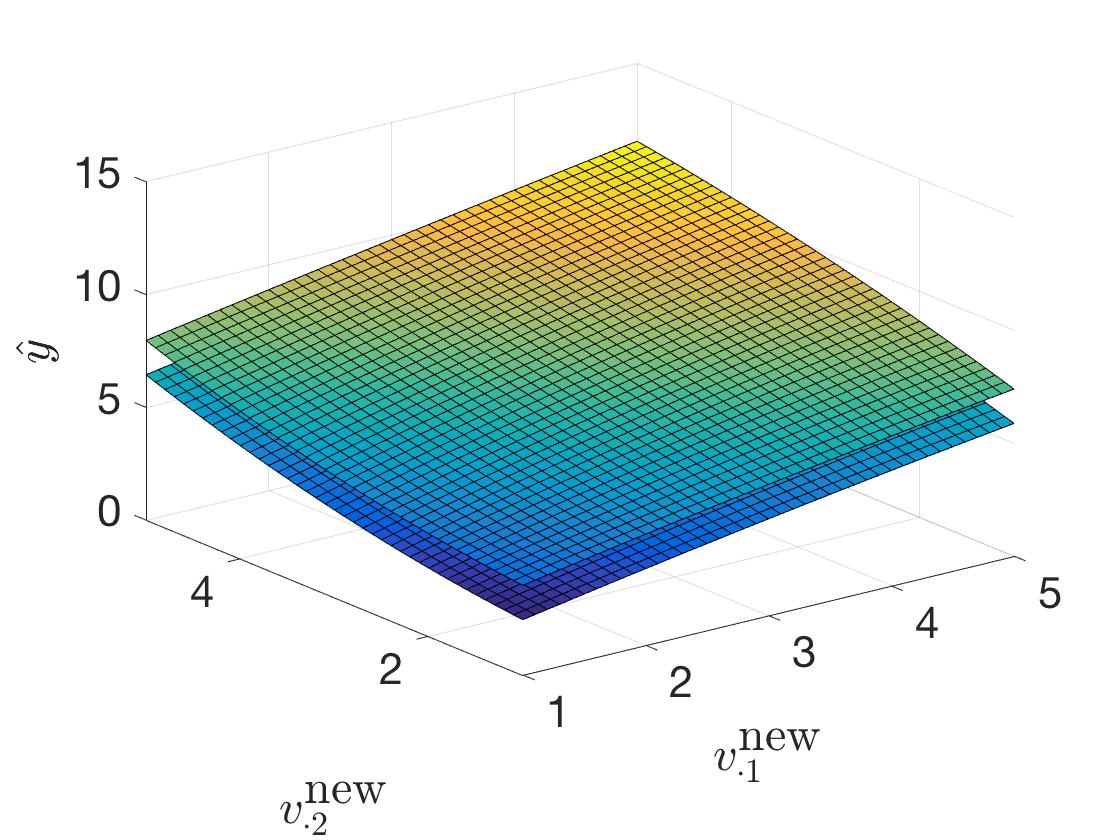}
		\includegraphics[width=3.75cm]{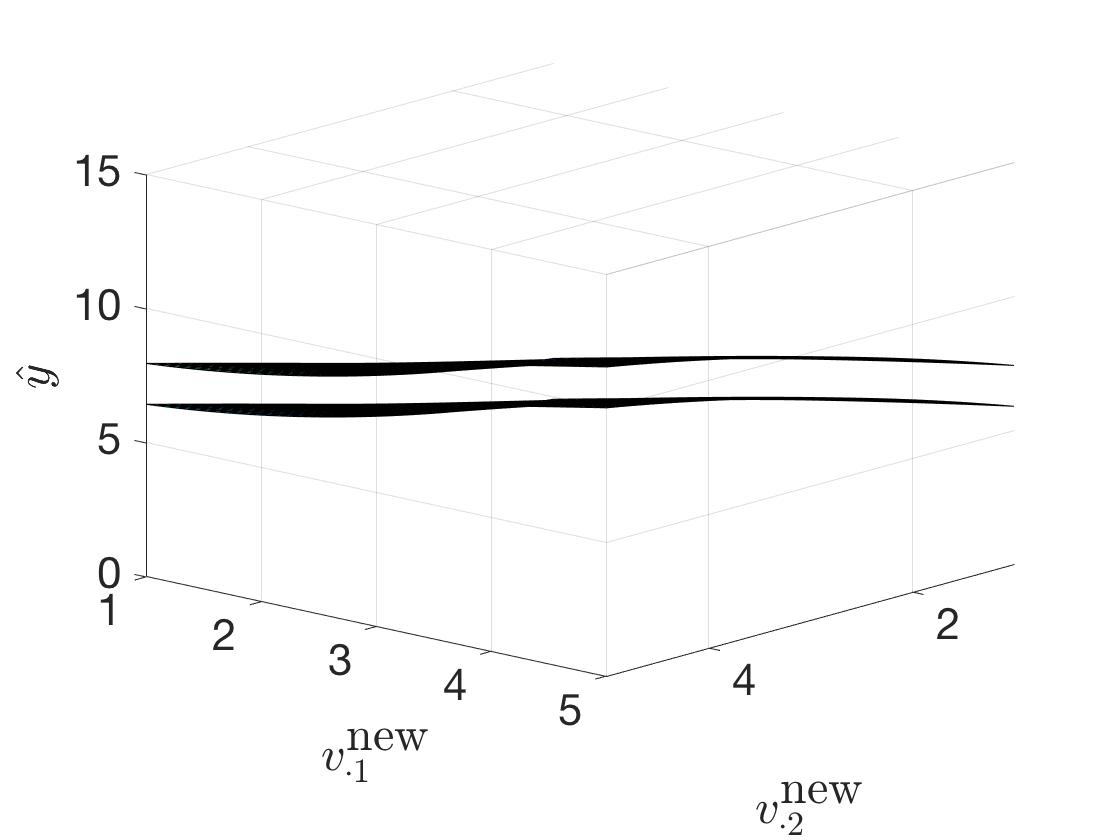}
		\caption{Prediction curves for minimum treatment effect}
	\end{subfigure}
	\vspace{3mm}
	\caption{
		\label{FigureG1} For each of the three panels, the left and right figures are two different vantage points of the same figure. Since the true data generation process in Equation (\ref{eq:true_process}) depends linearly on only $v_{\cdot 1}$ and $v_{\cdot 1}$, the optimal prediction curve as a function of $v_{\cdot 1}$ and $v_{\cdot 1}$ is a hyperplane. The addition of monomials to the observed $\x$ causes some overfitting.
		(a) All four prediction curves (max/min, treatment/control).
		(b) Prediction curves that yield the maximum treatment effect. The upper curve shows $\hat{y}_{\max,\treated}$ and the lower curve shows $\hat{y}_{\max,\untreated}$. The difference between the curves is the maximum treatment effect, $\beta_{0,\max}^*$. These curves correspond to the top and bottom curves in the top row of plots. 
		(c) Prediction curves that yield the minimum treatment effect. The upper curve shows $\hat{y}_{\min,\treated}$ and the lower curve shows $\hat{y}_{\min,\untreated}$. The difference between the curves is the minimum treatment effect, $\beta_{0,\min}^*$. These curves correspond to the middle two curves in the top row of plots. 
	}
\end{figure}

We simulated a dataset of $n=500$ observations and used Theorem \ref{TheoremGaryConfInterval} to find the values of $\bbeta_{\max}^*$, $\beta_{0,\max}^*$, $\bbeta_{\min}^*$, and $\beta_{0,\min}^*$,
where $\theta$ was set to 10\% higher than the least squares loss of $\bbeta_{LS}^*$. Table \ref{tbl:reg_results} gives the results for the coefficient on treatment indicator, $\beta_0$. To illustrate these results, on a grid of $v_{\cdot 1}^{\textrm{new}}$ and $v_{\cdot 2}^{\textrm{new}}$, we found the vector of monomials that would be observed by the researcher, $\xnew$, and evaluated the four possible outcome predictions (max and min, treatment and control):
\begin{align}
\hat{y}_{\max,\treated} &= \xnew \bbeta_{\max}^* + 1\times\beta_{0,\max}^*\label{eq:pred1}\\
\hat{y}_{\min,\treated} &= \xnew \bbeta_{\min}^* + 1\times\beta_{0,\min}^*  \\
\hat{y}_{\min,\untreated} &= \xnew \bbeta_{\min}^* + 0\times\beta_{0,\min}^* \\
\hat{y}_{\max,\untreated} &= \xnew \bbeta_{\max}^* + 0\times\beta_{0,\max}^*.\label{eq:pred4}
\end{align}
Equations (\ref{eq:pred1}) through (\ref{eq:pred4}) are ordered by value, from largest to smallest.
This gives four surface plots, shown in Figure \ref{FigureG1} from different rotations. Asymptotically, or if we had a larger number of points, the curves would be hyperplanes since the ground truth in Equation (\ref{eq:true_process}) depends linearly on $v_{\cdot 1}$ and $v_{\cdot 2}$. As it stands, the curves are very close to the optimal hyperplanes, overfitting only slightly.

We would like to consider \textit{individual} treatment effects, where the treatment effects can differ between units. The simple regression setting above will predict a constant treatment effect for all units, so we need to have a more flexible modeling approach. 

\begin{table}[]
	\centering
	\begin{tabular}{ccc}
		$\beta_{0,\min}^*$ & $\beta_{0,LS}^*$  & $\beta_{0,\max}^*$  \\ \hline
		1.52 & 2.16 & 2.80
	\end{tabular}
	\caption{Minimum, least-squares, and maximum coefficient on the treatment indicator. $[\beta_{0,\min}^*,\beta_{0,\max}^*]$ is the tethered hacking interval. The ground truth is $\beta_0 = 2$.}
	\label{tbl:reg_results}
\end{table}


\subsection{Scenario 2: Individual Treatment Effect} \label{sec:scen2}

For the second regression scenario we consider, the regression model is more flexible, including separate terms for treatment and control. Our goal is to find the range of treatment effects for a particular point $\xnew$. 

To explain the motivation for this problem, let us consider a new patient receiving a prediction of the expected treatment effect for a drug. Before taking the drug, the patient might want to know whether there are other reasonable models that give different predictions. That is, the patient might want to know the answer to the following:
\textit{Considering all reasonable models for predicting treatment effects, what are the largest and smallest possible predicted treatment effects for this drug on me?}

To determine the range, we solve:
\begin{align*}
\max_{\bbeta,\beta_0} f\left(\xnew\right) &\textrm{ s.t. } \sum_{i=1}^n \left(f(\xnewi)-y_i^{\textrm{(new)}}\right)^2\leq \theta.
\\
\min_{\bbeta,\beta_0} f\left(\xnew\right) &\textrm{ s.t.} \sum_{i=1}^n \left(f(\xnewi)-y_i^{\textrm{(new)}}\right)^2\leq \theta.
\end{align*}

The model is:
\begin{align*}
f(\x,\textrm{treated or control})=1_{\textrm{control}}[\beta_1^c x_{.1} + \beta_2^c x_{.2} + ... \beta_p^c x_{.p}] + 1_{\textrm{treated}}[\beta_1^t x_{.1} + \beta_2^t x_{.2} + ... \beta_p^t x_{.p}].
\end{align*}
Using notation $w_i=1$ for treatment points, and $w_i=0$ for control points, the least squares loss thus decouples, leading to separate regression problems for the treatment and control points:
\begin{eqnarray*}
	\lefteqn{\sum_{i=1}^n \left(f(\x_i,w_i) - y_i \right)^2 }\\
	&=& 
	\sum_{i:w_i=1} \left(f(\x_i,1) - y_i \right)^2 + 
	\sum_{i:w_i=0} \left(f(\x_i,0) - y_i \right)^2\\
	&=& 
	\sum_{i:w_i=1} \left([\beta_1^c x_{i1} + \beta_2^c x_{i2} + ... \beta_p^c x_{ip}] - y_i \right)^2+
	\sum_{i:w_i=0}\left([\beta_1^t x_{i1} + \beta_2^t x_{i2} + ... \beta_p^t x_{ip}] - y_i \right)^2.
\end{eqnarray*}
Because the first sum involves only the control observations and control coefficients, and the second sum involves only treatment observations and treatment coefficients, this decouples as two separate regressions, one for the control group, and one for the treatment group.
We will assume that the user wants neither of the regressions to be too suboptimal, so we will have separate constraints $\theta$ on the quality of each regression. We will find the maximum and minimum values for the control regression and the treatment regressions (four values). All of these optimization problems are very similar, so for simplicity, we solve the optimization problem on a generic regression problem, for point $\xnew$. Here $\xnew$ does not need to be one of the training observations.

\begin{theorem}[Hacking Intervals for Least-Squares Individual TE]
	\label{thm_cynthias_problem}
	Consider the hacking interval optimization problems:
	\begin{eqnarray*}
		&&\max_{\bbeta} (\mathbf{x}^{{\rm (new)}}\bbeta) \textrm{ such that } \sum_{i=1}^n (y_i-\x_i\bbeta)^2\leq \theta,\\
		&&\min_{\bbeta} (\mathbf{x}^{{\rm (new)}}\bbeta) \textrm{ such that } \sum_{i=1}^n (y_i-\x_i\bbeta)^2\leq \theta.
	\end{eqnarray*}
	Define $\bbeta^*_{LS}:=(\X^T\X)^{-1}\X^T\mathbf{Y} $, define $\Upsilon=(\X^T\X)^{-1}\x^{({\rm new})T}$, which is a vector of size $p$, $\SSE=\|\mathbf{Y} -\X\bbeta^*_{LS}\|^2$, and
	\begin{equation*}
	\tilde{\mu} = \frac{\sqrt{\theta -  \SSE}}{\|\X\Upsilon\|}.
	\end{equation*}
	The solutions to the optimization problems above are:
	\[
	\bbeta^*_{-} = \bbeta^*_{LS} -\tilde{\mu}\Upsilon, \qquad \bbeta^*_{+} = \bbeta^*_{LS}+\tilde{\mu}\Upsilon.
	\]

\end{theorem}


\begin{theorem}[Individual TE Hacking Intervals and Standard Confidence Intervals]
	\label{thm_ATE_CI}
	Start with a standard confidence interval for $\xnew\bbeta$ under usual assumptions (normality of errors given a linear model), which is given by the boundary points:
	\begin{equation*}
	\bbeta^*_{LS} \pm \tpn \sqrt{\frac{\SSE}{n-p-1}}\sqrt{\xnew (\X^T\X)^{-1} \xnewt}
	\end{equation*}
	where $\tpn$ is the $1-\alpha/2$ quantile of a $t$ distribution with $n-p-1$ degrees of freedom. Then, in order to keep the hacking interval from Theorem \ref{thm_cynthias_problem} the same as the standard one, we would take the following value for $\theta$:
	\begin{equation*}
	\theta = \SSE \left(1+ \frac{\tpn^2}{n-p-1}\right).
	\end{equation*}
	
\end{theorem}

We can use the result of Theorem \ref{thm_cynthias_problem} to determine the hacking interval, which in this case is the range of causal effect estimates for $\xnew$. Let us apply Theorem \ref{thm_cynthias_problem} to the treatment regression and the control regression separately. We thus obtain $\beta_+^{t*}$, $\beta_-^{t*}$, $\beta_+^{c*}$, and $\beta_-^{c*}$. To find the maximum of the causal effect estimate, use:
\begin{equation*}
\max\left(\xnew\bbeta_+^{t*}, \xnew\bbeta_-^{t*} \right) - \min \left(\xnew\bbeta_+^{c*}, \xnew\bbeta_-^{c*} \right). 
\end{equation*}
To find the minimum of the causal effect estimate, use:
\begin{equation*}
\min\left(\xnew\bbeta_+^{t*}, \xnew\bbeta_-^{t*} \right) - \max \left(\xnew\bbeta_+^{c*}, \xnew\bbeta_-^{c*} \right). 
\end{equation*}

\subsubsection{Illustration}
\begin{figure}
	\centering
	\begin{subfigure}{.49\textwidth}
		\centering
		\captionsetup{justification=centering}
		\includegraphics[width=3.75cm]{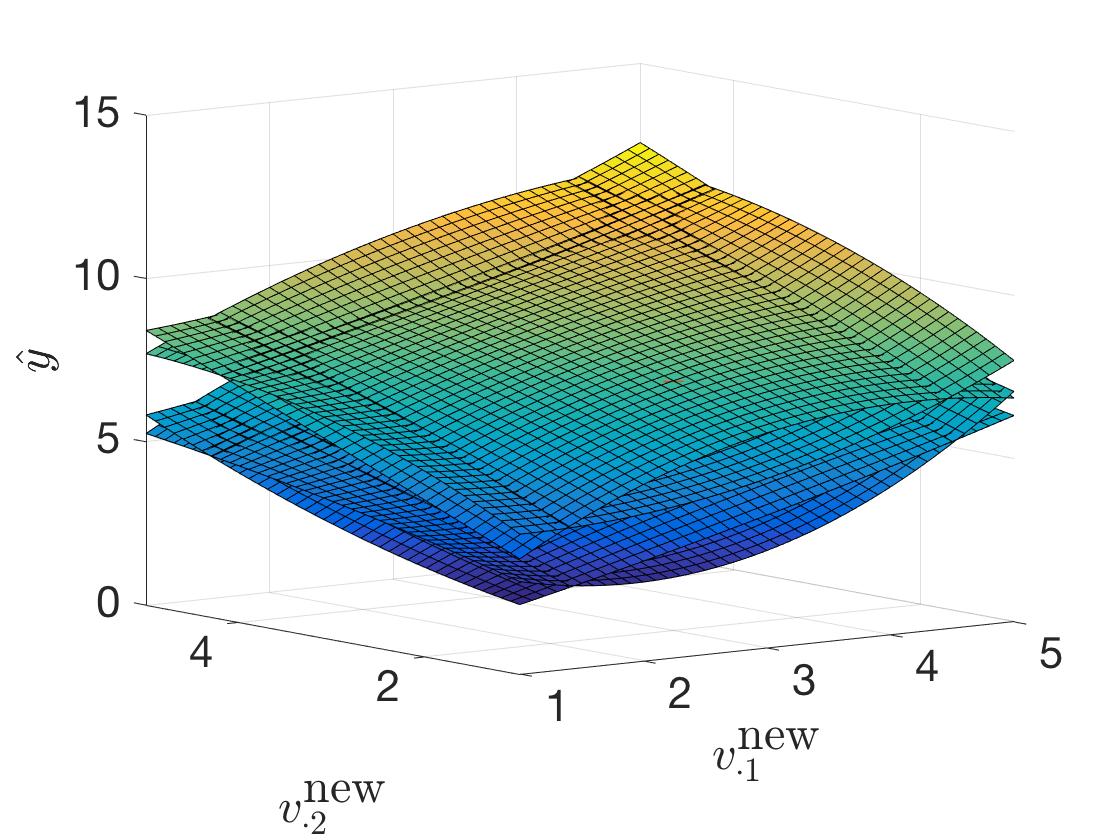}
		\includegraphics[width=3.75cm]{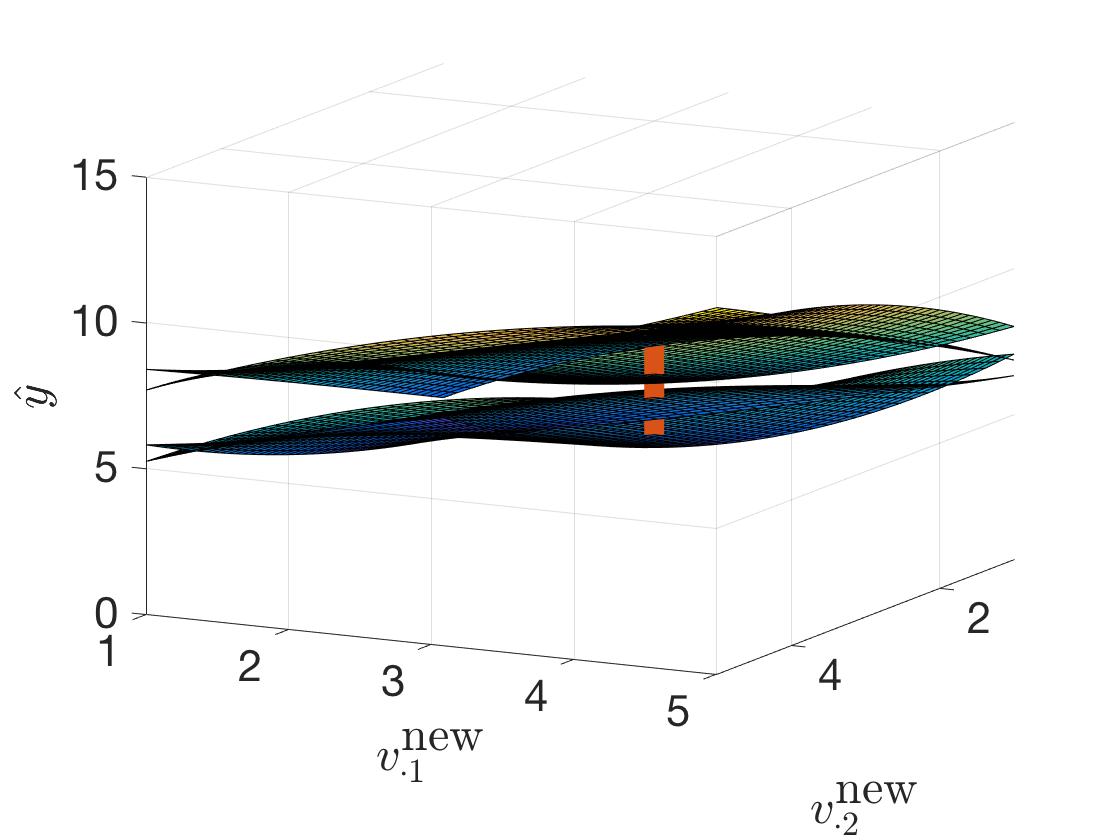}
		\caption{All four models: max and min at $\xnew$ of regressions for treatment and control.}
	\end{subfigure}\hfill

	\begin{subfigure}{.49\textwidth}
		\captionsetup{justification=centering}
		\centering
		\includegraphics[width=3.75cm]{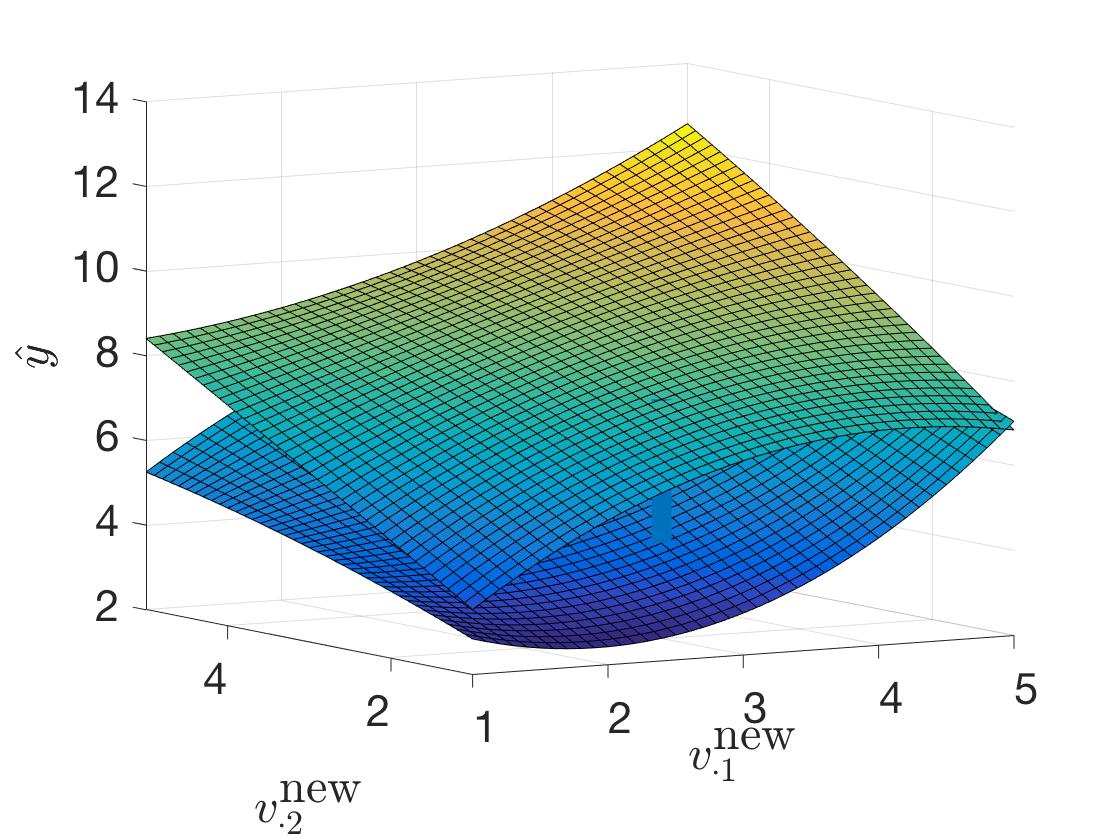}
		\includegraphics[width=3.75cm]{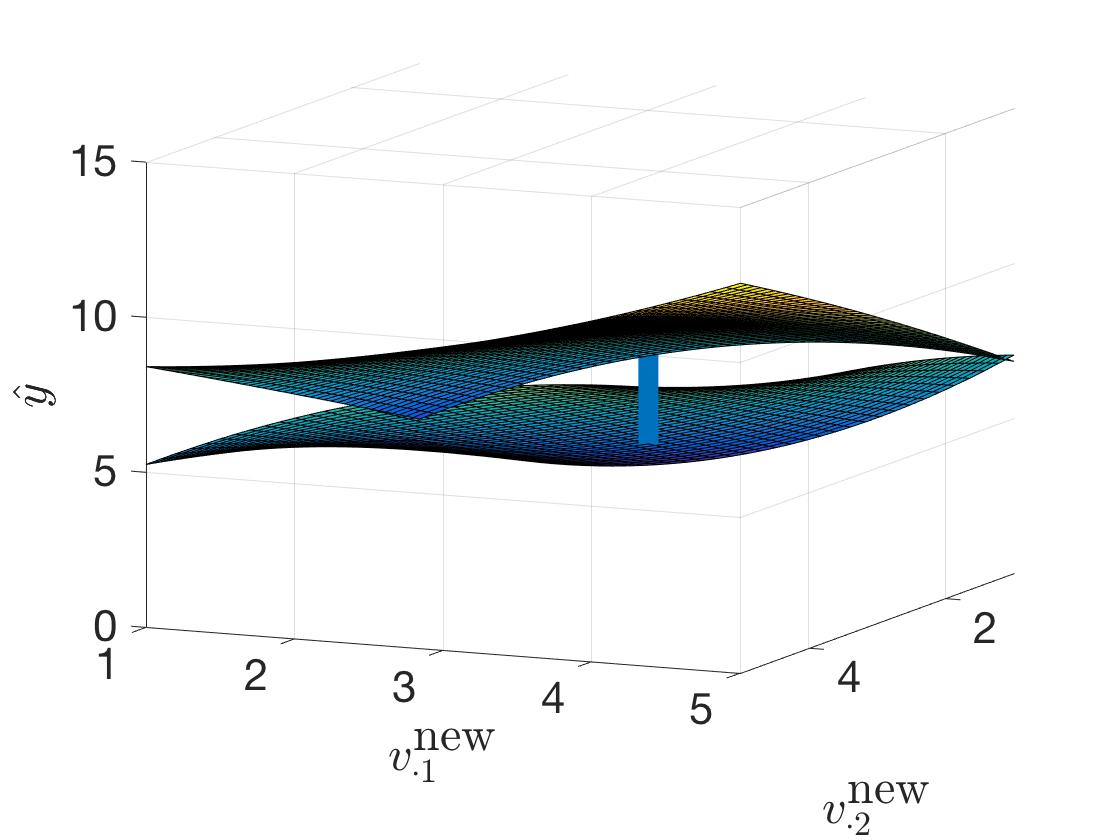}
		\caption{Max of treatment and min of control at $\xnew$. Vertical line drawn at $\xnew$.}
	\end{subfigure}
	\begin{subfigure}{.49\textwidth}
		\centering
		\captionsetup{justification=centering}
		\includegraphics[width=3.75cm]{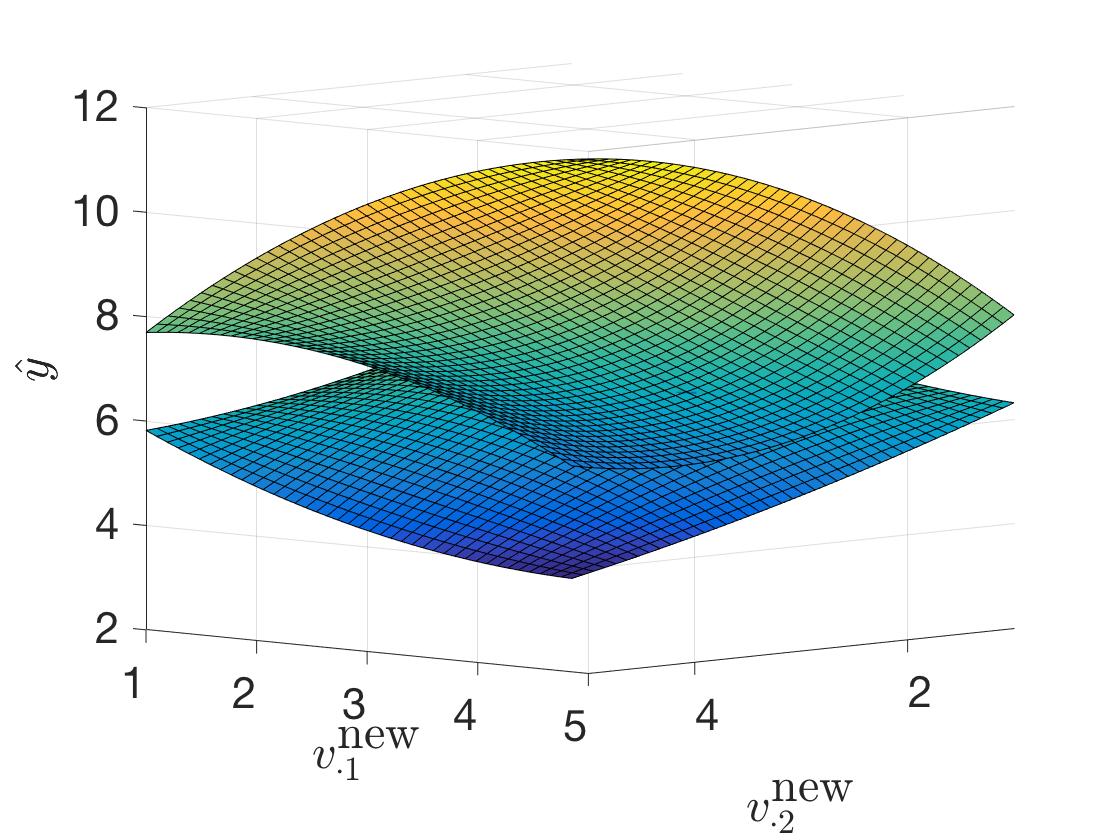}
		\includegraphics[width=3.75cm]{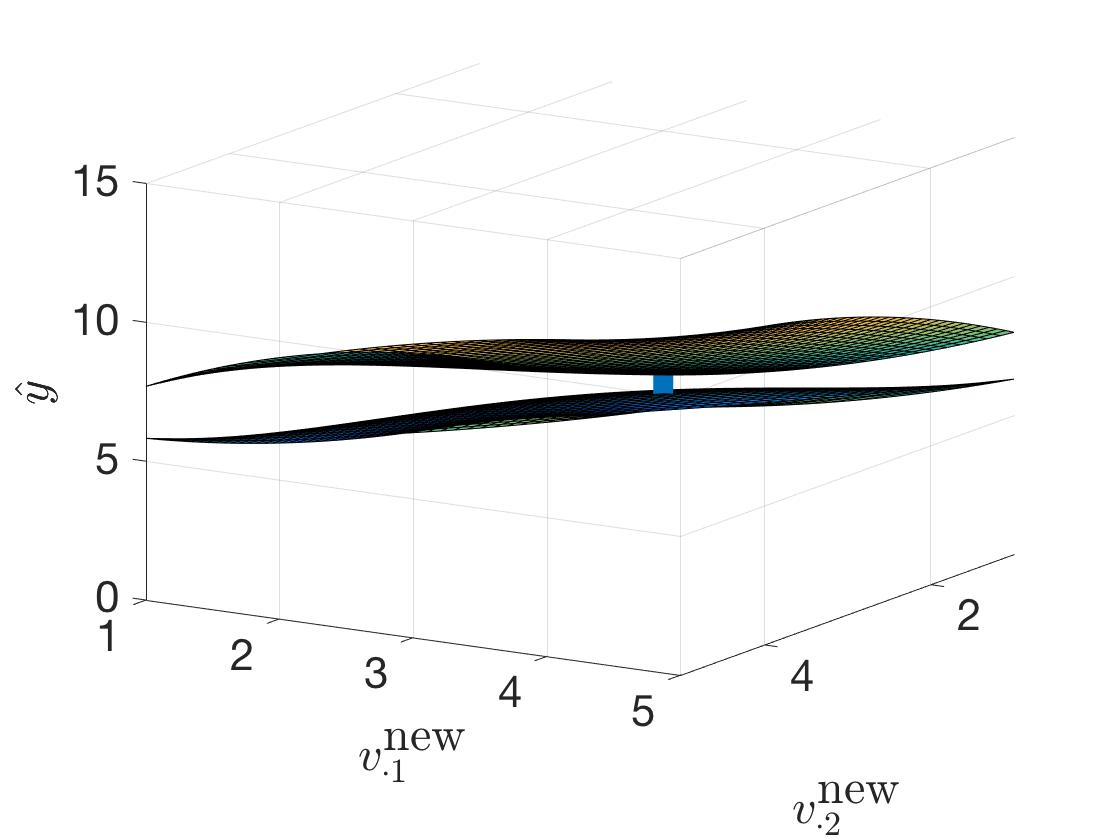}
		\caption{Min of treatment and max of control at $\xnew$. Vertical line drawn at $\xnew$.}
	\end{subfigure}
	\vspace{3mm}
	\caption{
		\label{FigureG2} For all three panels, the left and right figures are two different vantage points of the same figure. (a) All four regressions (max/min, treatment/control). 
		(b) Maximizing the gap between treatment and control at $\xnew$.  The upper curve is the regression for maximizing expected outcomes on the treated at $\xnew$. 
		The lower curve is the regression for minimizing expected outcomes on the control units at $\xnew$. One can see how the curves pull away from each other at $\xnew$ to make the differences between treatment and control as large as possible. 
		(c) Minimizing the gap between treatment and control at $\xnew$. The upper curve is the regression for minimizing outcomes on the treatment units at $\xnew$. The lower curve is the regression for maximizing the control outcomes at $\xnew$.
		Here the curves pull towards each other to minimize the estimated treatment effect.
	}
\end{figure}

We continue with the same data generation process we used in Section \ref{subsecillus}, where the ground truth outcomes are created as follows:
\[y_i = 2\times 1_{[\treated]} + v_{i 1} + v_{i 2}+ \epsilon.\]
We chose $\xnew$ to be created from the point $v_1^{\textrm{new}}=3$, $v_2^{\textrm{new}}=2$.
Here we created four separate regressions. One regression maximizes the expected outcome at $\xnew$ for the treatment observations. Another regression minimizes the expected outcome at $\xnew$ on the treatment observations. Analogous regressions are created for the control observations. Figure \ref{FigureG2} shows these regressions explicitly for $\xnew=\{3,2\}$. One can see the regressions starting to bend away from each other at $\xnew$ for the maximization problem, and bend towards each other for the minimization problem. We placed a blue line between the curves at the point $\xnew$.


\section{Application: Recidivism Prediction}\label{sec:app}

Understanding the potential impact of researcher choices on machine learning methods becomes especially important when issues of fairness are involved. Although there does not exist a widely accepted mathematical definition of fairness when assessing risk with machine learning \citep{berk}, if a machine learning method could reach opposing conclusions about a person or group of persons were small adjustments to a dataset or hyperparameters made, then this could potentially undermine any definition of fairness (one could simply argue a negative decision to be unfair because an equally good model exists that predicts the opposite).
A hacking interval quantifies the degree to which this can happen.

In the criminal justice system, algorithms are increasingly being used to make risk assessments about defendants, for example their risk of failing to appear in court or reoffending. Clearly, issues of fairness are involved. One such algorithm is COMPAS (or, Correctional Offender Management Profiling for Alternative Sanctions), created by Northpointe, Inc.
COMPAS produces three decile scores that indicate the risk that a defendant will fail to reappear in court, reoffend, or violently reoffend. 
As of October 2017 it was used by 4 of 58 counties in California \citep{california}.
It is a proprietary algorithm that bases its assessment on a questionnaire that is either pulled from criminal records or answered by the defendant. The data gathered by the questionnaire is not publicly available. 
ProPublica assembled COMPAS scores and other data --- including criminal history and demographic information --- on more than 7,000 defendants in Broward County, Florida, from 2013 through 2014 with the help of the Broward County Sheriff's Office \citep{ProPublica}. Using the same metric used by Northpointe --- whether or not a defendant was charged with a crime within two years of the COMPAS score calculation --- ProPublica concluded that COMPAS was biased against African Americans. For example, they found that of African American defendants who did not reoffend, 45 percent were misclassified as higher risk, while of Caucasian defendants who did not reoffend, only 23 percent were misclassified as higher risk.
Northpointe has issued a rebuttal that argues a definition of fairness based on a false-positive rate is not appropriate in this case \citep{northpointe}. 
\citet{angelino} argues that while COMPAS is ostensibly not influenced by race, its dependence on prior record could effectively induce dependence on race due to disproportionate arrest rates that count towards one's prior record. This agrees with the sentiment of other work on interpretable models for recidivism \citep{zeng}. More work on this dataset has provided further insight into how COMPAS may depend on prior record as well as age \citep{Rudin2020Age}.

In our analysis, we use the data collected by ProPublica, but our interest is not in comparing a risk assessment score like COMPAS against a given definition of fairness. Rather, we are interested in the impact that researcher choices could have on conclusions made about this dataset. 
In Section \ref{sec:recid_feat}, we use the methods of Section \ref{sec:ex_feat} to assess the impact that a new feature created by the researcher could have on inferences about the population, in this case the odds ratio of reoffending and gender. This is an example of a prescriptively-constrained hacking interval since we explicitly constrain researcher choices about the new feature. 
In Section \ref{sec:recid_svm}, we use the methods of Section \ref{sec:ex_svm} to assess the impact of researcher choices on the predictions of a support vector machine about individual defendants. This is an example of a tethered hacking interval since we constrain researcher choices only through their impact on the loss function. For both applications we use the following set of features:
\begin{itemize}
	\item \texttt{c\_charge\_degree\_F}: Binary indicator if the most recent charge prior to the COMPAS score calculation is a felony.
	
	\item \texttt{sex\_Male}: Binary indicator if the defendant is male.
	
	\item \texttt{age\_screening}: Age in years at the time of the COMPAS score calculation.
	
	\item \texttt{age\_18\_20}, \texttt{age\_21\_22}, \texttt{age\_23\_25}, \texttt{age\_26\_45}, and \texttt{age\_\_45}: Binary indicators based on \texttt{age\_screening} for age groups 18-20, 21-22, 23-25, 26-45, and greater than 45, respectively.
	
	\item \texttt{juvenile\_felonies\_\_0}, \texttt{juvenile\_misdemeanors\_\_0}, and \texttt{juvenile\_crimes\_\_0}: Binary indicators one or more juvenile felony, misdemeanor, or crime, respectively. We use binary indicators because the counts of each are highly right-skewed. 
	
	\item \texttt{priors\_\_0}, \texttt{priors\_\_1}, \texttt{priors\_2\_3}, and \texttt{priors\_\_3}: Binary indicators of whether the number of priors is 0, 1, 2-3, or more than 3, respectively.
\end{itemize}
We filtered the dataset to include only defendants whose most recent charge prior to the COMPAS score calculation was a felony or misdemeanor and occurred at most 30 days prior to the COMPAS score calculation (otherwise we assume this charge did not trigger the COMPAS score calculation, so it seems that data about this defendant are missing). The binary indicator variables for age and number of priors were added to the dataset because, in general, recidivism is highly nonlinear with respect to these features.

\subsection{Prescriptively-Constrained Example: Adding a New Feature} \label{sec:recid_feat}
We suppose a researcher is interested in the odds ratio between gender and recidivism but is allowed to create a new binary feature $u$, perhaps as a function of the existing features or by introducing new data. Notice this is not a valid causal question since gender is not assignable, but we only use the mathematical tools of causal sensitivity analysis. A benefit of this approach is that we do not need to understand exactly what the new feature is, only its relationship to the outcome $y$ (whether or not a defendant reoffends) and ``treatment'' $w$ (gender). In the setup described in Section \ref{sec:ex_feat}, this means the researcher specifies constraints $OR_{yu}\in[a,b]$, $|p_1 - p_0| \le c$, and $p_0\ge d$ (by specifying $a$, $b$, $c$ and $d$), where $p_0:=p(U\mid w=0)$, $p_1:=p(U\mid w=1)$. We will use a simple version where $OR_{yu}$ is fixed (or, equivalently, $a=b=OR_{yu}$). As shown in Section \ref{sec:ex_feat}, the hacking interval can be calculated as a function of $c$. 

\begin{figure}[!h]
	\centering
	\includegraphics[scale=0.3]{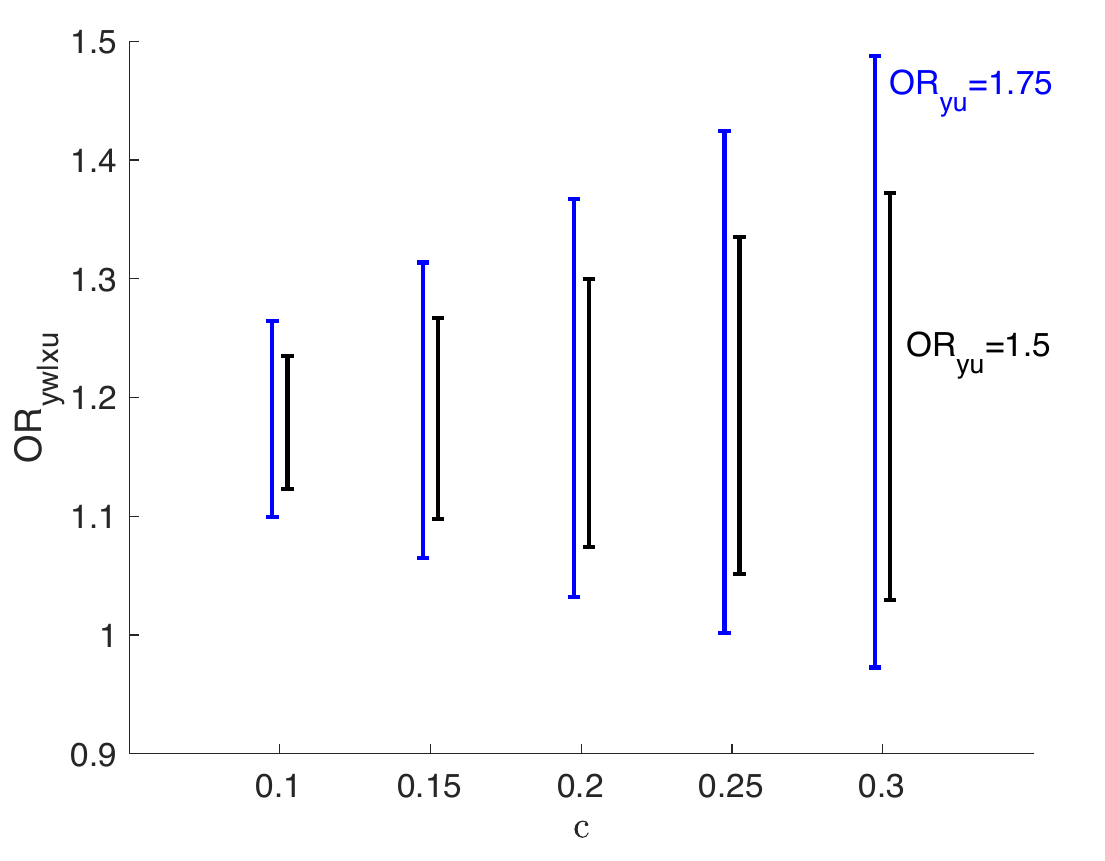}
	\caption{Hacking intervals for $OR_{yw\mid \x, u}$ for different values of constraints $c$ and $a=b=OR_{yu}$.}
	\label{fig:sensitivity}
\end{figure}

Figure \ref{fig:sensitivity} shows hacking intervals for $OR_{yw\mid \x, u}$ --- the odds ratio between recidivism and gender adjusted for the observed covariates $\x$ and the new feature $u$ --- for each combination of $c\in(0.1,0.15,0.2,0.25,0.3)$ and $OR_{yu}\in(1.5,1.75)$. These constraints are picked arbitrarily for illustration. In practice, the choice of these constraints describes the degree of freedom given to the researcher. For example, if the researcher were permitted to pick any new binary feature $u$ such that the odds ratio between the outcome and the new feature were $OR_{yu}=1.5$ and the difference between $p_1$ and $p_0$ (the probability of the new feature when the treatment $w$ is present or not present, respectively) were constrained to be less than or equal to $c=0.3$, then the value of $OR_{yw\mid \x, u}$ they could get would necessarily be in the hacking interval $[1.03,1.37]$. For the same restriction of $c=0.3$, if the researcher were permitted to pick $u$ such that $OR_{yu}=1.75$, indicating a stronger relationship between the new feature and the outcome, then they could obtain a value of $OR_{yw\mid \x, u}$ above or below one, since Figure \ref{fig:sensitivity} shows the hacking interval in this case overlaps with one. In other words, with this freedom given to the researcher, they could conclude that the odds ratio between recidivism and gender, after controlling for measured covariates and the new covariate they created, could be above or below one. 

\subsection{Tethered Example: SVM} \label{sec:recid_svm}
We now consider the impact of researcher hacking on predictions of two year recidivism for individual defendants. We use a support vector machine (SVM) as our predictive model. For prediction on a new defendant represented by $\xnew$, SVM calculates the distance of $\xnew$ to the hyperplane that minimizes the hinge loss. If the distance is positive, the model predicts the defendant will reoffend within two years. If the distance is negative, the model predicts the defendant will not reoffend within two years. By adjusting the hyperplane, the tethered hacking interval is the range of distances of $\xnew$ to the  hyperplane that can be achieved within a constraint on the loss. As discussed in Section \ref{sec:ex_svm}, we can find this range of values by solving the dual problem in Equation (\ref{eq:svm_dual}) for $s=-1$ and $s=1$. We do this using the \texttt{fmincon} function in \textsc{Matlab}. We thus solved two optimization problems for each defendant. 

\begin{figure}[!h]
	\centering
	\includegraphics[scale=.55]{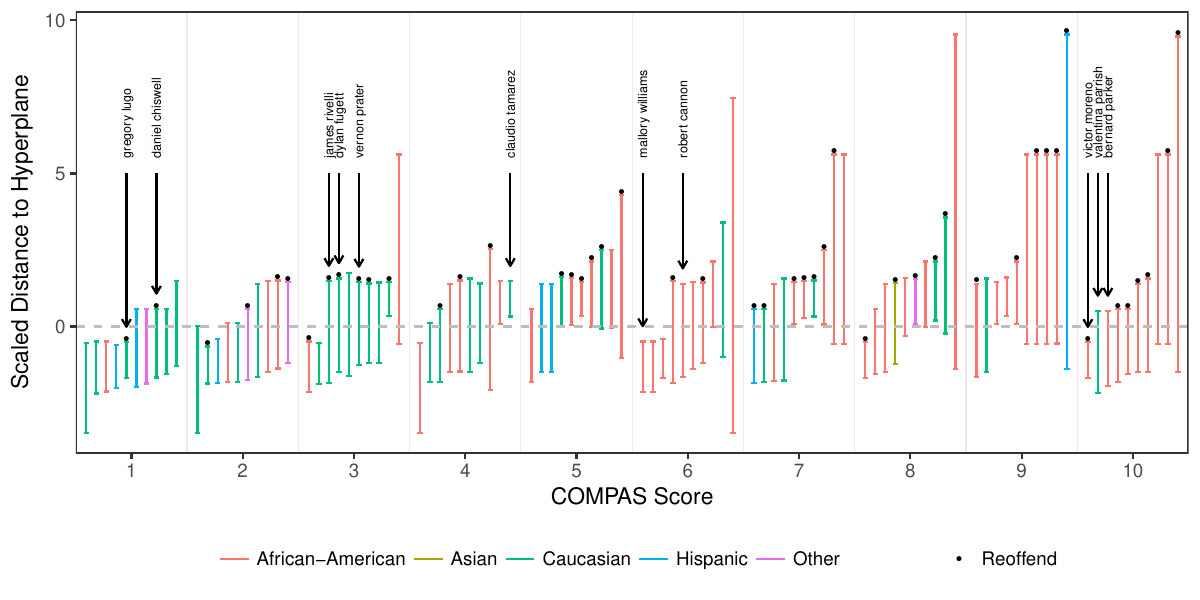}
	\caption{SVM hacking intervals for 10 defendants for each COMPAS score. Loss is constrained to be within 5\% of the minimum loss on a random sample of 1000 defendants.}
	\label{fig:svm_by_decile}
\end{figure}

Figure \ref{fig:svm_by_decile} shows the hacking intervals for 10 selected defendants from each group of COMPAS scores. We included a few individuals highlighted in an article by ProPublica \citep{ProPublica} and randomly selected the rest. The loss is constrained to be within 5\% of the minimum loss on a group of 1000 defendants randomly selected from the remaining defendants (so, each prediction in Figure \ref{fig:svm_by_decile} is out of sample).

Consider three possible cases: (i) The hacking interval is entirely below zero, (ii) the hacking interval is entirely above zero, or (iii) the hacking interval overlaps with zero. In case (i), this means there does not exist an SVM model such that the loss on the 1000 training observations is within 5\% of the minimum loss and the model predicts the defendant will reoffend; all ``reasonable'' models (i.e., within this loss constraint) predict the defendant \textit{will not} reoffend. In case (ii), when the hacking interval is entirely above zero, the interpretation is the same except all reasonable models predict the defendant \textit{will} reoffend. In (iii), when the hacking interval overlaps with zero, then reasonable SVM models exist that make either prediction. Although this is only a sample of the data, notice that of the ten defendants shown here with COMPAS scores of ten --- the riskiest possible COMPAS score --- nine of them have hacking intervals that overlap with zero. On the other hand, of the ten people shown here with COMPAS scores of one --- the least risky COMPAS score --- five of them have hacking intervals entirely above zero.

In the ProPublica article \citep{ProPublica}, several pairs of defendants are highlighted. For each pair, one defendant received a low COMPAS score despite a significant criminal history, while the other received a high COMPAS score despite a limited criminal history.  For example, James Rivelli and Robert Cannon were both charged with theft, but Rivelli was charged with felony grand theft and possession of heroin, while Cannon was charged with misdemeanor petit theft. In addition, Rivelli had three prior arrests, including for felony aggravated assault and felony grand theft, while Cannon had none. Despite this, Rivelli --- who is white --- received a low risk COMPAS score of three, while Cannon --- who is black --- received a medium risk COMPAS score of six. Rivelli later reoffended in Broward County with grand theft again while Cannon did not. Interestingly, the hacking intervals for both defendants overlapped with zero, indicating justifiable SVM models (on our limited feature set) could have made either prediction. 
The hacking intervals also overlap with zero for the similarly contrasting pair of Bernard Parker and Dylan Fugett, both arrested on drug charges. For the pair of Vernon Prater and Brisha Borden, both arrested on petty theft charges, the more experienced criminal Prater also has a hacking interval that overlaps with zero but we do not have data on Borden. The exception is Mallory Williams, who received a medium risk COMPAS score of six after a DUI arrest and only two prior misdemeanors. Her hacking interval is entirely below zero, meaning no justifiable SVM model would predict she would reoffend in this experiment. She did not reoffend. In general, we see a high degree of uncertainty from SVM models for the individuals discussed in this article. The counterpart to Mallory Williams in the ProPublica article, Gregory Lugo, illustrates how offense data can be easily misinterpreted. Gregory Lugo was charged with a DUI but had zero priors according to the data we used in our analysis. Not surprisingly, his COMPAS score was low and his hacking interval was entirely below zero. However, ProPublica claimed he had four priors, including three DUIs, and used this as an example of a poorly calibrated COMPAS score. This appears to be a misinterpretation of the data: all of his supposed prior offenses have the same offense date as the offense related to his COMPAS score calculation, so the supposed prior offenses appear to be re-recordings (perhaps for ordinary bureaucratic reasons) of the same offense.

There are other interesting examples in Figure \ref{fig:svm_by_decile}. 
Claudio Tamarez, a 30 year old Caucasian male, received a COMPAS score of 4, which means low risk, following a charge for possession of phentermine and despite 9 priors that included battery on an officer. In contrast, his hacking interval was entirely above zero. He did not recidivate within the 2 year follow-up period, though. 
Daniel Chiswell, a 41-year old Caucasian male, was assigned a COMPAS score of only one despite being charged with felony possession of heroin and having previously been charged with felony battery on an officer. His hacking interval overlapped with zero, meaning there exists a reasonable SVM model that would have predicted he would reoffend. He was charged again with felony possession of heroin later that year. 
Valentina Parrish, a 21 year old Caucasian female, was charged with driving under the influence and possession of less than 20 grams of cannabis. She was given a COMPAS score of ten. 
In contrast, her hacking interval, $[-2.16,0.50]$, was mostly below zero, though not entirely. She did not reoffend.
There are also examples that illustrate limitations of our limited feature set.
Victor Moreno, a 31 year old African American male, received a COMPAS score of 10 despite zero priors. However, the arrest related to his COMPAS score calculation included felony charges of battery, tampering with a victim, tampering with physical evidence, and delivering cocaine. Our SVM model, without access to the content of these charges, not surprisingly gave him a low hacking interval given his lack of prior offenses.

\begin{figure}[!h]
	\centering
	\includegraphics[scale=.55]{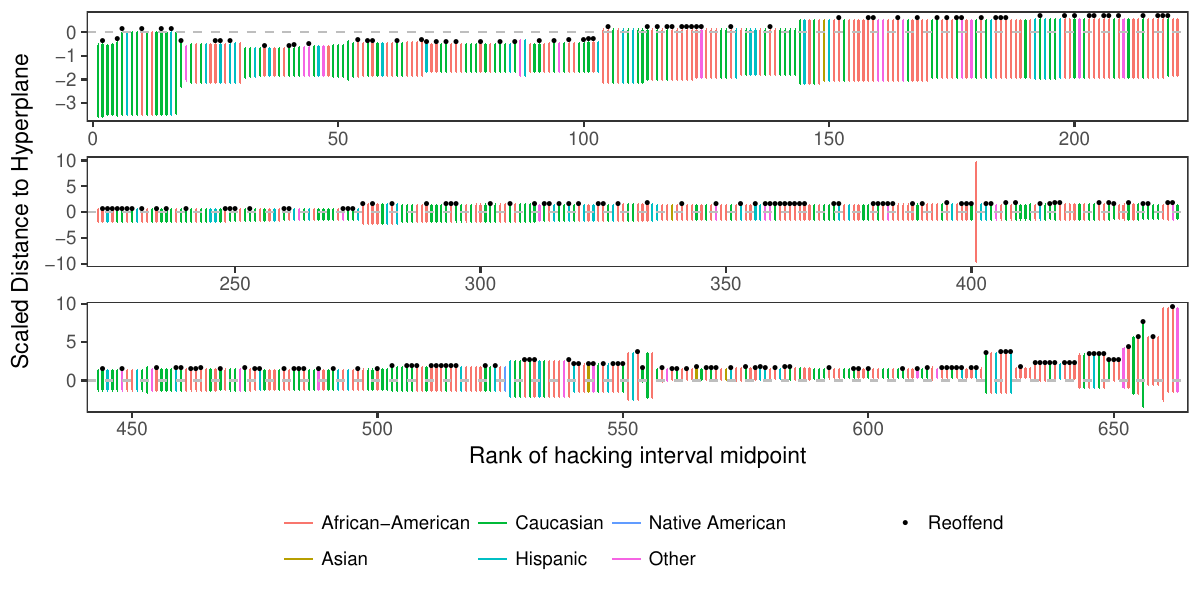}
	\caption{SVM hacking intervals for all defendants with a COMPAS score of 3. Loss is constrained to be within 5\% of the minimum loss on a random sample of 1000 defendants.}
	\label{fig:svm_by_decile3}
\end{figure}

\begin{figure}[!h]
	\centering
	\includegraphics[scale=.55]{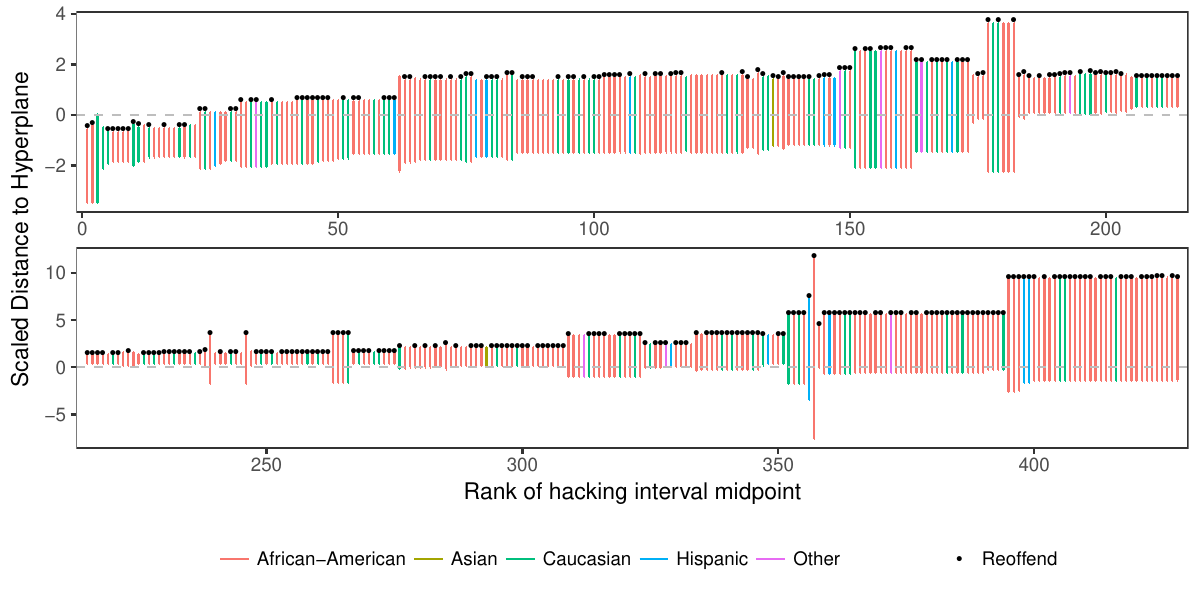}
	\caption{SVM hacking intervals for all defendants with a COMPAS score of 8. Loss is constrained to be within 5\% of the minimum loss on a random sample of 1000 defendants.}
	\label{fig:svm_by_decile8}
\end{figure}

Figures \ref{fig:svm_by_decile3} and \ref{fig:svm_by_decile8} show the hacking intervals for every defendant in our dataset with COMPAS scores of three and eight, respectively. The loss constraint is the same as above (within 5\% of the minimum loss on the same 1000 defendants). Of the 663 people in our dataset with COMPAS scores of three --- a ``low risk'' score --- 75 of them had hacking intervals entirely above zero. Again, this means that, had SVM been used for prediction, any reasonable model would have predicted that they would reoffend. These 75 people had an average of about 6.3 priors and 35 of them reoffended. 
Conversely, of the 428 people in our dataset with COMPAS scores of eight --- a ``high risk'' score --- 121 of them had hacking intervals entirely below zero, meaning any reasonable SVM model would predict that they would not reoffend. These 121 people had an average of about 8.75 priors and 94 of them reoffended. 
This potentially means we may be missing data on their past criminal history that is not in the dataset we use for our analysis. While it is possible that missing information can explain COMPAS scores that are high, it cannot explain COMPAS scores that are too low.

\begin{figure}[!h]
	\centering
	\includegraphics[scale=.55]{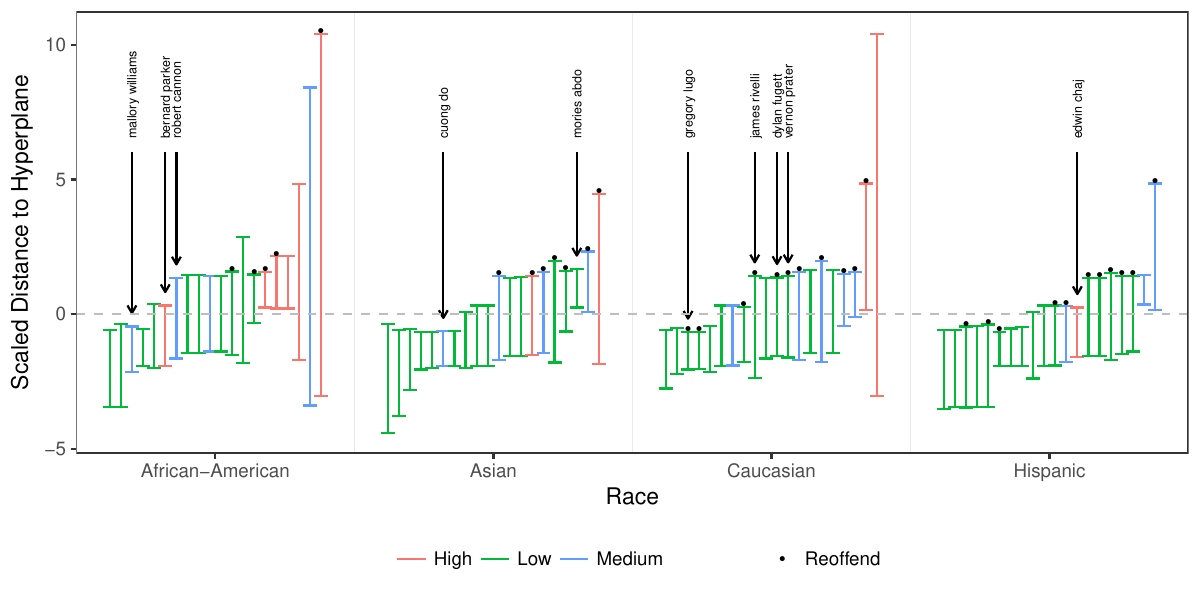}
	\caption{SVM hacking intervals for 10 randomly selected defendants for each race in the dataset (except Native American as there are only 11 in the dataset). Loss is constrained to be within 5\% of the minimum loss on a random sample of 1000 defendants. Color indicates COMPAS scores (high/medium/low).}
	\label{fig:svm_by_race}
\end{figure}

We also show hacking intervals grouped by race in Figure \ref{fig:svm_by_race}. As before, we allow for a 5\% tolerance on the loss on a sample of 1000 defendants, but for this figure we use a different sample of defendants. Each hacking interval in Figure \ref{fig:svm_by_race} is out of sample (i.e., the defendant corresponding to the hacking interval was not included in the 1000 defendant training sample used for the loss constraint). Some of the COMPAS scores again do not align with the hacking intervals. 
Consider Edwin Chaj, a 27 year-old Hispanic male with only one prior related to trespassing, received a COMPAS score of nine following a charge of disorderly intoxication. In contrast to the high-risk COMPAS score, his hacking interval was low ($[-1.59, 0.24]$), although not entirely below zero. He did not reoffend.
Similarly, Cuong Do, a 32 year old Asian male with no priors, received a COMPAS score 8 following charges with felony burglary and misdemeanor petit theft. In contrast to the high-risk COMPAS score, his hacking interval was entirely below zero. He did not reoffend. 
On the other hand, consider Mories Abdo, a 27 year old Asian male with 6 priors, received a COMPAS score of 3 following a battery charge. In contrast to the low-risk COMPAS score, his hacking interval was entirely above zero.
He did not reoffend during the 2 year follow-up period, but did commit felony Aggravated Assault with a Firearm just after the follow-up period ended, according to the Broward County Clerk of the Courts.\footnote{Mories Abdo also committed a Municipal Ordinance for Possession of a Controlled Substance during the two year follow-up period, but this charge does not count as a reoffense in our dataset (there are many charges, like ordinary traffic violations, that do not count as reoffenses).}
%
Figure \ref{fig:svm_by_race} also indicates the individuals discussed in the ProPublica article. Since the 1000-defendant training sample is different from Figure \ref{fig:svm_by_decile}, the hacking intervals are slightly different, but they are each in the same category (below zero, overlapping with zero, or above zero).


We summarize Figures \ref{fig:svm_by_decile} through \ref{fig:svm_by_race} with a couple observations:
\begin{itemize}
	\item \textit{If we had used SVM on our limited data set rather than the COMPAS score to predict re-offense, then for most people there is enough uncertainty in predictions that we could justifiably predict either reoffend or not reoffend.} This can be seen in Figures \ref{fig:svm_by_decile3} and \ref{fig:svm_by_decile8}, where 75\% and 67\% of defendants with COMPAS scores of three and eight, respectively, have hacking intervals that overlap with zero, meaning justifiable SVM models exist that could make either prediction. Even for the extreme cases discussed in the ProPublica article, the hacking intervals often overlapped with zero.
	
	\item \textit{There are many individuals for which no justifiable SVM model would agree with the COMPAS score using our feature set}. In the case of an individual with a low COMPAS score, this means their hacking interval is entirely above zero, while in the case of an individual with a high COMPAS score, this means their hacking interval is entirely below zero. In either case, this is suggestive of an error in the COMPAS calculation. Figure \ref{fig:svm_by_decile3} shows 75 examples of the former case and Figure \ref{fig:svm_by_decile8} shows 121 examples of the latter case. 
\end{itemize}

\section{Related Work and Discussion}
\label{sec:related_work}

Hacking intervals are designed to quantify a form of uncertainty that is usually ignored in statistical inference. This could have implications for scientific research; let us discuss this first. 

\subsubsection*{Problems with replication of scientific studies and proposed solutions}

The evidence for $p$-hacking primarily comes from two types of meta-analyses: replication studies, and the distribution of $p$-values for a set of independent findings \citep[or ``p-curve,''][]{Simonsohn}. For an example of the former approach, a major 2015 study attempted to replicate 100 studies and found very few findings could be reproduced \citep{OpenScienceCollaboration2015}, although a replication of this replication found that the percent of studies that were replicated was not statistically different from the fraction that would be expected to replicate due to chance alone \citep{GilKinPet16}. \citet{Camerer2016} found a higher initial percentage being replicable in 18 economic studies, but still reflecting a problem in the field. In commercial applications, large corporations are keenly aware of this problem: based on their own comparisons, Bayer HealthCare found that only about 20-25\% of preclinical studies were completely in line with their in-house findings \citep{Bayer}. Amgen replicated 11\% of 53 scientific findings \citep{Amgen}.

For the $p$-curve approach, a uniform distribution of $p$-values across articles indicates a lack of significant results, a right-skew indicates a general existence of significant results, and a left-skew, especially near the 0.05 threshold, supposedly indicates $p$-hacking. \citet{Head} concluded that the evidence indicates the existence of ``widespread'' evidence for $p$-hacking after searching all Open Access papers in the PubMed database ($\sim 100,000$ papers). This type of analysis has also been contested \citep{Bishop2016}, and not all meta-analyses have found evidence of $p$-hacking \citep{Jager2014}.

Several types of solutions to $p$-hacking have been proposed: 
\begin{itemize}
	\item We could require researchers to ``pre-register'' the details of their study, so that they cannot selectively make choices to achieve significant results, but this rules out learning from the data in any other way.
	
	\item Another proposal is to reduce the significance threshold \citep{Monogan15,Humphreys,Simmons,Gelman2013}, because when explicitly considering multiple comparisons, decreasing the threshold for significance is sensible (e.g., the Bonferroni correction). Recently, a group of 72 scientists advocated reducing it to 0.005 \citep{Benjamin}, which might lessen false positives but would also invalidate the quantitative meaning of the $p$-value in the first place. This is also a drastic measure, leading to a higher true negative rate, and thus many important results being dismissed as insignificant.
	
	
	\item We could create Bayesian confidence intervals or Bayesian hierarchical models. In comparison to frequentist hypothesis testing, Bayesian hypothesis testing provides a more comfortable interpretation of the conclusion (the probability that the alternative hypothesis is true), but it is still subject to hacking: the introduction of a prior gives the researcher even more discretion, which may lead to more user choices \citep[see][for examples of complicated priors leading to bias]{Gelman2009}. If we place a prior on analysts' decisions, it is easy to argue that any given prior is wrong. An example of this, discussed earlier, is the choice of matching algorithm for treatment and control units in a matched pairs experiment. This is a case where uniform priors do not make sense, but any other choice of prior is not defensible either. 
	
	\item In the case where the researcher does variable selection, \textit{post selection inference} can be used to adjust classical confidence intervals in order to account for the variables being chosen after examining the data. In the case of linear regression, \citet{taylor} present a framework for \textit{specific} variable selection procedures (forward stepwise regression, least angle regression, and the lasso regularization path) and \citet{berk2013} present a framework that holds for \textit{all} variable selection procedures that is more conservative than Scheff\'{e} protection \citep{scheffe}. Hacking intervals differ from post-selection inference in at least two ways: (1) Hacking intervals are more general, as they could include uncertainty to many choices made by the analyst for \textit{any} prediction problem (not just regression), and do not necessarily require \textit{i.i.d.} Gaussian errors; (2) post selection is useful when you already have a model selected and you want to do regular inference, whereas hacking intervals consider robustness to other models that \textit{could} have been selected. Post-selection confidence intervals can be combined with hacking intervals to account for other researcher choices. 
	
	\item The work of \cite{dwork} provides a method to avoid $p$-hacking in a setting where data are provided sequentially, chosen iid from the same distribution. Our setting is very different; in our work, the data could be subject to pre-processing, and the underlying distribution may not exist.
	
\end{itemize}
These solutions are obviously sometimes useful, but often unfulfilling, highlighting the importance, inherent difficulty, and urgency of the problem.



\subsubsection*{Problems with classical inference that are easy to overlook} 

Here we highlight some drawbacks to classical inference, including frequentist, Bayesian, and fiducial inference \citep[see][for a review of a modern version of fiducial inference]{hannig}, in the way they are used in practice and how hacking intervals can help to fix these issues. \\

\textit{In cases where a superpopulation exists, the null hypothesis for data analysis is not the correct null hypothesis.} 
The entire confidence interval (CI) calculation for an observed dataset is conditional on statistical assumptions about measurement, distributions, asymptotics, and modeling, among others. Changes in any of these can greatly impact the resulting substantive conclusions, a problem known as \emph{model dependence} \citep{KinZen06,IacKinPor11}. The null hypothesis used for the analysis depends on the processed data and thus is subject to model dependence. Let us say we want to know whether a pharmaceutical drug causes a side effect. We might process data by choosing covariates, choosing a match assignment, performing regression with a choice of regularization, and so on. The ``true" null hypothesis is that the drug does not have any side effect. Instead, the null hypothesis that is actually tested is that the drug has no effect after the researcher's pre-processing is done to future instances of raw data. It is not clear which pre-processing steps will make the researcher's null hypothesis close to the true null hypothesis on the correct superpopulation. If the researcher's results are robust to a range of possible data processing options, then this range may include processing that brings the data closer to a sample drawn from the true superpopulation. To analyze the data in this case, we would want a combination of a hacking interval (for the data processing choices) and a regular confidence interval (for the processed data) to ensure robustness both to user manipulation and to randomness in the sample of data. We discuss such combinations in Section \ref{sec:hacking_plus_data} for regression.
To summarize, hacking intervals help to ensure that the conclusions about the true null hypothesis with respect to the true superpopulation are valid.\\

\textit{It does not make sense to explicitly model analyst choices.}
In the case of Bayesian model averaging or other decision-making frameworks, one might try to model the way the analyst might treat the data and average over realistic choices an analyst might make. However, this makes little sense. The hypothesis is about the ground truth, not about researcher choices. We would like the result to be robust to \textit{any} choices made by a reasonable researcher. 

The example of matching, discussed earlier, is an example where placing a prior on analyst choices of matching method does not make sense.

\subsubsection*{Enumerative approaches similar to prescriptively-constrained hacking intervals}
In the social sciences, there are several works that propose enumerating all reasonable model specifications and computing the effect estimate of interest for each specification. The ``extreme bound analysis'' of \citet{leamer1983} focuses on covariate combinations, the ``specification curve'' of \citet{simonsohn2015} proposes a graphical display of all effect estimates and a method for conducting joint inference across all specifications; and the method of \citet{young2017} investigates a variety of model specification types, including functional form, and develops a model influence analysis showing how each model component impacts the effect estimate. Each of these approaches proposes brute force calculation of all chosen model specifications, which can be costly. \citet{munoz2018} draws on the framework of \citet{young2017} to a simulated dataset, fitting a total of 9 billion linear regression models on a simulated dataset, but the computation takes several months. Tethered hacking intervals differ in that they aim only to identify the smallest and largest effect estimates, which permits an optimization-based approach. We also provide a variety of examples for machine learning models in addition to linear models, whereas the approaches mentioned above only consider linear or generalized linear models. 

\subsubsection*{Mathematical equivalence of hacking intervals to other problems, but with different meaning}
In some contexts, hacking intervals bear mathematical equivalence to other problems, which means we can leverage existing methods in some cases. Prescriptively-constrained hacking intervals often fall under a form of sensitivity analysis \citep{Leamer10}. If we consider uncertainty in the inputs to a mathematical model (usually in an applied-math context), they fall under the field of Uncertainty Quantification. If we consider uncertainty in prior specification, they fall under Robust Bayesian Analysis. If we consider uncertainty in assumptions for causal inference, they fall under (causal) Sensitivity Analysis. See \citet{ghanem}, \citet{berger1994}, and \citet{liu} for overviews of these fields, respectively. Uncertainty Quantification provides useful computational tools, like Monte-Carlo simulation and surrogate models \citep{sudret}. In the latter two methods, theoretical bounds on effect estimates have been proven. \citet{berger1990} determines the range of a posterior quantity for priors contained in a certain class. In linear regression, DFBETAS and DFFITS measure the change in a coefficient and a prediction of a linear model, respectively, when a single observation is removed and can be computed without refitting the model \citep{belsley}. These results can be used to compute tethered hacking intervals for linear models when the space of data-adjustment functions includes those that remove a single observation. In causal inference, we can find the range of effect estimates subject to an unmeasured confounder being within specified bounds on its relationship to both the treatment and the outcome \citep{lin, Vanderweele}. If we think of an unmeasured confounder as an additional feature created by a researcher, we can use these results to find the prescriptively-constrained hacking interval under this researcher degree of freedom. We applied this idea in Section \ref{sec:ex_feat}. Tethered hacking intervals are equivalent to profile likelihood confidence intervals \citep{Bjornstad1990} when the loss function corresponds to a likelihood. We discuss this in more detail in 
\textbf{Appendix C.1}.

Finding hacking intervals can be viewed as a form of robust optimization. Robust optimization serves as a worst case analysis in decision theory. Uncertainty sets are the primitives for hacking intervals, namely the ranges of user choices we are willing to consider. In prescriptively-constrained hacking intervals, the uncertainty set is the range of prescriptive choices the researcher is allowed to make. In tethered hacking intervals, the uncertainty set is determined by the set of functions achieving low loss. If we cannot easily determine the uncertainty set in advance, we may be able to learn the uncertainty sets from related problems if data (from other sources) are available. This is done by
\citet{TulabandhulaRuISAIM14RO} for machine learning to determine uncertainty sets for decision making.

The ``Machine Learning with Operational Costs'' framework \citep{TulabandhulaRu14MLJ,TulabandhulaRu13} computes a tethered hacking interval  of the cost that a company might incur to enact an optimal policy in response to any good predictive model. The work of \citet{LethamLeRuBr16} uses tethered hacking intervals in the setting of uncertainty quantification and optimal experimental design for dynamical systems. They recommend to perform an experiment that would most reduce the hacking interval on the quantity the experimenter wishes to estimate.

\subsubsection*{Teaching of hacking intervals}
A major benefit of hacking intervals is that they are easy to explain. Confidence intervals and $p$-values are difficult to teach and interpret, and are regularly misinterpreted. In response, the American Statistical Association recently issued a document explaining hypothesis testing to users \citep{Wasserstein2016}, and the field of Basic and Applied Social Psychology banned $p$-values altogether \citep{Trafimow2015}, but as the authors of these proposals recognize, this does not fully solve the problem. 

Hacking intervals are easy to explain, do not require knowledge of probability to understand, and sometimes capture as much, if not more, uncertainty as regular confidence intervals. Teaching hacking intervals first may give a gentle introduction to the effect of uncertainty on conclusions.




\section{Conclusion}
In this work, we presented an alternative theory of inference. It complements existing theories in that it handles a form of uncertainty that arises from analyst choices, rather than from randomness in the data. We presented several examples of hacking intervals stemming from regression and classification, as well as dimension reduction and feature selection. 
We showed in a real example how hacking intervals can be helpful --- in particular, our results indicate that a commonly used model for pre-trial risk analysis may sometimes be miscalculated, potentially leading to suboptimal judicial decision-making throughout the U.S. Our examples indicate that it is possible that these incorrectly computed risk scores could lead (or have led) to high-risk individuals being released or low-risk individuals being detained. 

\newpage
\begin{appendix}
	\section{\texttt{R} Package} 
\let\clearpage\relax 
\label{sec:rpackage}

This section describes the \texttt{R} package that accompanies this paper. The package computes tethered and prescriptively-constrained hacking intervals for linear models relative to a linear ``base'' model. The package also computes an interval that considers both types of hacking at once; in other words, the range of results that could be achieved by a model that fits the data almost as well as any model obtainable via a modification (within the prescriptive constraints) of the base model. 

The prescriptive constraints are defined as exactly one of any of the following manipulations of the base model:
\begin{itemize}
	\item Removing an outlier.
	\item Removing any additive term in the model.
	\item Adding any variable included in the dataset but excluded from the model. 
	\item Adding an interaction term between existing linear terms in the model. 	
	\item Adding a transformation of a feature (this package implements \verb!x^2! and discretizing continuous variables into indicator variables based on quantiles). 
\end{itemize}
The loss tolerated for tethered hacking is expressed as a percentage of the base model.

Recall that tethered hacking intervals find the range of results over the set of models that fit the data well compared to the base model, supposing that each such model could be obtained by an unidentified manipulation. This provides robustness to \textit{any} (sufficiently small) manipulation, including ones difficult for optimization like:
\begin{itemize}
	\item Changing the values of covariates or outcomes. 
	\item Removing any number of observations.
	\item Adding new observations or new features.
	\item Changing hyperparameters.  
\end{itemize}
Next we walk through a brief tutorial of the package. 
\vspace{.5cm}

\subsection*{Installation}
\vspace{.5cm}
\begin{Verbatim}[frame=single]
devtools::install_github("beauCoker/hacking")
\end{Verbatim}

\subsection*{Quick demo}
Start with a dataset. For this demo we'll generate a toy dataset \texttt{data}.
\vspace{.5cm}

\begin{Verbatim}[frame=single]
set.seed(0)

N = 50 # Number of observations
data <- data.frame(
y = rnorm(N), # Response variable (continuous)
w = rbinom(N, 1, .5), # Treatment variable (binary)
X = matrix(rnorm(N*3), nrow=N), # Covariates included in base model
Z = matrix(rnorm(N*3), nrow=N) # Covariates excluded from base model
)
\end{Verbatim}

Next, fit a linear model with \texttt{lm}. We'll call this the ``base'' model.
\vspace{.5cm}

\begin{Verbatim}[frame=single]
mdl <- lm(y ~ w + X.1*X.2, data=data)
(beta_0 <- mdl$coefficients['w'])

#>         w 
#> 0.2696223
\end{Verbatim}
\vspace{.5cm}

This tells us the ordinary least squares estimate for the coefficient \texttt{beta\_0} on the treatment variable $w$ in the base model is about 0.27. A standard question in statistics is to ask, ``What could happen if I estimated \texttt{beta\_0} using a different dataset drawn from the same distribution?'' This is conceptually what a standard confidence interval tells you. It can be computed with \texttt{R}'s built-in \texttt{confint} function:
\vspace{.5cm}

\begin{Verbatim}[frame=single]
(ci <- confint(mdl)['w',])

#>      2.5 %     97.5 % 
#> -0.2619048  0.8011494
\end{Verbatim}

Now we get to the hacking interval part. What if instead you ask, ``What if the scientist that reported this estimate threw out some important observations, or manipulated the data in some other way? What's the range of estimates that could have been reported?'' This is conceptually what a hacking interval tells you. For linear models, it can be computed with the \texttt{hackint\_lm}  function in our package. The parameter \texttt{theta} tells you what percentage of loss is tolerated for the tethered variety of hacking.
\vspace{.5cm}

\begin{Verbatim}[frame=single]
library(hacking)
output <- hackint_lm(mdl, data, theta=0.1, treatment = 'w')

#>                       result      value                     manipulation
#> 1             Tethered (LB): -0.2901996                         Tethered
#> 2             Tethered (UB):  0.8294442                         Tethered
#> 3          Constrained (LB):  0.1675830            Remove observation 29
#> 4          Constrained (UB):  0.3508530            Remove observation 13
#> 5 Constrained+Tethered (LB): -0.4013983 Remove observation 29 + Tethered
#> 6 Constrained+Tethered (UB):  0.9234986      Add variable Z.2 + Tethered
\end{Verbatim}

In the output above, \texttt{LB} and \texttt{UB} stand for the lower bound and upper bound, respectively, of the corresponding hacking interval. It says that a tethered hacking interval around the base model is $(-0.29, 0.83)$, a prescriptively constrained hacking interval around the base model is $(0.17, 0.35)$, and a hacking interval that considers both types hacking is $(-0.4, 0.92)$. Notice either of the tethered intervals are wider than the standard confidence interval, $(-0.26, 0.8)$, but note that hacking intervals and standard confidence invervals measure different forms of uncertainty. Either could be larger, and hacking intervals need not even be centered on the point estimate.

The \texttt{hackint\_lm} function works by enumerating all of the manipulations within the prescriptive constraints and, for each manipulation, computing the ordinary least squares coefficient estimate as well as a tethered hacking interval around this estimate (i.e., where the model under the manipulation is essentially treated as a new base model). This complete list is available as a dataframe, with \texttt{Estimate} denoting the coefficient estimate and \texttt{(LB,UB)} denoting the tethered hacking interval. The prescriptively-constrained hacking interval is the range of \texttt{Estimate}. The hacking interval that considers prescriptive constraints and tethering is given by the minimum of \texttt{LB} and the maximum of \texttt{UB}. This list is useful for diagnosing which manipulations are most impactful. The output is sorted by the largest absolute difference \texttt{largest\_diff} of any value (\texttt{LB}, \texttt{Estimate}, or \texttt{UB}) from \texttt{beta\_0}:
\vspace{.5cm}

\begin{Verbatim}[frame=single]
output$hacks_all

#> # A tibble: 62 x 6
#>    manipulation          type           LB Estimate    UB largest_diff
#>    <chr>                 <chr>       <dbl>    <dbl> <dbl>        <dbl>
#>  1 Remove observation 29 remove_obs -0.401    0.168 0.737        0.671
#>  2 Add variable Z.2      add_term   -0.222    0.351 0.923        0.654
#>  3 Remove observation 35 remove_obs -0.373    0.192 0.757        0.643
#>  4 Remove observation 13 remove_obs -0.200    0.351 0.901        0.632
#>  5 Remove observation 3  remove_obs -0.199    0.350 0.899        0.629
#>  6 Remove observation 30 remove_obs -0.256    0.318 0.892        0.622
#>  7 Remove observation 48 remove_obs -0.227    0.332 0.891        0.622
#>  8 Remove observation 41 remove_obs -0.349    0.199 0.747        0.619
#>  9 Remove observation 32 remove_obs -0.220    0.330 0.881        0.611
#> 10 Remove observation 24 remove_obs -0.241    0.320 0.880        0.610
#> # … with 52 more rows
\end{Verbatim}

\subsection*{Focusing on most influential observations}
The optional argument \verb!frac_remove_obs! (default value 1) specifies the fraction of observations that are considered for removal in evaluating prescriptively-constrained hacking intervals. If \verb!frac_remove_obs! is less than 1, then only observations with the highest Cook's distance \citep{cook} are considered for removal. This will speed up computation for small datasets but does not provide any theoretical guarantees of accuracy.

\section{Prescriptively-Constrained Hacking Intervals for Matching} 
\let\clearpage\relax 
\label{sec:matching}

Matching methods in causal inference are a key example of where a hacking interval can be useful in strengthening or weakening conclusions made from data. An analyst's choice of matching algorithm may have a huge impact on the conclusion, and this impact could be much larger than the uncertainty due to randomness in the data. It is entirely possible that two well-meaning analysts, given the same data, would choose two different matching algorithms for treatment and control units and reach two entirely different conclusions. It does not make sense to model the set of choices that analysts would make when choosing between possible matching algorithms. The analyst's choice could be arbitrary; it depends on the algorithms available at the time, the popularity of these algorithms among scientists, the order of the data in the database, and other choices that are totally separate from the ground truth treatment effect. The analyst does not choose matches from a uniform distribution among reasonable matching possibilities (simple examples where one good match assignment is clearly better than another can show why this would be an unreasonable assumption). In the work of \citet{Noor1}, the authors take a hacking interval approach by specifying that an \textit{unreasonable} match assignment would have at least one matched treatment/control pair whose covariates are far away from each other. The converse of this set consists of  \textit{reasonable} match assignments, even though some of these match assignments would not be chosen by any matching algorithm that we could envision.  \citet{Noor1} compute a prescriptively-constrained hacking interval, in that they compute the range of treatment effect estimates corresponding to all reasonable match assignments. For some datasets, they find that all reasonable match assignments yield the same conclusion. In other cases, the range of treatment effects corresponding to reasonable matches is very large. Their technique uses mixed-integer programming, so that they can determine the maximum and minimum test statistics over all match assignments without having to enumerate them.

\begin{figure}[!h]
	\centering
	\includegraphics[scale=0.25]{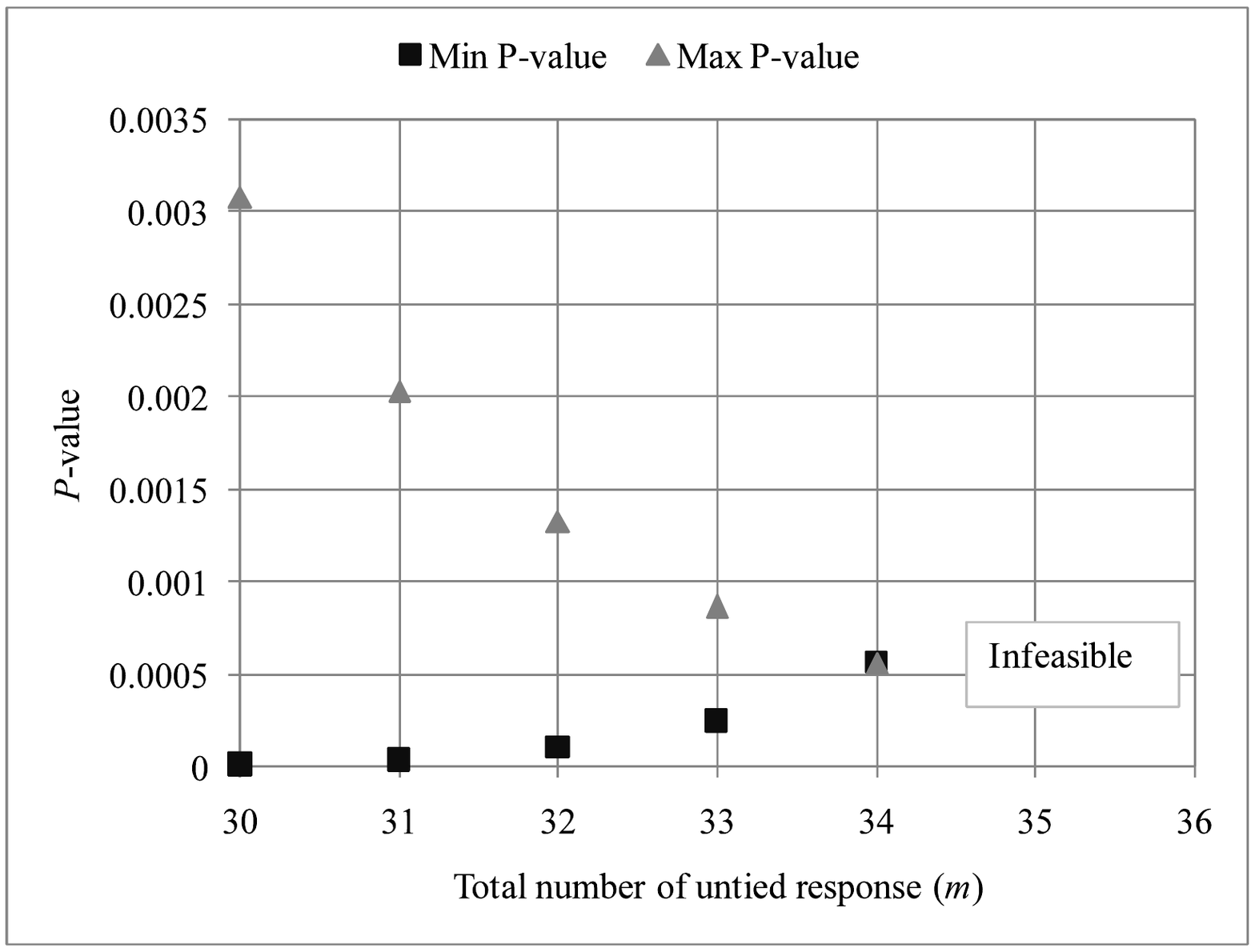}
	\caption{Prescriptively-constrained hacking intervals of the $p$-value of McNemar's test for different numbers of matched groups. Reproduced from Figure 3 in \citet{Noor1}.}
	\label{fig:glow}
\end{figure}

If we find that \textit{any} reasonable match assignment yields the same conclusion, then it is a much stronger result than saying that \textit{one} reasonable match assignment (as is typically considered) yields a particular conclusion. Consider the experiment done by \citet{Noor1} on the GLOW (Global Longitudinal study of Osteoporosis in Women) data \citep{glow}. The goal was to determine whether smoking causes bone fractures using McNemar's test. Their experiments showed that no matter which reasonable analyst creates the matched pairs, the conclusion is the same: smoking causes bone fractures. Figure \ref{fig:glow}, reproduced from \citet{Noor1}, shows the hacking intervals of the $p$-value of McNemar's test for different numbers of matched groups. The figure shows that if any reasonable analyst constructs match assignments with 30 or more ``unmatched'' pairs (where treatment and control outcomes differ), the worst (highest) $p$-value possible they can achieve is 0.003, meaning the result will be significant at the 0.05 level no matter which matching method the analyst uses. This example demonstrates how hacking intervals can sometimes strengthen scientific conclusions. (In other cases, hacking intervals can weaken conclusions).

\section{Tethered Hacking Intervals Supplement} 
\let\clearpage\relax 
\textbf{Appendix \ref{sec:profile}} describes the relationship between tethered hacking intervals and profile likelihood confidence intervals, \textbf{Appendix \ref{sec:bound}} makes an assumption about the true data generating procedure to derive an appropriate generalization bound, and \textbf{Appendix \ref{sec:kernel_reg}} and \textbf{Appendix \ref{sec:pca}} describe two additional examples of tethered hacking intervals: predictions made by kernel regression and features selected using PCA. 

\subsection{Connection to Profile Likelihood}\label{sec:profile}

Tethered hacking intervals are closely related to profile likelihood confidence intervals. When the loss function $L$ corresponds to a likelihood function $\mathcal{L}$ and the test statistic $t$ is a single parameter of the learned prediction function $f$, then a tethered hacking interval is mathematically equivalent to a profile likelihood confidence interval for an appropriate choice of the loss threshold $\theta$. In this section we quantify this equivalence.

Suppose a likelihood function $\mathcal{L}(\lambda,\xi)$ depends on a low-dimensional parameter of interest $\lambda$ and a higher-dimensional nuisance parameter $\xi$. The \textit{profile likelihood} $\mathcal{L}_p$ focuses attention on $\lambda$ by ``profiling out'' $\xi$:
\begin{align*}
\mathcal{L}_p (\lambda) &:= \sup_{\xi} \mathcal{L}(\lambda,\xi) 
= \mathcal{L}(\lambda, \hat{\xi}_{\lambda}),
\end{align*}
where $\hat{\xi}_\lambda$ maximizes the likelihood when $\lambda$ is fixed. The profile likelihood $\mathcal{L}_p$ is a lower dimensional function that for many purposes can be used instead of the full, higher-dimensional likelihood $\mathcal{L}$. Notably, $[\hat{\lambda}, \hat{\xi}]$ maximizes the full likelihood if any only if $\hat{\lambda}$ maximizes the profile likelihood. Interesting for our purposes is the property that the ratio of profile likelihoods $\mathcal{L}_p(\lambda_0)/ \mathcal{L}_p(\hat{\lambda})$ equals the ratio of likelihoods $\Lambda_{\lambda_0}$ for testing the null hypothesis $H_0: \lambda=\lambda_0$:
\begin{align*}
\frac{\mathcal{L}_p(\lambda_0)}{\mathcal{L}_p(\hat{\lambda})} 
= \frac{\sup_\xi \mathcal{L} (\lambda_0, \xi)}{\sup_\xi \mathcal{L} (\hat{\lambda}, \xi)} 
= \frac{\sup_\xi \mathcal{L} (\lambda_0, \xi)}{\sup_{\lambda,\xi} \mathcal{L} (\lambda, \xi)} := \Lambda_{\lambda_0}.
\end{align*}
By Wilks' Theorem \citep{wilks}, if $H_0$ is true and a few regularity conditions are met, then as the sample size $n\to\infty$,
\begin{align*}
-2\log{\Lambda}_{\lambda_0} \xrightarrow{d} \chi^2_m,
\end{align*}
where $m$ is the difference in the dimensions of $\lambda$ and $\xi$. A hypothesis test for $H_0$ rejects $H_0$ if $-2\log{\Lambda_{\lambda_0}}$ is large, which happens when the maximum likelihood under $H_0$, $\sup_\xi \mathcal{L} (\lambda_0, \xi)$, is small.
If $\lambda$ is a scalar, the set of $\lambda_0$ for which $H_0:\lambda=\lambda_0$ is not rejected provides a confidence interval for $\lambda$. That is, a $1-\alpha$ \textit{profile likelihood likelihood confidence interval} for $\lambda$ is given by $[p_{\min}, p_{\max}]$, where:
\begin{align*}
p_{\min} &:= \min \lambda \quad\text{s.t.}\quad -2\log(\Lambda_{\lambda}) \le \chi^2_{m, 1-\alpha}, \\
p_{\max} &:= \max \lambda \quad\text{s.t.}\quad -2\log(\Lambda_{\lambda}) \le \chi^2_{m, 1-\alpha},
\end{align*}
and $\chi^2_{m, 1-\alpha}$ is the $1-\alpha$ quantile of a $\chi^2$ distribution with $m$ degrees of freedom. Equivalently, in terms of the profile likelihood we have:
\begin{align}
p_{\min} &= \min \lambda \quad\text{s.t.}\quad \log \mathcal{L}_p(\lambda) \ge \log \mathcal{L}_p(\hat{\lambda}) - \frac{1}{2} \chi^2_{m, 1-\alpha}, \label{eq:pl_ci_min}\\
p_{\min} &= \max \lambda \quad\text{s.t.}\quad \log \mathcal{L}_p(\lambda) \ge \log \mathcal{L}_p(\hat{\lambda}) - \frac{1}{2} \chi^2_{m, 1-\alpha} \label{eq:pl_ci_max}.
\end{align}
If we define $\theta_p(\alpha):= \log \mathcal{L}_p(\hat{\lambda}) - \frac{1}{2} \chi^2_{m, 1-\alpha}$ then we have:
\begin{align}
p_{\min} &= \min \lambda \quad\text{s.t.}\quad \log \mathcal{L}_p(\lambda) \ge \theta_p(\alpha), \label{eq:pl_ci_min2}\\
p_{\min} &= \max \lambda \quad\text{s.t.}\quad \log \mathcal{L}_p(\lambda) \ge \theta_p(\alpha) \label{eq:pl_ci_max2}.
\end{align}
Equations (\ref{eq:pl_ci_min2}) and (\ref{eq:pl_ci_max2}) are similar to Equations  (\ref{eq:tethering_min}) and (\ref{eq:tethering_max}) that define tethered hacking intervals if the summary statistic $t$ is a single parameter $\lambda$ of the prediction function $f_{\lambda,\xi}$; that is, if $t(f_{\lambda,\xi})=\lambda$. In this case, both the profile likelihood confidence interval and the tethered hacking interval are the minimum and maximum of the summary statistic that can be achieved subject to a constraint on how well the prediction function fits the observed data. In the case of a profile likelihood confidence interval, the fit constraint is a lower bound $\theta_p(\alpha) := \log \mathcal{L}_p(\hat{\lambda}) - \frac{1}{2} \chi^2_{m, 1-\alpha}$ on the profile likelihood. In the case of a tethered hacking interval, the fit constraint is an upper bound $\theta$ on the loss. To summarize, if $t(f_{\lambda,\xi})=\lambda$, then:
\begin{align*}
\text{Profile likelihood interval:}&\quad \underset{\lambda}{\text{max/min}}\ \lambda \ \text{s.t.}\  \log \mathcal{L}_p(\lambda) \ge \theta_p(\alpha),
\\ \text{Tethered hacking interval:}&\quad \underset{\lambda,\xi}{\text{max/min}}\ \lambda \ \text{s.t.}\  L(Z, f_{\lambda,\xi},\psi_d) \le \theta.
\end{align*}
Notice that the profile likelihood confidence interval is an optimization over $\lambda$ only, while the tethered hacking interval is an optimization over $\lambda$ and $\xi$. However, since the objective function of the tethered hacking interval does not depend on $\xi$, and the loss is constrained by an upper bound, we can do no better in the optimization than by plugging in the $\xi$ that minimizes the loss for a fixed $\lambda$, $\hat{\xi}_\lambda := \text{argmin}_\xi L(Z, f_{\lambda,\xi},\psi_d)$, into the tethered hacking interval constraint. In other words, we can ``profile out'' the nuisance parameter $\xi$ from the loss function as we did with the likelihood function. Therefore, we have:
\begin{align}
\text{Profile likelihood interval:}&\quad \underset{\lambda}{\text{max/min}}\ \lambda \ \text{s.t.}\  \log \mathcal{L}_p(\lambda) \ge \theta_p(\alpha) \label{eq:ppl}\\
\text{Tethered hacking interval:}&\quad \underset{\lambda}{\text{max/min}}\ \lambda \ \text{s.t.}\  L(Z, f_{\lambda,\hat{\xi}_\lambda},\psi_d) \le \theta. \label{eq:hack}
\end{align}
This shows the equivalence of a profile likelihood confidence interval and a tethered hacking interval when the summary statistic is a single parameter $\lambda$ of the prediction function. Notice the profile likelihood confidence interval requires the existence of a likelihood model, whereas the tethered hacking interval requires the existence of only a loss function.

If we are given a likelihood function $\mathcal{L}(\lambda,\xi)$ and threshold $\theta_p(\alpha)$ that define a profile likelihood confidence interval for the parameter $\lambda$ at confidence level $\alpha$, we can construct a loss function $L(\Z, f_{\lambda,\xi}, \psi_d)$ and loss threshold $\theta$ that define an equivalent tethered hacking interval. We do this by defining the loss function as the negative log likelihood and the loss threshold as the negative likelihood threshold:
\begin{align}
L(\Z, f_{\lambda,\xi}, \psi_d) &:= -\log \mathcal{L}(\lambda,\xi),\label{eq:like2loss} \\
\theta&:=-\theta_p(\alpha).\label{eq:like2loss_theta}
\end{align}
We assume the function class $\mathcal{F}$ is clear from the definition of the likelihood.
Taking the infimum over $\xi$ of Equation (\ref{eq:like2loss}) we have:
\begin{align*}
\inf_\xi L(\Z, f_{\lambda,\xi}, \psi_d) &= \inf_\xi -\log \mathcal{L}(\lambda,\xi) \\
\iff L(\Z, f_{\lambda,\hat{\xi}_\lambda}, \psi_d) &= -\sup_\xi \log \mathcal{L}(\lambda,\xi) \\
&= - \log \sup_\xi \mathcal{L}(\lambda,\xi) \\
&= - \log \mathcal{L}_p(\lambda).
\end{align*}
This means:
\begin{align*}
\left\{\lambda \mid \log \mathcal{L}_p(\lambda) \ge \theta_p(\alpha) \right\}
&= \left\{\lambda \mid -\log \mathcal{L}_p(\lambda) \le -\theta_p(\alpha) \right\} \\
&= \left\{\lambda \mid L(\Z, f_{\lambda,\hat{\xi}_\lambda}, \psi_d) \le \theta \right\}.
\end{align*}
The profile likelihood confidence interval and tethered hacking interval given in Equations (\ref{eq:ppl}) and (\ref{eq:hack}), respectively, will therefore be the same, since each is defined by the minimum and maximum value of the same objective function over the same set.

We illustrate the construction of a tethered hacking interval from a profile likelihood confidence interval with the example of linear regression. Suppose the outcomes $\Y=(y_1,\dotsc,y_n)^T$ are generated by the following linear model with independent Gaussian noise of known variance $\sigma^2$:
$$
\Y \sim N(\X \bxi +  \W \lambda , \sigma^2 \bm{I}),
$$
where $\X = (\x_1,\dotsc,\x_n^T)^T$ and $\W=(w_1,\dotsc,w_n)^T$ are, respectively, $n\times p$ and $n\times 1$ matrices of covariates, $\bxi$ is a $p\times 1$ vector of nuisance parameters, and $\lambda$ is the scalar parameter of interest. The log likelihood for this model is:
\begin{align}
\log \mathcal{L}(\bxi, \lambda_0) 
&= \log \prod_{i=1}^n (2\pi\sigma^2)^{-1/2} \exp\left\{ -\frac{1}{2\sigma^2}\left( \x_i^T \bxi + w_i\lambda - y_i \right)^2 \right\}\nonumber\\
&= -\frac{n}{2}\log(2\pi\sigma^2) - \frac{1}{2\sigma^2}\sum_{i=1}^n \left( \x_i^T \bxi + w_i\lambda - y_i \right)^2. \label{eq:lin_like}
\end{align}
By Equation (\ref{eq:ppl}), the profile likelihood confidence interval at confidence level $\alpha$ is defined by the minimum and maximum values of $\lambda$ for which the following inequality holds:
\begin{align}
\log \mathcal{L}_p(\lambda) &\ge \theta_p(\alpha) \nonumber
\\
\log \mathcal{L}_p(\lambda) 
&\ge \log \mathcal{L}_p(\hat{\lambda}) - \frac{1}{2} \chi^2_{1, 1-\alpha} \nonumber
\\
-\frac{n}{2}\log(2\pi\sigma^2) 
- \frac{1}{2\sigma^2}\sum_{i=1}^n \left( \x_i^T \hat{\bxi}_{\lambda} + w_i \lambda - y_i \right)^2 \nonumber
&\ge
-\frac{n}{2}\log(2\pi\sigma^2) 
- \frac{1}{2\sigma^2}\sum_{i=1}^n \left( \x_i^T \hat{\bxi} + w_i\hat{\lambda} - y_i \right)^2
- \frac{1}{2} \chi^2_{1, 1-\alpha} \nonumber
\\
\sum_{i=1}^n \left( \x_i^T \hat{\bxi}_{\lambda} + w_i \lambda - y_i \right)^2
&\le
\sum_{i=1}^n \left( \x_i^T \hat{\bxi} + w_i\hat{\lambda} - y_i \right)^2
+ \sigma^2 \chi^2_{1, 1-\alpha}.
\label{eq:lambda_ppl}
\end{align}
To construct an equivalent tethered hacking interval, we define the loss by Equation (\ref{eq:like2loss}): 
\begin{align*}
L(\Z, f_{\xi,\lambda}, \psi_d) &:= - \log \mathcal{L}(\bxi, \lambda) \\
&=\frac{n}{2}\log(2\pi\sigma^2) + \frac{1}{2\sigma^2}\sum_{i=1}^n \left( \x_i^T \bxi + w_i\lambda - y_i \right)^2
\end{align*}
and the loss threshold by Equation (\ref{eq:like2loss_theta}):
\begin{align*}
\theta &:= - \theta_p(\alpha) \\
&=- \log \mathcal{L}_p(\hat{\lambda}) + \frac{1}{2} \chi^2_{m, 1-\alpha} \\
&= \frac{n}{2}\log(2\pi\sigma^2) 
+ \frac{1}{2\sigma^2}\sum_{i=1}^n \left( \x_i^T \hat{\bxi} + w_i\hat{\lambda} - y_i \right)^2
+ \frac{1}{2} \chi^2_{1, 1-\alpha}.
\end{align*}
By Equation (\ref{eq:hack}), the tethered hacking interval is defined by the minimum and maximum values of $\lambda$ for which the following inequality holds:
\begin{align}
L(Z, f_{\lambda,\hat{\xi}_\lambda},\psi_d) &\le \theta \nonumber
\\
\frac{n}{2}\log(2\pi\sigma^2) + \frac{1}{2\sigma^2}\sum_{i=1}^n \left( \x_i^T \hat{\bxi} + w_i\lambda - y_i \right)^2
&\le
\frac{n}{2}\log(2\pi\sigma^2) 
+ \frac{1}{2\sigma^2}\sum_{i=1}^n \left( \x_i^T \hat{\bxi} + w_i\hat{\lambda} - y_i \right)^2
+ \frac{1}{2} \chi^2_{1, 1-\alpha} \nonumber
\\
\sum_{i=1}^n \left( \x_i^T \hat{\bxi}_{\lambda} + w_i \lambda - y_i \right)^2 
&\le
\sum_{i=1}^n \left( \x_i^T \hat{\bxi} + w_i\hat{\lambda} - y_i \right)^2
+ \sigma^2 \chi^2_{1, 1-\alpha}. \label{eq:lambda_hack}
\end{align}
Since Equations (\ref{eq:lambda_ppl}) and (\ref{eq:lambda_hack}) are the same, the profile likelihood confidence interval and tethered hacking intervals are the same, since each is defined by the minimum and maximum value of $\lambda$  over the same set of $\lambda$. Notice any monotonic function applied to the loss function and loss threshold will yield the same set of $\lambda$ for which the loss function exceeds the loss threshold.

\subsection{Generalization Bound}\label{sec:bound}

Tethered hacking intervals do not require the assumption of a true data generating process; however, if we do assume a true data generating process $\mu$, we can extend the interpretation of tethered hacking intervals to this setting and, in the case of learning classification functions, derive a generalization bound that incorporates both the uncertainty due to observing a random draw from the true data generating process \textit{and} the uncertainty due to researcher hacking. The interpretation of a hacking interval in this setting begins with the following sequence of events:
\begin{enumerate}
	\item The true distribution $\mu$ generates independently and identically a ``pristine'' dataset $\Zp$ of $n$ observations.
	
	\item $\tilde{\phi}: \mathcal{Z} \to \mathcal{Z}$ transforms the pristine data, $\Zp$, into what the researcher actually observes, $\Zo$.
	
	\item $\phi: \mathcal{Z} \to \mathcal{Z}$ transforms the observed data, $\Zo$, into the hacked data the researcher uses in their analysis, $\Zh$. This is the same $\phi$ as in the tethered hacking interval definition.\label{step3}
\end{enumerate}
In other words, in order to do inference or prediction on the true data generating process $\mu$, a researcher would like to perform their analysis on a dataset generated by $\mu$, which we call the pristine data $\Zp$. However, we assume the researcher only observes $\Zo$, which is the pristine data after applying an unknown data adjustment function $\tilde{\phi}$. In an attempt to undo $\tilde{\phi}$, the researcher chooses a data adjustment function of their own, $\phi$. The result of applying $\phi$ to $\Zo$ is the hacked data $\Zh$.
Schematically, we can write this procedure as:
\begin{align*}
\underbrace{\Zp}_{\substack{\text{Pristine} \\ \text{data}}}
\xrightarrow[]{\tilde{\phi}} 
\underbrace{\Zo}_{\substack{\text{Observed} \\ \text{data}}}
\xrightarrow[]{\phi}
\underbrace{\Zh}_{\substack{\text{Hacked} \\ \text{data}}}.
\end{align*}

On each of the three datasets --- $\Zp$, $\Zo$, and $\Zh$ --- there is at least one function from a function class $\mathcal{F}$ that minimizes the empirical risk on a loss function $L(\Z,f)$.\footnote{Both the function class $\mathcal{F}$ and the loss function $L(\Z,f)$ can depend on hyperparameters $\psi_d$, but hyperparameters are not important to this section so we suppress their notation.} 
We call these functions $\fp$, $\fo$, and $\fh$, respectively. That is, we define the following:
\begin{itemize}
	
	\item $\fp\in \arg\min_{f\in\mathcal{F}} R^{\emp}_{\Zp}(f) :=  \arg\min_{f\in\mathcal{F}} L(\Zp, f)$ minimizes the empirical risk of the pristine dataset, $\Zp$. Since we do not observe $\Zp$ we cannot learn it.
	
	\item $\fo \in \arg\min_{f\in\mathcal{F}} R^{\emp}_{\Zo}(f) :=  \arg\min_{f\in\mathcal{F}} L(\Zo, f)$ minimizes the empirical risk of the observed dataset, $\Zo$. This is the function that would be learned if we did not allow for hacking.
	
	\item $\fh \in \arg\min_{f\in\mathcal{F}} R^{\emp}_{\Zh}(f) :=  \arg\min_{f\in\mathcal{F}} L(\Zh, f)$ minimizes the empirical risk of the hacked dataset, $\Zh$. It is learned by the researcher.
\end{itemize}
In a classical statistical setting --- where the researcher observes $\Zp$ and computes $\fp$ --- a question from statistical learning theory is how the true risk of $\fp$, $R^{\true}_\mu (\fp) := \mathbb{E}_{Z\sim\mu} L(Z,\fp)$, differs from the empirical risk of $\fp$ on $\Zp$, $R^{\emp}_{\Zp}(\fp) := L(\Zp, \fp)$. If $\mathcal{F}$ consists of classification functions, a bound on this difference can be found using statistical learning theory. In a hacking setting --- where the researcher observes $\Zo$, adjusts the data to obtain $\Zh$, and computes $\fh$ --- there is additional uncertainty due to the data adjustments. In order to derive an analogous bound, this time to understand how the true risk of $\fp$ differs from the empirical risk of $\fh$ on $\Zh$, we will need to make a few assumptions about the impact of the data adjustments:
\begin{itemize}
	\item $\left| R^{\emp}_{\Zp} (\fp) - R^{\emp}_{\Zo} (\fp)  \right| \le \theta_1$, which we call ``reverse tethering (part 1).'' It means that the function $\fp$ that minimizes the loss on the pristine data does not yield a loss too different on the observed data.
	
	\item $\left| R^{\emp}_{\Zo} (\fp) - R^{\emp}_{\Zo} (\fo)  \right| \le \theta_2$, which we call ``reverse tethering (part 2).'' It means that the functions learned from the pristine data and the observed data do not have losses too different from each other on the observed data.
	
	\item $\left| R^{\emp}_{\Zo} (\fo) - R^{\emp}_{\Zo} (\fh)  \right| \le \theta_3$, which is our standard tethering constraint used in Equations (\ref{eq:tethering_min}) and (\ref{eq:tethering_max}). It means that the functions learned from the observed data and the hacked data do not have losses too different from each other on the observed data.
	
	\item Let $\left| R^{\emp}_{\Zo} (\fh) - R^{\emp}_{\Zh} (\fh)  \right| = \theta_4$. This is not an assumption since it can be calculated by the researcher. For the function $\fh$ that minimizes the loss on the hacked data, it is the difference between the loss on the observed data and the hacked data.
	
\end{itemize}
We can bound the difference between the true risk of $\fh$ and the empirical risk of $\fh$ on $\Zh$ by applying the triangle inequality several times and bounding each intermediate difference. The assumptions about $\theta_1$, $\theta_2$, and $\theta_3$, and the calculated $\theta_4$, provide bounds on all but one of these intermediate differences. The final intermediate difference is between the true risk of $\fh$ and the empirical risk of $\fh$ on $\fp$. This can be bounded by the same learning theory bound we would derive in a classical statistical setting, since it holds uniformly for all functions in $\mathcal{F}$. Table \ref{tbl:gen_bound} summarizes the relationships and Proposition \ref{thm_gen_bound} gives the result. The proof is available in \textbf{Appendix \ref{sec:proofs}}.

\begin{prop}[Generalization Bound for Hacked Data]
	\label{thm_gen_bound}
	
	If $\mathcal{F}$ is a set of classification functions with Vapnik-Chervonenkis dimension $h$, then, for all $\delta>0$, with probability of at least $1-\delta$ with respect to data $\Zp$ drawn \textit{i.i.d.} from an unknown distribution $\mu$ on $\R^{n\times p}\times\{-1,1\}^n$:
	\begin{equation*}
	\left| R^{\true}_{\mu} (\fp) - R^{\emp}_{\Zh} (\fh)  \right| \le  2\sqrt{2\frac{h\log{\frac{2eh}{n}}+\log{\frac{4}{\delta}}}{n}} + \sum_{i=1}^4 \theta_i.
	\end{equation*}
\end{prop}

\begin{table}[]
	\centering
	\def\arraystretch{1.5}
	\begin{tabular}{cccccccc}
		& $\mu$ & &$\Zp$ & & $\Zo$ & & $\Zh$ \\ \cline{2-8} 
		\multicolumn{1}{c|}{$\fp$} & $R^{\true}_{Z_\mu}(\fp)$ & $\xrightarrow{{\scriptstyle VC}}$ &$R^{\emp}_{\Zp}(\fp)$ & $\xrightarrow{{\scriptstyle \theta_1}}$ &$R^{\emp}_{\Zo}(\fp)$ &  &$R^{\emp}_{\Zh}(\fp)$ \\
		\multicolumn{1}{c|}{} &  &  & &  & $\left\downarrow\rule{0cm}{.3cm}\right.{\scriptstyle \theta_2}$ &  &  \\
		\multicolumn{1}{c|}{$\fo$} & $R^{\true}_{Z_\mu}(\fo)$ &  &$R^{\emp}_{\Zp}(\fo)$ &  &$R^{\emp}_{\Zo}(\fo)$ &  & $R^{\emp}_{\Zh}(\fo)$ \\
		\multicolumn{1}{c|}{} &  &   &  &  & $\left\downarrow\rule{0cm}{.3cm}\right.{\scriptstyle \theta_3}$  &  &  \\
		\multicolumn{1}{c|}{$\fh$} & $R^{\true}_{Z_\mu}(\fh)$ & &$R^{\emp}_{\Zp}(\fh)$ &  & $R^{\emp}_{\Zo}(\fh)$ & $\xrightarrow{{\scriptstyle \theta_4}}$ & $R^{\emp}_{\Zh}(\fh)$
	\end{tabular}
	\caption{We can relate the true risk of $\fh$, $R^{\true}_{Z_\mu}(\fp)$, to the empirical risk of $\fh$ on $\Zh$, $R^{\emp}_{\Zh}(\fh)$, by applying a VC bound and the triangle inequality four times.}
	\label{tbl:gen_bound}
\end{table}

\subsection{Kernel Regression}\label{sec:kernel_reg}
\begin{figure}[!h]
	\centering
	\includegraphics[scale=0.35]{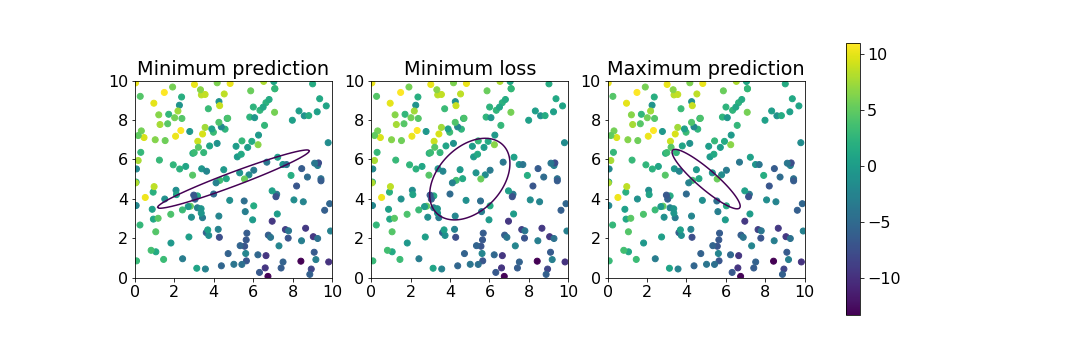}
	\caption{Level sets of the Mahalanobis distance that defines the kernel regression function $f_A$. The middle panel corresponds to the minimum loss prediction function, while the left and right panels correspond to the prediction functions that minimize and maximize prediction on $\xnew=(5,5)^T$, respectively, within a loss constraint $\theta=2000$ and for $\psi_d=1$.}
	\label{fig:kernel_reg}
\end{figure}

Consider the form of kernel regression where prediction models $f\in\mathcal{F}$ evaluated at a point $\x_i\in\mathbb{R}^p$ are weighted averages of the labels of the other points, $\{y_j\}_{j\neq i}$, and where the weight is determined by a kernel function $k_{\psi_d}(d_A(\x_i,\x_j))$ that depends inversely on hyperparameters $\psi_d$ and the distance $d_A(\x_i,\x_j)$ between points $\x_i$ and $\x_j$ for parameters $A$. That is, $f$ is of the form:
$$
f_A(\x_i) = \frac
{\sum_{j\neq i} y_i k_{\psi_d}\left(d_A(\x_o,\x_i)\right)}
{\sum_{j\neq i} k_{\psi_d}\left(d_A(\x_i,\x_i)\right)}.
$$
We suppose a quadratic loss function $L\left(\Z,f_A,\psi_d\right) = \sum_i (f_A(\x_i) - y_i)^2$, 
a Gaussian kernel $k_{\psi_d}(d(\x_i,\x_j)) = 1/\sqrt{2\pi \psi_d^2}\exp{\left(-d(\x_i,\x_j)/\psi_d^2\right)}$, and a Mahalanobis distance $d_A(\x_i,\x_j) = (x_i - x_j)^T A^T A (x_i - x_j)$, where $A$ is a $p\times p$ matrix of parameters. Methods for learning $A$ are considered by \citet{Weinberger}. A tethered hacking interval $[b_{\min}, b_{\max}]$ for prediction on a new point $\xnew$ is given by:
\begin{align*}
b_{\min}&=\min_{A} \frac
{\sum_{j} y_j k\left(d_A(\xnew,\x_j)\right)}
{\sum_{j} k\left(d_A(\xnew,\x_j)\right)}
\quad \text{s.t.} \quad
\sum_i \left( y_i -  \frac
{\sum_{j\neq i} y_j k\left(d_A(\x_i,\x_j)\right)}
{\sum_{j\neq i} k\left(d_A(\x_i,\x_j)\right)}
\right)^2 \le \theta
\\
b_{\max}&=\max_{A} \frac
{\sum_{j} y_j k\left(d_A(\xnew,\x_j)\right)}
{\sum_{j} k\left(d_A(\xnew,\x_j)\right)}
\quad \text{s.t.} \quad
\sum_i \left( y_i -  \frac
{\sum_{j\neq i} y_j k\left(d_A(\x_i,\x_j)\right)}
{\sum_{j\neq i} k\left(d_A(\x_i,\x_j)\right)}
\right)^2 \le \theta.
\end{align*}
Figure \ref{fig:kernel_reg} shows an example. Covariates are uniformly distributed on $[0,10]\times[0,10]$ and any points on a line of slope one have the same mean outcome. Therefore, the prediction function $f_A(\x)$ should assign higher weight to the points on the upper right and on the lower left of $\x$ since these points should have outcomes similar to the outcome of $\x$. Figure \ref{fig:kernel_reg} shows this result in the level sets of the Mahalanobis distance metric that defines $f_A$. The middle panel corresponds to the minimum loss prediction function, while the left and right panels correspond to the prediction functions that minimize and maximize prediction on $\xnew=(5,5)^T$, respectively, within a loss constraint $\theta=2000$ and for $\psi_d=1$. These minimum and maximum predictions, which define a tethered hacking interval, are $-0.94$ and $1.17$. The minimum loss prediction is $-0.32$.

\subsection{PCA Feature Selection}\label{sec:pca}
Hacking intervals can also be used in the context of feature selection. We consider the example where principal components analysis is employed.  There are a number of proposed methods \citep{Mccabe1984, Jolliffe1972, Xu2008}, but we focus on one proposed by \citet{Guo}, which is similar to \citet{Krzanowski1987}. 

In this method, we start with a matrix $\X$ of $n$ observations and $p$ features and we wish to find the matrix $\X_q$ of $n$ observations and $q<p$ of these features that gives the closest approximation to $\X$ when we compare the principal component scores of $\X$ and $\X_q$. That is, we do a principal component decomposition of each matrix, writing $\S=\X\W$ and $\S_q=\X_q\W_q$, where $\S$ and $\S_q$ are the matrices of principal component scores of $\X$ and $\X_q$, respectively, and $\W$ and $\W_q$ the matrices of eigenvectors of $\X^T\X$ and $\X_q^T\X_q$, respectively. We assume the columns of $\W$ and $\W_q$ are ordered from largest to smallest eigenvalue, so that the columns of $\S$ and $\S_q$ are ordered from largest to smallest variance. We then pick an integer $k\le q$ and compare the first $k$ columns (\textit{i.e.}, principal component scores) of $\S$ and $\S_q$ by ``superimposing'' one matrix on the other. That is, we find their Procrustes distance, which compares the matrices after optimal translation, scaling, and rotation. 
If we define the function space $\mathcal{F}$ as the set of all selector functions that map the complete set $\{1,\dotsc,p\}$ of $p$ feature indices to the selected set of $q<p$ feature indices, then the $f$ (i.e., the choice of $q$ features) that maximizes this Procrustes distance for a given $k$ will minimize the following loss function:
\begin{align}
\label{eq:loss_pca}
L(\X,f,k) = \text{trace}(\S^{(k)T} \S^{(k)} + \S_q^{(k)T} \S^{(k)}_q -2\bm{\Sigma})
\end{align}
where $\S^{(k)}$ and $\S_q^{(k)}$ are the first $k$ columns of $\S$ and $\S_q$, respectively, and $\bm{\Sigma}$ is the diagonal matrix of singular values of $\S^{(k)T}\S^{(k)}_q$. Equation (\ref{eq:loss_pca}) represents the loss of structural information in a candidate subset. In practice we scale the score matrices so that the loss is between 0 and 100. The number of component scores $k$ is a hyperparameter. Note that while the researcher must choose the number of selected variables $q$, this number actually defines the problem, so we do not consider it a hyperparameter.

Among all of the $q$ feature subsets that result in a loss of less than a small threshold $\theta$, there are three questions we wish to ask. 
(i) Is a particular subset $j\in\{1,2,\dotsc, {p\choose q}\}$ one of these subsets? That is, does subset $j$ of $q$ features yield a small loss of information? 
(ii) Is feature $i\in\{1,2,\dotsc,p\}$ included or not included in any of these subsets? That is, to achieve a small loss of information using $q<p$ features, can we determine if a particular feature $i$ must or must not be used?
(iii) What is the maximum Hamming distance of these subsets to the optimal subset (assuming we represent the subsets as binary indicators for each feature)? That is, how different could a subset of $q$ features that yields a small loss of information be from the subset of $q$ features that yields the least loss of information? Each of these questions corresponds to a different summary statistic. 
In (i) it is a binary indicator equal to 1 for subset $j$ and $0$ otherwise. 
In (ii) it is a binary indicator equal to 1 if variable $i$ is included in a subset of $q$ features and $0$ if not. 
In (iii) it is the Hamming distance between two subsets. Notice that in the first two cases, the hacking interval is either $[0,0]$, $[1,1]$, or $[0,1]$, while in the last case the hacking interval is between $0$ and, at most, $2q$, since at most $q$ variables can differ.

\begin{figure}
	\centering
	\includegraphics[width=10cm]{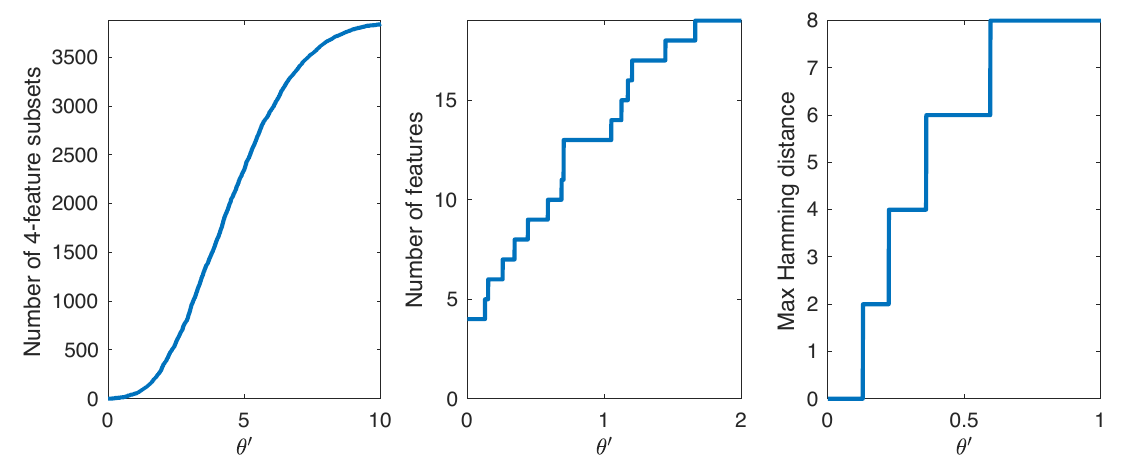}
	\caption{
		\textit{Left:} Number of $q$-feature subsets that yield a loss within the tolerance. 
		\textit{Middle:} Number of unique features that are included in at least one subset that yields a loss within the tolerance. 
		\textit{Right:} Maximum Hamming distance between the optimal $q$-feature solution and any $q$-feature solution that yields a loss within the tolerance. $\theta'$ is a number added to the minimum loss (which is out of 100).}
	\label{fig:pca}
\end{figure}

Several PCA variable selection papers \citep{Jolliffe1972, Guo, Krzanowski1987} have used a dataset on alate alleges (winged aphids), so we will do the same for comparison. This dataset consists of 40 observations of 19 variables. See \citet{Jeffers} for a full description. Keeping with common practice on this dataset \citep{Guo}, we will restrict our analysis to selecting $q=4$ and set a default $k=4$ for the number of principal component scores. Reading the three panels from left to right in Figure \ref{fig:pca} we see answers to our three questions for this dataset. Note that in each case, $\theta'$ is a number added to the minimum loss (which is out of 100).  

Of the fourteen feature selection methods analyzed on this dataset by \citet{Guo}, the selected features have losses given by Equation (\ref{eq:loss_pca}) that differ by as much as 7.65 (although, note that only three of the methods seek to minimize this particular loss) and are concentrated around a group of fifteen of the nineteen features (\textit{i.e.}, four of the features are not selected by any of the fourteen methods).
Our analysis shows that this greatly underestimates the diversity of features that could be selected, where ``selected'' means a feature subset yields a loss within a given tolerance of the minimum loss. 
Examining Figure \ref{fig:pca}, we find that within a loss tolerance of 7.65, more than 92\% of the 3876 possible 4-feature subsets could be selected (left panel); within a loss tolerance of only 1.66, each of the 19 features is contained in at least one 4-feature subset that could be selected (middle panel); and within loss tolerance of only 0.59, a four-feature subset could be selected that is disjoint from the optimal 4-feature subset (right panel). This illustrates the advantage of our systematic approach over the approach of aggregating past studies.  

\section{Proofs}
\let\clearpage\relax 
\label{sec:proofs}
All proofs are available in this section.

\proof{Proof of Proposition \ref{thmSVM} (Hacking Intervals for SVM).}

Notice that the dual problem given by Equation (\ref{eq:svm_dual}) is convex. The last two terms are linear, $1/\beta$ is convex, and the coefficient on this term is positive since it is a square. We will assume Slater's condition is satisfied. That is, that there exists a primal solution for which all inequality constraints are strictly satisfied. In this case, the KKT conditions provide necessary and sufficient conditions for optimality. We start by writing down the Lagrangian:
\begin{multline*}
\mathcal{L}(\blambda, \lambda_0, \bxi, \bm{\alpha}, \bm{r}, \beta) = 
s\blambda^T \xnew + s\lambda_0 
- \sum_{i=1}^n \alpha_i \left[ y_i(\blambda^T \x_i + \lambda_0)-1+\xi_i \right] \\
- \sum_{i=1}^n r_i \xi_i
+ \beta \left[ \frac{1}{2} \norm{\blambda}_2^2 + \psi_d\sum_{i=1}^n \xi_i - \theta \right],
\end{multline*}
where we have introduced dual variables $\{\alpha_i\}_{i=1}^n$, $\{r_i\}_{i=1}^n$, and $\beta$.

Next, we check the KKT conditions:
\begin{itemize}
	
	\item \textit{Lagrangian stationarity}: $\grad \mathcal{L}=0$. So, we have:
	\begin{align}
	\fracpartial{\mathcal{L}}{\blambda} = s\xnew - \sum_{i=1}^n \alpha^*_i y_i \x_i + \beta^* \blambda^* &=0 \nonumber\\
	\blambda^* &= \frac{1}{\beta^*} \left(-s\xnew + \sum_{i=1}^n\alpha^*_i y_i \x_i \right), \label{eq:kkt2}
	\end{align}
	\begin{align}
	\fracpartial{\mathcal{L}}{\lambda_0} = s- \sum_{i=1}^n \alpha^*_i y_i &= 0  \label{eq:kkt3}\\ 
	\sum_{i=1}^n \alpha^*_i y_i & = s, \label{eq:kkt4}
	\end{align}
	\begin{align}
	\fracpartial{\mathcal{L}}{\xi_i} = -\alpha^*_i - r^*_i + \beta^* \psi_d &= 0 \label{eq:kkt5}\\
	r^*_i &= \beta^* \psi_d - \alpha^*_i. \label{eq:kkt6}
	\end{align}
	
	\item \textit{Complementary slackness}:
	There are three conditions:
	\begin{gather}
	\alpha^*_i \left[ y_i(\blambda^{*T} \x_i + \lambda^*_0)-1+\xi^*_i \right] = 0,\  \forall i \label{eq:kkt7}\\
	r^*_i \xi^*_i = 0,\  \forall i \label{eq:kkt8}\\
	\beta^* \left[ \frac{1}{2} \norm{\blambda^*}_2^2 + \psi_d\sum_{i=1}^n \xi^*_i - \theta \right]=0.\nonumber
	\end{gather}
	
	\item \textit{Dual feasibility}: There are three conditions: 
	\begin{gather}
	\alpha^*_i \ge 0,\  \forall i \label{eq:kkt10}\\
	r^*_i \ge0,\  \forall i  \nonumber\\
	\beta^* \ge 0. \label{eq:kkt12}
	\end{gather}
	Notice if we plug in the stationarity condition given by Equation (\ref{eq:kkt6}), we have:
	\begin{gather}
	r^*_i \ge 0 \iff \beta^* \psi_d - \alpha^*_i \ge 0 \iff \alpha^*_i \le \beta^* \psi_d.\label{eq:kkt13}
	\end{gather}
	
	\item \textit{Primal feasibility}: There are three conditions:
	\begin{gather*}
	y_i(\blambda^{*T} x + \lambda^*_0)-1+\xi^*_i \ge 0,\  \forall i \\
	\xi^*_i \ge 0,\  \forall i \\
	\frac{1}{2} \norm{\blambda^*}_2^2 + \psi_d\sum_{i=1}^n \xi^*_i - \theta \le 0.
	\end{gather*}
	
\end{itemize}

Now, let us simplify the Lagrangian:
\begin{equation*}
\begin{split}
\mathcal{L}(\blambda, \lambda_0, \bxi, \alpha, r, \beta) 
&= s\blambda^T \xnew + s\lambda_0 - \blambda\sum\alpha_i y_i \x_i 
- \lambda_0 \sum\alpha_i y_i \\
&\phantom{=} + \sum\alpha_i 
- \sum\alpha_i \xi_i
- \sum r_i \xi_i
+\beta \left[ \frac{1}{2} \norm{\blambda}_2^2 + \psi_d \sum_{i=1}^n \xi_i - \theta \right]
\\
&= \blambda^T \left( s\xnew - \sum\alpha_iy_ix_i\right)
+ \lambda_0 \left\{s - \sum\alpha_iy_i \right\} \\
&\phantom{=} + \sum\alpha_i 
+ \sum \{-\alpha_i - r_i + \beta \psi_d\}\xi_i
+ \beta \frac{1}{2} \norm{\blambda}_2^2
- \beta\theta
\\
&= -\beta \blambda^T \left\{ \frac{1}{\beta} \left( - s\xnew + \sum\alpha_iy_ix_i\right) \right\}
+ \lambda_0 \left\{s - \sum\alpha_iy_i \right\} \\
&\phantom{=} + \sum\alpha_i 
+ \sum \{-\alpha_i - r_i + \beta \psi_d\}\xi_i
+ \beta \frac{1}{2} \blambda^T\blambda
- \beta\theta.
\end{split}
\end{equation*}
Next we plug the KKT conditions (\ref{eq:kkt2}), (\ref{eq:kkt3}), and (\ref{eq:kkt5}) into the Lagrangian:
\begin{align*}
\mathcal{L}(\blambda^*, \lambda^*_0, \bxi^*, \bm{\alpha}^*, \bm{r}^*, \beta^*)
&= -\beta^*\blambda^{*T}\{ \blambda^* \}
+\lambda^*_0 \{ 0\}
+ \sum\alpha^*_i
+ \sum \{0\}\xi^*_i
+\beta^* \frac{1}{2} \blambda^{*T}\blambda^*
- \beta^*\theta
\\
&= -\beta \frac{1}{2} \blambda^{*T}\blambda^* + \sum\alpha^*_i - \beta^*\theta.
\end{align*}
Plugging in Equation (\ref{eq:kkt2}) we have:
\begin{align*}
\mathcal{L}(&\blambda^*, \lambda^*_0, \bxi^*, \bm{\alpha}^*, \bm{r}^*, \beta^*) \\
&= -\beta^* \frac{1}{2} \left\{ \frac{1}{\beta^*} \left(-s\xnew + \sum_{i=1}^n\alpha^*_i y_i \x_i \right)^T  \frac{1}{\beta^*} \left(-s\xnew + \sum_{i=1}^n\alpha^*_i y_i \x_i \right)\right\} 
+ \sum\alpha^*_i - \beta^*\theta 
\\
&= -\frac{1}{2\beta^* } \left[
s^2 \xnewt \xnew 
-2s\sum\alpha^*_i y_i \x_i^T \xnew
+s^2 \sum_i \sum_k \alpha^*_i \alpha^*_k y_i y_k \x_i^T \x_k
\right]
+ \sum\alpha^*_i - \beta^*\theta
\\
&= -\frac{1}{2\beta^*} \left[
\xnewt \xnew 
-2s\sum\alpha^*_i y_i \x_i^T \xnew
+\sum_i \sum_k \alpha^*_i \alpha^*_k y_i y_k \x_i^T \x_k
\right]
+ \sum\alpha^*_i - \beta^*\theta.
\end{align*}
Notice $s^2=1$, so we can eliminate it. 
Since the Lagrangian only depends on the dual variables $\bm{\alpha}$ and $\beta$, the dual problem is:
\begin{align*}
\max_{\bm{\alpha}, \beta}
-\frac{1}{2\beta} \left[
\xnewt \xnew 
-2s\sum\alpha_i y_i \x_i^T \xnew
+\sum_i \sum_k \alpha_i \alpha_k y_i y_k \x_i^T \x_k
\right]
+ \sum\alpha_i - \beta\theta \\
\text{s.t.}\quad
\begin{cases}
0\le \alpha_i \le \beta \psi_d,\  \forall i \\
\sum_{i=1}^n \alpha_i y_i =s \\
\beta \ge 0,
\end{cases}
\end{align*}
where the constraints come from Equations (\ref{eq:kkt10}), (\ref{eq:kkt4}), (\ref{eq:kkt12}), and (\ref{eq:kkt13}). Once the optimal $(\bm{\alpha}^*, \beta^*)$ have been found, we find the optimal $\blambda^*$ from Equation (\ref{eq:kkt2}).

To find $\lambda^*_0$ we can use the complementary slackness conditions. For some $i_{sv}$ such that $r^*_{i_{sv}}>0$ and $\alpha^*_{i_{sv}}>0$, we have $\xi^*_{i_{sv}}=0$ and $y_{i_{sv}}(\blambda^{*T} \x_{i_{sv}} + \lambda^*_0)-1+\xi^*_{i_{sv}}=0$, by Equations (\ref{eq:kkt7}) and (\ref{eq:kkt8}), respectively.
Then, $y_{i_{sv}}(\blambda^{*T} \x_{i_{sv}} + \lambda^*_0)=1$, so $\lambda^*_0 = y_{i_{sv}} - \blambda^{*T} \x_{i_{sv}}$.
\endproof
\hfill \break

\proof{Proof of Theorem \ref{TheoremGaryConfInterval} (Hacking Interval for Least-Squares ATE).}

Let us rewrite the problems as follows:
\begin{equation*}
{\max/\min}_{\bbeta\in\R^p,\beta_0\in\R} \beta_0 \qquad\textrm{ such that }\qquad g(\bbeta,\beta_0)-\theta\leq 0,
\end{equation*}
where $g(\bbeta,\beta_0)=\sum_{i=1}^n\left(y_i-\bbeta\x_i-\beta_0\one_{[i \textrm{ treated}]}\right)^2$.

Let us write some of the KKT stationarity conditions:
\[\grad \beta_0 =\mu \grad g(\bbeta,\beta_0).\] 
In particular, we consider the gradients along the $\beta_j$'s:
\[\textrm{for }j=1\ldots p, \;\left.\frac{\partial \beta_0}{\partial \beta_j}\right|_{\beta_0^*,\bbeta^*}=0,\qquad \left.\frac{\partial g(\bbeta,\beta_0)}{\partial \beta_j}\right|_{\beta_0^*,\bbeta^*}= 
-2\sum_{i=1}^n \left(y_i -\x_i^T\bbeta^* - \beta_0^*\one_{[i \textrm{ treated }]}  \right)x_{ij},
\]
that is:
\[\mathbf{0}=\X^T\mathbf{Y}-\X^T\X\bbeta^* - \beta_0^*\X^T\one_{[\treated]}
\]
and solving for $\bbeta^*$:
\begin{equation}\label{eqn2a}
\bbeta^*=(\X^T\X)^{-1}\X^T\mathbf{Y}-\beta_0^*(\X^T\X)^{-1}\X^T\one_{[\treated]}.
\end{equation}
Changing notation in \eqref{eqn2a},  
\begin{equation}\label{eqn2}
\bbeta^*=\bbeta^*_{LS}-\beta_0^*(\X^T\X)^{-1}\X^T\one_{[\treated]},
\end{equation}
where $\bbeta^*_{LS}= (\X^T\X)^{-1}\X^T\mathbf{Y}$ is the optimal least square solution.
We do not yet know $\beta_0^*$, and for that, we will use complementary slackness.
\begin{equation}\label{eqn1}
\mu g(\bbeta^*\beta_0^*)= 0 \rightarrow g(\bbeta^*\beta_0^*)= 0, \textrm{ that is, }
\sum_{i=1}^n (y_i-\x_i\bbeta^* - \beta_0^*1_{[i\; \treated]})^2-\theta=0.
\end{equation}
Equations \eqref{eqn2} and \eqref{eqn1} will suffice to find all solutions. 
Substituting \eqref{eqn2} into \eqref{eqn1},
\begin{eqnarray*}
	0&=&\sum_{i=1}^n (y_i-\x_i\bbeta^*_{LS}-\beta_0^*
	[\x_i(\X^T\X)^{-1}\X^T\one_{[\treated]} - 1_{[i\; \treated]}])^2-\theta ,\\
	0&=&\sum_i (d_i + \beta_0^* h_i)^2 - \theta,
\end{eqnarray*}
where we have defined differences $d_i:=y_i-\x_i\bbeta^*_{LS}$ and $h_i:=\x_i(\X^T\X)^{-1}\X^T\one_{[\treated]} - 1_{[i\; \treated]}$. Continuing,
\begin{eqnarray*}
	0&=&\left(\sum_i d_i^2-\theta\right) +2\beta_0^* \left(\sum_i h_id_i \right)+ \beta_0^{*2} \sum_i h_i^2.
\end{eqnarray*}
Thus, using the quadratic formula:
\begin{align*}
\beta_0^*&=\frac{
	-2\sum_i h_id_i \pm \sqrt{\left(2\sum_i h_id_i \right)^2-4(\sum_i h_i^2)\left(
		(\sum_id_i^2)-\theta
		\right)
}}
{2\sum_ih_i^2}\\
&=
-\frac{\mathbf{h}^T\mathbf{d}}{\|\mathbf{h}\|^2}
\pm 
\frac{1}{\|\mathbf{h}\|}
\sqrt{\frac{\left(\mathbf{h}^T\mathbf{d} \right)^2}{\|\mathbf{h}\|^2}
	-\|\mathbf{d}\|^2
	+\theta
}.
\end{align*}
Now, suppose $\theta$ were set to the $\SSE$. Then the contents of the square root must become 0, since if this were not true, the solution to the robust optimization problem would disagree with the solution to the least squares minimization problem in the case where $\theta$ is the $\SSE$, which would be a contradiction. In that case, we are back to the least square solution, which must be both $\tilde{\beta}^*_{0,LS}$ and $-\frac{\mathbf{h}^T\mathbf{d}}{\|\mathbf{h}\|^2}$. Next, setting the contents of the square root to 0 and solving for $\theta$ (which was set to the $\SSE$), we find $\theta=\|\mathbf{d}\|^2-\frac{(\mathbf{h}^T\mathbf{d})^2}{\|\mathbf{h}\|^2}$. Putting this together we have:
\begin{align}
\beta_0^*=\bbeta^*_{LS} \pm \frac{1}{\|\mathbf{h}\|} \sqrt{\theta - \SSE}.\label{eq:beta0_h}
\end{align}
Notice we can write $V_{tt}$, the diagonal entry corresponding to the treatment variable of $\left[\tilde{\X}^T\tilde{\X}\right]^{-1}$, as follows:
\begin{align}
V_{tt} &:= 
\left( \left[\tilde{\X}^T\tilde{\X}\right]^{-1}\right)_{tt}  \nonumber
\\ &= 
\left( \left[
\begin{bmatrix}
\X^T \\[10pt]
\one_{\textrm{treated}}^T
\end{bmatrix}
\begin{bmatrix}
\X & \one_{\textrm{treated}}
\end{bmatrix}
\right]^{-1} \right)_{tt} \nonumber
\\[10pt]
&= \left( 
\begin{bmatrix}
\X^T \X & \X^T \one_{\textrm{treated}} \\
(\X^T \one_{\textrm{treated}})^T & \one_{\textrm{treated}}^T\one_{\textrm{treated}})
\end{bmatrix}^{-1}
\right)_{tt}\nonumber
\\[10pt]
&=
\begin{bmatrix}
\left(\X^T \X\right)^{-1} + \frac{1}{k} \left(\X^T \X\right)^{-1} \X^T \one_{\textrm{treated}} \one_{\textrm{treated}}^T \X \left(\X^T \X\right)^{-1}& \frac{1}{k} \left(\X^T \X\right)^{-1} \X^T \one_{\textrm{treated}} \\
\frac{1}{k}\one_{\textrm{treated}}^T \X \left(\X^T \X\right)^{-1}  & \frac{1}{k}
\end{bmatrix}_{tt} \label{eq:mat_inv}
\\[10pt]
&=\frac{1}{k} \label{eq:Vtt},
\end{align}
where $k= \one_{\textrm{treated}}^T \one_{\textrm{treated}} - \one_{\textrm{treated}}^T \X \left(\X^T \X\right)^{-1}  \X^T \one_{\textrm{treated}} = \one_{\textrm{treated}}^T (\I_n - \H) \one_{\textrm{treated}}$, $\H=\X \left(\X^T \X\right)^{-1}  \X^T$ is the hat matrix, and $\I_n$ is an identity matrix of size $n$. Equation (\ref{eq:mat_inv}) follows from a common block matrix inversion formula (see Proposition 2.8.7 in \citet{bernstein}, for example). Next, notice that $\|\mathbf{h}\|^2$ simplifies to:
\begin{eqnarray}
\mathbf{h}^T\mathbf{h}
&=&-\one_\textrm{treated}\X(\X^T\X)^{-1}\X^T\one_{\textrm{treated}}+\one_\textrm{treated}^T\one_\textrm{treated} \nonumber \\
&=&\one_\textrm{treated}^T(\I_n -\X(\X^T\X)^{-1}\X^T)\one_{\textrm{treated}} \nonumber \\
&=&\one_\textrm{treated}^T(\I_n -\H)\one_{\textrm{treated}} \nonumber \\
&=& k. \label{eq:hh}
\end{eqnarray}
Therefore, from Equations (\ref{eq:Vtt}) and (\ref{eq:hh}), we have $V_{tt} = 1/\|\mathbf{h}\|^2$. Plugging this result into Equation (\ref{eq:beta0_h}) we have the desired result for $	\beta_{0,\max}$ and $	\beta_{0,\min}$:
\begin{eqnarray*}
	\beta_{0,\max}^{*}&=&\tilde{\beta}^*_{0,LS}+\sqrt{V_{tt}}\sqrt{\theta - \SSE}\\
	\beta_{0,\min}^{*}&=&\tilde{\beta}^*_{0,LS}-\sqrt{V_{tt}}\sqrt{\theta - \SSE}.
\end{eqnarray*}
Finally, defining $\bm{\gamma}^*_{LS} := (\X^T\X)^{-1}\X^T\one_{[\treated]}$ as the optimal least square solution from regressing $\one_{[\treated]}$ on $\X$, from Equation (\ref{eqn2}) we have:
\begin{eqnarray*}
	\bbeta_{\max}^*&=&\bbeta^*_{LS}-\beta_{0,\max}^{*}\bm{\gamma}^*_{LS},\\
	\bbeta_{\min}^*&=&\bbeta^*_{LS}-\beta_{0,\min}^{*}\bm{\gamma}^*_{LS}.
\end{eqnarray*}

\endproof
\hfill \break

\proof{Proof of Theorem \ref{thm_CI} (ATE Hacking Intervals and Standard Confidence Intervals).}

Equating the upper bound of the least-squares ATE hacking interval $\beta_{0,\max}^{*}$ to the upper bound of a standard confidence interval, given in Equations (\ref{eq:beta_max}) and (\ref{eq:std_CI}), respectively, and solving for $\theta$ we have:
\begin{align*}
\sqrt{V_{tt}}\sqrt{\theta - \SSE} &= \tpn \sqrt{\frac{\SSE}{n-p-1}}\sqrt{V_{tt}} \\
\theta &= \SSE\left(1 + \frac{\tpn^2}{n-p-1}\right).
\end{align*}
The calculation is the same for the lower bounds of the two intervals. 

\endproof 
\hfill \break

\proof{Proof of Theorem \ref{thm_var} (Variance of Least-Squares ATE Hacking Interval Bounds).}

Let $\beta_{0,s}^{*} = \tilde{\beta}^*_{0,LS}+s\sqrt{V_{tt}}\sqrt{\theta - \SSE}$, where $s=1$ gives $\beta_{0,\max}^{*}$ and $s=-1$ gives $\beta_{0,\min}^{*}$ as defined by Equations (\ref{eq:beta_min}) and (\ref{eq:beta_max}), respectively.
It is well-known that for linear regression, the maximum likelihood estimates $\tilde{\bbeta}_{LS} = (\tilde{\X}'\tilde{\X})^{-1}\tilde{\X}^T\Y$ and $\tilde{\sigma}^2_{LS} = \frac{1}{n} (\Y - \tilde{X}\tilde{\bbeta}_{LS})^T(\Y - \tilde{X}\tilde{\bbeta}_{LS})$ for $\bbeta$ and $\sigma^2$, respectively, have the following properties:
\begin{itemize}
	\item \textit{Property 1}: $\tilde{\bbeta}_{LS} \sim N(\bbeta, (\tilde{X}^T \tilde{X})^{-1}\sigma^2)$.  
	
	\item \textit{Property 2}: $\frac{n\tilde{\sigma}_{LS}^2}{\sigma^2} = \frac{\SSE}{\sigma^2} \sim \chi^2_{n-p-1}$. Consequently, $\sqrt{\SSE/\sigma^2}$ has a
	has a chi distribution with mean $\mu$ given by Equation (\ref{eq:mu}) for $n-p-1$ degrees of freedom. 
	
	\item \textit{Property 3}: $\tilde{\bbeta}_{LS}$ and $\tilde{\sigma}^2_{LS}$ are independent conditional on $\tilde{\X}$. Consequently, $\tilde{\beta}^*_{0,LS}$ and $s\sqrt{r V_{tt}}\sqrt{n \tilde{\sigma}^2_{LS}} = s\sqrt{r V_{tt}}\sqrt{\SSE}$ are also independent for fixed $\tilde{\X}$. 
\end{itemize}
Therefore:
\begin{align}
\mathbb{V}\left[ \beta_{0,s}^{*} \mid \tilde{\X} \right]
&= \mathbb{V}\left[ \tilde{\beta}^*_{0,LS}+s\sqrt{V_{tt}}\sqrt{\theta - \SSE} \mid \tilde{\X} \right] \nonumber
\\
&= \mathbb{V}\left[ \tilde{\beta}^*_{0,LS}+s\sqrt{V_{tt}}\sqrt{(1+r)\SSE - \SSE} \mid \tilde{\X} \right] \nonumber
\\
&= \mathbb{V}\left[ \tilde{\beta}^*_{0,LS}+s\sqrt{r V_{tt}}\sqrt{\SSE} \mid \tilde{\X} \right] \nonumber
\\
&= \mathbb{V}\left[ \tilde{\beta}^*_{0,LS}\right]+\mathbb{V}\left[s\sqrt{r V_{tt}}\sqrt{\SSE} \mid \tilde{\X} \right] \label{eq:thm_var_s1}
\\
&= \sigma^2 V_{tt}+s^2r V_{tt}\mathbb{V}\left[\sqrt{\SSE} \mid \tilde{\X} \right]  \label{eq:thm_var_s2}
\\
&= \sigma^2 V_{tt}+r \sigma^2 V_{tt}\mathbb{V}\left[\sqrt{\SSE/\sigma^2} \mid \tilde{\X} \right] \nonumber
\\
&= \sigma^2 V_{tt} + r V_{tt} (n-p-1 - \mu^2)  \label{eq:thm_var_s3}
\\
&= \sigma^2 V_{tt}\left( 1 + r(n-p-1 - \mu^2) \right),\nonumber
\end{align}
where $\mu$ is given by Equation (\ref{eq:mu}).
Equation (\ref{eq:thm_var_s1}) follows from Property 3,
the first term in Equation (\ref{eq:thm_var_s2}) follows from Property 1, and Equation (\ref{eq:thm_var_s3}) follows from the variance formula for a chi distribution.  
Notice the final result does not depend on $s$, so it holds for both $\beta_{0,\max}^{*}$ and $\beta_{0,\min}^{*}$.

\endproof
\hfill \break

\proof{Proof of Theorem \ref{thm_cynthias_problem} (Hacking Intervals for Least-Squares Individual TE).}

Starting again with the stationarity conditions:
\[\grad (\mathbf{x}^{{\rm (new)}}\bbeta) =\mu \grad \left[\sum_{i=1}^n (y_i-\x_i\bbeta)^2 - \theta\right],\]
evaluating the gradients with respect to $\beta_j$, we know that the optimal solution $\bbeta^*$ obeys
\begin{eqnarray*}
	x^{\textrm{(new)}}_j = 2\mu \left[\sum_i (y_i-\x_i\bbeta^*)(-x_{ij}) \right]
\end{eqnarray*}
and the full gradients in vector form are:
\begin{eqnarray*}
	\xnew = 2\mu (\X^T \X \bbeta^*-\X^T\mathbf{Y}).
\end{eqnarray*}
Solving for $\bbeta$,
\begin{eqnarray}\label{eqnC1}
	\bbeta^* =\frac{1}{2\mu} 
	\left[ 
	(\X^T\X)^{-1}\mathbf{x}^{\textrm{(new)}T}  
	\right]
	+
	(\X^T\X)^{-1}\X^T\mathbf{Y} 
	=
	\tilde{\mu} 
	\Upsilon
	+
	\bbeta^*_{LS}
\end{eqnarray}
where $\bbeta^*_{LS}$ is the optimal least squares solution, $(\X^T\X)^{-1}\X^T\mathbf{Y} $, we defined $\tilde{\mu}=1/(2\mu)$, and $\Upsilon=(\X^T\X)^{-1}\xnew$. Again using complementary slackness, 
\[
\mu \left(\sum_{i=1}^n (y_i-\x_i\bbeta^*)^2- \theta\right)=0 \rightarrow \left(\sum_{i=1}^n (y_i-\x_i\bbeta^*)^2- \theta\right)=0.
\]
Substituting from \eqref{eqnC1},
\[
\sum_{i=1}^n (y_i
-\tilde{\mu} \x_i \Upsilon
-\x_i\bbeta^*_{LS})^2- \theta=0, \textrm{ and simplifying yields }
\],
\begin{eqnarray*}
	0&=&\left[\sum_i (y_i -\x_i\bbeta^*_{LS})^2 -\theta \right]- 2\tilde{\mu}\sum_i (\x_i \Upsilon)(y_i -\x_i \bbeta^*_{LS}) + \tilde{\mu}^2\sum_i (\x_i\Upsilon)^2\\
	0&=&\left[\sum_i (y_i -\x_i\bbeta^*_{LS})^2 -\theta \right]- 2\tilde{\mu}V + \tilde{\mu}^2\sum_i (\x_i\Upsilon)^2,
\end{eqnarray*}
where we let $V=\sum_i (\x_i \Upsilon)(y_i -\x_i \bbeta^*_{LS})$. Notice it is equal to zero:
\begin{eqnarray*}
	V &=& \sum_i (\x_i \Upsilon)(y_i -\x_i \bbeta^*_{LS}) \\
	&=& (\X\Upsilon)^T(Y-\X\bbeta^*_{LS}) \\
	&=& (\X (\X^T\X)^{-1} \mathbf{x}^{\textrm{(new)T}}  )^T(Y-X\bbeta^*_{LS}) \\
	&=& (\xnew (\X^T\X)^{-1}X^T)(Y-\X\bbeta^*_{LS}) \\
	&=& \xnew (\X^T\X)^{-1}\X^T Y - \xnew (\X^T\X)^{-1}(\X^T\X) \bbeta^*_{LS} \\
	&=& \xnew \bbeta^*_{LS} - \xnew \bbeta^*_{LS}\\
	&=& 0.
\end{eqnarray*}
Therefore, the quadratic formula yields:
\begin{align*}
\tilde{\mu} 
&= \frac{\pm \sqrt{-4\left(\sum_i (\x_i\Upsilon)^2\right)\left[\sum_i (y_i -\x_i\bbeta^*_{LS})^2 -\theta \right]   }}{2\sum_i (\x_i\Upsilon)^2} \\
&= \frac{\sqrt{\theta -  \SSE}}{\|\X\Upsilon\|}.
\end{align*}
Abusing notation by letting $\tilde{\mu}$ be the positive solution and plugging these solutions back into \eqref{eqnC1} yields the result:
\[
\bbeta^*_- = \bbeta^*_{LS}-\tilde{\mu} \Upsilon, \qquad \bbeta^*_+ = \bbeta^*_{LS}+\tilde{\mu} \Upsilon.
\]
\endproof
\hfill \break

\proof{Proof of Theorem \ref{thm_ATE_CI} (Individual TE Hacking Intervals and Standard Confidence Intervals).}

The boundary points of a standard confidence interval are:
\begin{eqnarray*}
	\xnew \bbeta^*_{LS} &\pm& \tpn \sqrt{\text{MSE} (\xnew (\X^T\X)^{-1} \xnewt)} \\
	\xnew \bbeta^*_{LS} &\pm& \tpn \sqrt{\SSE/(n-p-1)} \sqrt{\xnew \Upsilon}
\end{eqnarray*}
By Theorem \ref{thm_cynthias_problem}, the boundary points of the robust interval are:
\begin{eqnarray*}
	\xnew \bbeta^*_{LS} &\pm& \tilde{\mu} \xnew \Upsilon \\
	\xnew \bbeta^*_{LS} &\pm& \frac{\sqrt{\theta - \SSE}}{\|\X\Upsilon\|} \xnew \Upsilon \\
	\xnew \bbeta^*_{LS} &\pm& \frac{\sqrt{\theta - \SSE}}{\sqrt{(\X\Upsilon)^T(\X\Upsilon)}} \xnew \Upsilon \\
	\xnew \bbeta^*_{LS} &\pm& \frac{\sqrt{\theta - \SSE}}{\sqrt{(\X(\X^T\X)^{-1}\xnewt)^T(\X(\X^T\X)^{-1}\xnewt)}} \xnew \Upsilon \\
	\xnew \bbeta^*_{LS} &\pm& \frac{\sqrt{\theta - \SSE}}{\sqrt{\xnew (\X^T\X)^{-1} \xnewt}} \xnew \Upsilon \\
	\xnew \bbeta^*_{LS} &\pm& \frac{\sqrt{\theta - \SSE}}{\sqrt{\xnew \Upsilon}} \xnew \Upsilon \\
	\xnew \bbeta^*_{LS} &\pm& \sqrt{\theta - \SSE} \sqrt{\xnew \Upsilon} \\
	\xnew \bbeta^*_{LS} &\pm& \sqrt{\theta - \SSE} \sqrt{\xnew \Upsilon}.
\end{eqnarray*}
Comparing the standard and robust confidence intervals, we have:
\begin{eqnarray*}
	\tpn \sqrt{\SSE/(n-p-1)} &=& \sqrt{\theta - \SSE} \\
	\theta &=& \SSE \left(1+\frac{\tpn^2}{n-p-1}\right).
\end{eqnarray*}

\endproof

\proof{Proof of Proposition \ref{thm_gen_bound} (Generalization Bound for Hacked Data).}

As a result of \citet{vapnik1981} and the Vapnik-Chervonenkis-Sauer-Shelah lemma (proved independently by \citet{vapnik1971}, \citet{sauer}, and \citet{shelah}), we have that with probability $1-\delta$:
\begin{equation}
\left| R^{\true}_{\mu} (\fp) - R^{\emp}_{\Zp} (\fp)  \right| \le 2\sqrt{2\frac{h\log{\frac{2eh}{n}}+\log{\frac{4}{\delta}}}{n}}. \label{eq:growth}
\end{equation}
Applying the triangle inequality, Equation (\ref{eq:growth}), the assumptions about $\theta_1$, $\theta_2$, and $\theta_3$, and the calculated $\theta_4$ gives:
\begin{align*}
\left| R^{\true}_{\mu} (\fp) - R^{\emp}_{\Zh} (\fh)  \right|\le
&\left| R^{\true}_{\mu} (\fp) - R^{\emp}_{\Zp} (\fp)  \right| \\
&\quad+\left| R^{\emp}_{\Zp} (\fp) - R^{\emp}_{\Zo} (\fp)  \right| \\
&\quad+\left| R^{\emp}_{\Zo} (\fp) - R^{\emp}_{\Zo} (\fo)  \right| \\
&\quad+\left| R^{\emp}_{\Zo} (\fo) - R^{\emp}_{\Zo} (\fh)  \right| \\
&\quad+\left| R^{\emp}_{\Zo} (\fh) - R^{\emp}_{\Zh} (\fh)  \right| \\
\le & 2\sqrt{2\frac{h\log{\frac{2eh}{n}}+\log{\frac{4}{\delta}}}{n}} + \sum_{i=1}^4 \theta_i.
\end{align*}
\endproof
\hfill \break
\end{appendix}

\section*{Acknowledgments}
Special thanks to Aaron Fisher for insightful comments and assistance with proofs.

\bibliographystyle{plainnat}
\bibliography{./Bibliography/Range_of_Effects}  

\end{document}